\documentclass{article}

\usepackage{microtype}
\usepackage{graphicx}
\usepackage{subcaption}
\usepackage{booktabs} 

\usepackage{hyperref}

\usepackage[preprint]{icml2026}

\usepackage{amsmath}
\usepackage{amssymb}
\usepackage{mathtools}
\usepackage{amsthm}
\usepackage{graphicx}
\usepackage{bm}

\usepackage{makecell}
\usepackage{booktabs}
\usepackage{multirow}
\usepackage{amssymb} 

\usepackage[capitalize,noabbrev]{cleveref}

\theoremstyle{plain}

\theoremstyle{definition}

\theoremstyle{remark}

\usepackage[textsize=tiny]{todonotes}

\icmltitlerunning{General Self-Prediction Enhancement for Spiking Neurons}

\begin{document}

\twocolumn[
  \icmltitle{General Self-Prediction Enhancement for Spiking Neurons}



  \icmlsetsymbol{equal}{*}

  \begin{icmlauthorlist}
    \icmlauthor{Zihan Huang}{pku1}
    \icmlauthor{Zijie Xu}{pku1}
    \icmlauthor{Yifan Huang}{pku1}
    \icmlauthor{Shanshan Jia}{pku1}
    \icmlauthor{Tong Bu}{pku1}
    \icmlauthor{Yiting Dong}{pku1}
    \icmlauthor{Wenxuan Liu}{pku1}
    \icmlauthor{Jianhao Ding}{pku1}
    \icmlauthor{Zhaofei Yu}{pku1}
    \icmlauthor{Tiejun Huang}{pku1}
  \end{icmlauthorlist}

    \icmlaffiliation{pku1}{Peking University, Beijing, China}

  \icmlcorrespondingauthor{Zhaofei Yu}{xxxx}

  \icmlkeywords{Self-prediction, Spiking Neuron, Spiking Neural Networks}

  \vskip 0.3in
]



\printAffiliationsAndNotice{}  

\begin{abstract}

Spiking Neural Networks (SNNs) are highly energy-efficient due to event-driven, sparse computation, but their training is challenged by spike non-differentiability and trade-offs among performance, efficiency, and biological plausibility.
Crucially, mainstream SNNs ignore predictive coding, a core cortical mechanism where the brain predicts inputs and encodes errors for efficient perception.
Inspired by this, we propose a self-prediction enhanced spiking neuron method that generates an internal prediction current from its input–output history to modulate membrane potential.
This design offers dual advantages, it creates a continuous gradient path that alleviates vanishing gradients and boosts training stability and accuracy, while also aligning with biological principles, which resembles distal dendritic modulation and error-driven synaptic plasticity.
Experiments show consistent performance gains across diverse architectures, neuron types, time steps, and tasks demonstrating broad applicability for enhancing SNNs.
\end{abstract}

\section{Introduction}

Spiking Neural Networks (SNNs) constitute a biologically plausible class of neural network models that communicate through discrete, time-encoded spikes~\cite{maass1997networks}. Unlike conventional Artificial Neural Networks (ANNs), which rely on continuous activation values, SNNs activate only when a neuron’s membrane potential crosses a firing threshold~\cite{gerstner2014neuronal}. This event-driven mechanism leads to highly sparse computation, making SNNs exceptionally energy-efficient for deploying intelligent systems on resource-constrained edge hardware~\cite{li2024bic}. Recent neuromorphic platforms such as Loihi~\cite{davies2018loihi}, TrueNorth~\cite{debole2019truenorth}, Darwin~\cite{ma2017darwin} and Tianjic~\cite{pei2019towards} have demonstrated the potential of SNNs to enable low-power, real-time inference. Despite these benefits, training high performance SNNs remains a big challenge. The core issue stems from the binary, non-differentiable nature of spike events, which is not suitable for standard backpropagation. Moreover, simulating temporal dynamics during training often demands substantial memory and computational overhead.

To circumvent these limitations, researchers have developed two main paradigms for building high-performance SNNs: direct training and ANN-to-SNN conversion. Direct training approaches attempt to learn SNN weights end-to-end by approximating gradients through surrogate gradient functions~\cite{neftci2019surrogate,duan2022temporal}, or by leveraging biologically motivated rules like spike-timing-dependent plasticity (STDP)~\cite{masquelier2007unsupervised,masquelier2008spike,masquelier2009competitive}. While these methods aim to fully exploit the temporal information of spiking dynamics, they often suffer from long convergence times, and accuracy gaps on benchmark datasets~\cite{gygax2024elucidatingtheoreticalunderpinningssurrogate}. In contrast, ANN-to-SNN conversion first trains a high-accuracy ANN using standard deep learning methods and then converting it into an SNN that approximates its input–output behavior through rate-based or temporal coding schemes~\cite{cao2015spiking,rueckauer2017conversion,han2020rmp,li2021free,bu2022optimal,huang2024towards}. Although the converted SNN is still more energy-efficient than the original ANN, it usually incurs a non-negligible accuracy drop especially under tight latency. Meanwhile, the resulting SNN merely replicates ANN functionality without leveraging temporal spiking dynamics, and thus lacks biological interpretability.

Despite those advances in SNNs, they largely ignore a core principle of cortical computation: the brain predicts sensory inputs and computes deviations from prediction errors. Evidence shows that individual auditory neurons fire selectively to omitted expected sounds, directly signaling prediction errors at the single-neuron level~\cite{lao2023neuronal,lao2025sound}. Pre-activated neural templates resembling actual stimulus responses further support this predictive coding framework~\cite{sanmiguel2013hearing}, with recent work identifying neurons specifically tuned to negative prediction errors~\cite{yaron2025auditory}. Beyond population-level mechanisms, prediction may also occur within single neurons. \cite{hawkins2016neurons} proposed that dendritic branches act as independent pattern detectors, enabling internal predictions. This aligns with models where dendrites predict somatic activity, and local prediction errors drive synaptic plasticity~\cite{urbanczik2014learning,brea2016prospective,sacramento2018dendritic}. Yet such biologically grounded, neuron-intrinsic predictive mechanisms are missing from most SNNs. 

Inspired by these, we propose a self-prediction enhanced spiking neuron method that is compatible with most spiking neuron model within arbitrary spiking neural networks. The core idea is to enable each neuron to leverage its own historical inputs and outputs to internally generate a self-prediction error to further generate prediction current, which dynamically modulates its membrane potential as an auxiliary input current. This prediction current can pre-activate expected spiking events and generate a prediction error signal when the actual output deviates from the prediction, thereby facilitating or suppressing subsequent spike emissions.
Our main contributions are summarized as follows:
\begin{itemize}
    \item We propose a prediction current based on self-prediction error and demonstrate how it enhances arbitrary spiking neurons.
    \item We provide a thorough analysis of the self-prediction enhanced neuron from the perspectives of dynamics, gradient propagation, and biological plausibility. 
    \item We evaluate our method on image classification tasks across various network architectures, spiking neuron models, and datasets, demonstrating its effectiveness.  
    \item We further validate our approach in sequential classification tasks and reinforcement learning tasks with multiple neuron types to verify its broad applicability and robustness.
\end{itemize}

\section{Related Works}

\subsection{Direct Training of SNNs}
Recent years have witnessed significant progress in directly training deep SNNs. Pioneering works such as STBP~\cite{wu2018spatio}, SuperSpike~\cite{zenke2018superspike}, and SLAYER~\cite{shrestha2018slayer} introduced surrogate gradient methods to avoid the non-differentiability of spike emission, enabling end-to-end supervised learning. Subsequent efforts further enhanced SNN performance through more expressive neuron models such as learnable membrane time constants~\cite{fang2021incorporating}, adaptive firing thresholds with moderate dropout~\cite{wang2022ltmd}, complementary LIF dynamics to mitigate temporal gradient vanishing~\cite{huang2024clif}, and unified gated LIF frameworks that integrate diverse biological features~\cite{yao2022glif}. Concurrently, architectural advances in residual architectures such as Spiking ResNet~\cite{hu2021spiking}, SEW ResNet~\cite{fang2021deep}, and MS-ResNet~\cite{hu2024advancing} have mitigated degradation and enabled stable training of deep networks; Meanwhile, attention-based models including Spikformer~\cite{zhou2022spikformer}, Spike-driven Transformer~\cite{yao2023spike}, and QKFormer~\cite{zhou2024qkformer}, have successfully integrated self-attention mechanisms into the spiking paradigm.

Despite these advances, most approaches prioritize engineering performance over biological interpretability. To bridge this gap, researchers have incorporated recurrent or feedback connections to emulate the brain’s intrinsic recurrent processing~\cite{liang2015recurrent, liang2015convolutional, yin2020effective, liao2016bridging}. While biologically inspired, such recurrent SNNs often suffer from high computational costs, slow convergence, and difficulty in scaling to large architectures, limiting their practical utility. This trade between efficiency and plausibility calls for a new neuron-level mechanism that intrinsically integrates predictive feedback dynamics without relying on costly global recurrence.

\subsection{Predictive Coding Theory}
Predictive coding (PC) aims to minimizes redundancy by propagating prediction errors rather than raw sensory signals~\cite{elias2003predictive}. In neuroscience, PC has been formalized as a hierarchical inference framework where higher areas generate predictions and lower areas compute local mismatches~\cite{huang2011predictive, spratling2017review}.
A growing body of work indicate that prediction operates not only at the network level but within individual neurons~\cite{hawkins2016neurons}. Consistent with this view, several computational models formalize dendritic processing as a form of predictive coding, where basal or apical dendrites learn to predict somatic activity, and synaptic plasticity is driven by local prediction errors~\cite{urbanczik2014learning, brea2016prospective, sacramento2018dendritic}.

Predictive coding has also inspired several SNN models, such as PC-SNN~\cite{lan2022pc}, which uses firing time prediction errors for supervised learning, and SNN-PC~\cite{lee2024predictive}, which employs dedicated positive and negative error neurons for unsupervised reconstruction. 
\cite{yin2018autapses} point out that autapses connections formed by pyramidal cells, where the axon synapses onto the neuron’s own dendrites or soma, are a key component of neural circuitry. However, none of them embed the prediction process within the SNN neuron itself by leveraging its own past inputs and outputs for self-prediction, which can effectively improve efficiency compared to layer-level recurrence prediction error computation. 

To address these gaps, we propose a self-prediction paradigm applicable to diverse spiking neuron models, wherein each neuron uses its past inputs and outputs to compute an online local prediction error signal.

\section{Preliminaries}
\subsection{Neurons in SNNs}\label{sec:neurons_in_snn}
In SNNs, neuron models serve as the fundamental computational units responsible for encoding, transmitting, and processing temporal information through discrete spikes. Among various models, the Leaky Integrate-and-Fire (LIF) neuron is the most widely adopted due to its balance between biological plausibility and computational efficiency.

The dynamics of a LIF neuron layer $l$ at time-step $t$ can be formulated as follows:
\begin{align}
\bm{I}^l[t] &= \bm{x}^{l-1}[t]=\bm{W}^l[t]\bm{s}^{l-1}[t],\label{eq:I_lif}\\
\bm{m}^l[t] &= (1-\frac{1}{\tau^l})\bm{v}^{l}[t-1] + \frac{1}{\tau^l}\bm{I}^l[t], \label{eq:m_lif} \\
\bm{s}^l[t] &= \text{H}(\bm{m}^l[t]-\theta^l),\label{eq:s_lif}
\end{align}
where  $\bm{I}^l[t]$  denotes the synaptic input current, which equal to the weighted output $x^{l-1}[t]$ of last layer,  $\bm{W}^l$  is the weight matrix,  $\bm{s}^{l-1}[t]$  is the spike vector from the previous layer,  $\bm{m}^l[t]$  is the membrane potential before reset,  $\tau^l$  is the membrane time constant,  $\theta^l$  is the firing threshold, and  $\text{H}(\cdot)$  is the Heaviside step function. After spiking, the membrane potential  $\bm{v}^l[t]$  is reset. Two common reset strategies are hard reset and soft reset as in Eq. \eqref{eq:lif_hard_reset} and \eqref{eq:lif_soft_reset}.
\begin{align}
\bm{v}^l[t] &= \bm{m}^{l}[t] - \bm{s}^l[t](\bm{m}^{l}[t]-v_{reset}), \label{eq:lif_hard_reset}  \\
\bm{v}^l[t] &= \bm{m}^{l}[t] - \bm{\theta}^l\bm{s}^l[t], \label{eq:lif_soft_reset} 
\end{align}
where  $v_{\text{reset}}$  is a fixed reset potential often set to 0.

The Integrate-and-Fire (IF) neuron is a simplified variant of the LIF model that omits membrane potential leakage. Its update rule replaces Eq.~\eqref{eq:m_lif} with:
\begin{align}
\bm{m}^l[t] &= \bm{v}^{l-1}[t] + \bm{I}^l[t]. \label{eq:m_if} 
\end{align}
This model assumes perfect integration of inputs until a spike is emitted, making it computationally efficient but less biologically realistic.

The Parametric Leaky Integrate-and-Fire (PLIF) neuron extends the LIF model by treating the membrane time constant  $\tau^l$  as a learnable parameter rather than a fixed hyperparameter. This allows the network to adapt the temporal dynamics of each neuron during training, enhancing its capacity to capture diverse temporal patterns and improving overall learning performance.

The Complementary Leaky Integrate-and-Fire (CLIF) neuron introduces an auxiliary complementary membrane potential to enrich neuronal dynamics. Specifically, CLIF employs soft reset and removes input decay. Its dynamics are defined as:
\begin{align}
\bm{m}^l[t] &= (1-\frac{1}{\tau^l})\bm{v}^{l}[t-1] + \bm{I}^l[t], \label{eq:m_clif} \\
\bm{s}^l[t] &= H(\bm{m}^l[t]-\theta^l), \\
\bm{m}_c^l[t] &= \bm{m}_c^{l}[t-1]\cdot\sigma(\frac{1}{\tau^l}\bm{m}^{l-1}[t]) + \bm{s}^l[t], \\
\bm{v}^l[t] &= \bm{m}^{l}[t] - \bm{s}^l[t](\theta^l+\sigma(\bm{m}^l_c[t])),
\end{align}
where  $\bm{m}_c^l[t]$  is a complementary membrane state that accumulates spike history modulated by a sigmoid gate  $\sigma(\cdot)$ , and the reset magnitude is dynamically adjusted based on both the threshold and this complementary state.

Together, various neuron models represent a spectrum of design choices in SNNs, ranging from simplicity and efficiency to enhanced temporal expressivity and adaptive dynamics.

%
\subsection{SNN Training with Surrogate Gradient}
In SNNs employing common neuron models such as IF, LIF, and PLIF, backpropagation through time follows the chain rule.  
As suggested by \cite{neftci2019surrogate}, removing the gradient path through the reset terms in Eqs.~\eqref{eq:lif_hard_reset} and \eqref{eq:lif_soft_reset} improves training stability and overall network performance. The resulting gradient computations are as follows:
\begin{align}
&\frac{\partial L}{\partial \bm{I}^l[t]} = \frac{\partial L}{\partial \bm{m}^l[t]}\frac{\partial \bm{m}^l[t]}{\partial \bm{I}^l[t]}, \label{eq:L_I}\\
&\scalebox{1.2}{$
\frac{\partial L}{\partial \bm{m}^l[t]} = \frac{\partial L}{\partial \bm{s}^l[t]}\frac{\partial \bm{s}^l[t]}{\partial \bm{m}^l[t]} +\frac{\partial L}{\partial \bm{m}^l[t+1]}\frac{\partial \bm{m}^l[t+1]}{\partial \bm{m}^l[t]}. \label{eq:L_m}
$}
\end{align}
The term $\frac{\partial \bm{s}^l[t]}{\partial \bm{m}^l[t]}$ is non-differentiable because $\bm{s}^l[t] = \text{H}(\bm{m}^l[t] - \theta^l)$ involves the Heaviside step function. To enable gradient-based optimization, surrogate gradient methods replace this derivative with a smooth, differentiable approximation. A common choice is the sigmoid function:
\begin{align}
    \frac{\partial \bm{s}^l[t]}{\partial \bm{m}^l[t]} \approx \sigma'(\bm{m}^l[t] - \theta^l), \quad 
\end{align}
where $\quad \sigma(x) = \frac{1}{1 + e^{-k x}}, k > 0$. $k$ controls the steepness of the surrogate, larger $k$ yields a closer approximation to the true Heaviside function. With this substitution, gradients can be propagated through spiking layers, enabling end-to-end training of SNNs while preserving their discrete, event-driven nature in the forward pass.

\section{Method}
In this section, we first introduce our self-prediction enhanced neuron method from the perspective of predictive coding and provide a detailed analysis of its dynamics. We then explain why self-prediction enhances neuronal efficacy based on gradient analysis, and finally offer a biologically plausible interpretation of the proposed mechanism.

\subsection{Self-prediction Enhanced Spiking Neurons}\label{sec:pred_neuron}

\begin{figure}[t]
\begin{center}
\vskip 0.1in
\centerline{\includegraphics[width=1.0\columnwidth, trim=0.0cm 0.0cm 0.0cm 0.0cm, clip]{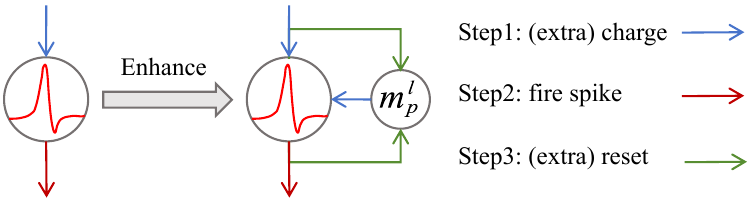}}
\caption{Left: original neuron, Right: our proposed self-prediction enhanced neuron. Neurons operate in the sequence of charging, firing, and resetting at each time-step.}
\label{pic:neuron}
\end{center}
\vskip -0.2in
\end{figure}

Predictive coding theory posits that the brain continuously generates internal predictions about sensory inputs and updates its internal model using the prediction error, which is the discrepancy between actual observations and predictions. At the neuronal level, many neurons exhibit pre-activate activity prior to an expected event, which can be interpreted as a prediction about whether a spike should occur at the current time-step.

Inspired by this principle, we propose that if a spiking neuron can locally predict its next state based on its own history of inputs and outputs, and inject this prediction into its dynamics as an auxiliary current, it can simultaneously enhance both biological plausibility and computational performance.
A schematic illustration of the self-prediction enhanced neuron is shown in Figure~\ref{pic:neuron}.
Specifically, we introduce a weak prediction current $ \bm{m}_p^l[t] $ , which is added to the original input current to dynamically modulate the membrane potential. The design objectives of this current are as follows:

(1) When a spike is predicted, the membrane potential is slightly elevated to bring it close to but still below the firing threshold, thereby pre-activating the neuron in anticipation of the expected event;

(2) When no spike is predicted, moderately lower the membrane potential to suppress noise-induced unexpected spikes;

(3) When a spike is predicted but not emitted, generate a positive prediction error to increase sensitivity to similar future inputs;

(4) When no spike is predicted but one is emitted, generate a negative prediction error to suppress over-responsive behavior in the future.

Begin by regarding prediction error term as:
\begin{align}
\bm{x}^{l-1}[t] - \bm{s}^l[t]/\tau^l,\label{eq:pred_error_term}
\end{align}
where, $ \tau^l $  is the neuronal membrane time constant. The division by  $ \tau^l $  compensates for the exponential decay of input currents, rendering  $ \bm{s}^l[t]/\tau^l $  a reasonable estimate of the effective contribution of the current spike to the membrane potential.

Consider a standard spiking neuron in layer  $ l $ , with raw input  $ \bm{x}^{l-1}[t] $  and binary output  $ \bm{s}^l[t] \in \{0,1\} $ . We enhance the input by introducing the prediction current  $ \bm{m}_p^l[t] $ :
\begin{align}
\bm{I}^l[t] &= \bm{x}^{l-1}[t] + \bm{m}_p^l[t-1].
\end{align}

The prediction current is updated dynamically via a low-pass filter that computes a moving average of the prediction error:
\begin{align}
\bm{m}_p^l[t]=(1 - \tau_p^l) \bm{m}_p^l[t-1] + \tau_p^l ( \bm{x}^{l-1}[t] - \frac{\bm{s}^l[t]}{\tau^l}),\label{eq:m_p_raw}
\end{align}
where $ \tau_p^l \in (0,1) $  controls the update rate of the prediction current.

\begin{figure}[t]
\vskip 0.1in
\begin{center}
\centerline{\includegraphics[width=1.0\columnwidth, trim=0.0cm 0.0cm 0.0cm 0.0cm, clip]{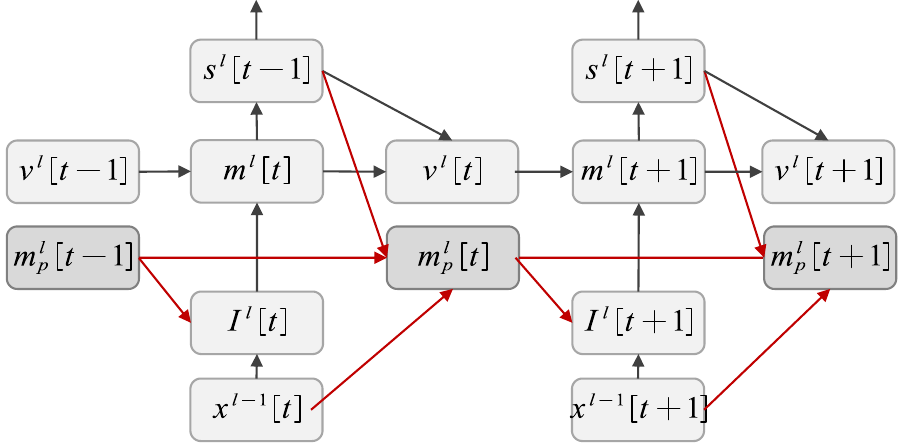}}
\caption{Forward propagation pathway of the self-prediction enhanced LIF neuron. The red lines indicate the additional forward path associated with self-prediction.}
\label{pic:forward}
\end{center}
\vskip -0.2in
\end{figure}

Specifically, taking a hard-reset LIF neuron as an example, the full dynamics of the enhanced neuron can be written as:
\begin{align}
\bm{I}^l[t] &=\bm{x}^{l-1}[t]+\bm{m}_p^l[t-1],\\
\bm{m}^l[t] &= (1-\frac{1}{\tau^l})\bm{v}^{l-1}[t] + \frac{1}{\tau^l}\bm{I}^l[t],\\
\bm{s}^l[t] &= \text{H}(\bm{m}^l[t]-\theta^l), \\
\bm{v}^l[t] &= \bm{m}^{l}[t] - \bm{s}^l[t](\bm{m}^{l}[t]-v_{reset}),\\
\bm{m}_p^l[t] &= (1-\tau^l_p)\bm{m}_p^l[t-1]+\tau^l_p(\bm{x}^{l-1}[t]-\frac{\bm{s}^l[t]}{\tau^l}).\label{eq:lif_m_p_raw}
\end{align}
The forward propagation pathway is shown in Figure~\ref{pic:forward}.

Noe analyzing the prediction error term  $ \bm{x}^{l-1}[t] - \bm{s}^l[t]/\tau^l $  in Eq.\eqref{eq:lif_m_p_raw} from a dynamical perspective:

(1) \textbf{If the current input is high and a spike is emitted}, i.e., $\bm{x}^{l-1}[t]$ is large and $\bm{s}^l[t] = 1$. This implicitly indicates that the prediction at the current time-step is correct and that the neuron tends to predict a spike at the next time-step. Typically, $(\bm{x}^{l-1}[t] - \bm{s}^l[t]/\tau^l)$ should be close to zero but slightly positive. This is because, in previous time-steps, $\bm{x}^{l-1}[t]$ must have exceeded $\bm{s}^l[t]/\tau^l$ to trigger the spike. The predictive current raise the membrane potential, thereby moderately increasing the likelihood of the expected spike and promoting its stable generation.

(2) \textbf{If the current input is low and no spike is emitted}, i.e., $\bm{x}^{l-1}[t]$ is small and $\bm{s}^l[t] = 0$. This implicitly indicates that the prediction at the current time-step is correct and that the neuron tends to predict no spike at the next time-step. The error term should be close to zero or negative, and the predictive current tends toward zero or becomes slightly negative, keeping the membrane potential at a low level and effectively suppressing noise-induced spurious spikes. If the error term is slightly greater than zero and the input remains low in subsequent time-steps, no spike will be emitted. Due to the decay mechanism, both the membrane potential $\bm{v}^l[t]$ and the predictive current $\bm{m}_p^l[t]$ remain at low levels.

(3) \textbf{If the current input is high but no spike is emitted}, i.e., $\bm{x}^{l-1}[t]$ is large yet $\bm{s}^l[t] = 0$. This indicates a prediction error that a spike was expected but did not occur. In this case, $\bm{m}_p^l[t]$ increases in the positive direction, amplifying the missed signal in subsequent time-steps and driving the network to enhance its sensitivity to inputs.

(4) \textbf{If the current input is low yet an unexpected spike occurs}, i.e., $\bm{x}^{l-1}[t]$ is small but $\bm{s}^l[t] = 1$. This also reflects a prediction error that a spike occurred when none was expected. Consequently, $\bm{m}_p^l[t]$ is updated in the negative direction, encouraging the network to learn to suppress such error responses in the subsequent time-step.

Moreover, the moving average form in Eq.~\eqref{eq:lif_m_p_raw} offers two key advantages. One is that it establishes a learnable temporal window that links past inputs to current outputs, enabling short-term prediction; and Another is the low-pass filtering property suppresses high-frequency noise, allowing the prediction current to robustly reflect genuine patterns.

\subsection{Gradient Analysis}\label{sec:gradient}

For a typical spiking neuron model, the gradient with respect to the membrane potential  $\bm{m}^l[t]$ as in Eqs.~\eqref{eq:L_I} and \eqref{eq:L_m} can be expanded over time as follows:
\begin{align}
\frac{\partial L}{\partial \bm{m}^l[t]} = \sum_{i=t}^{T}\frac{\partial L}{\partial \bm{s}^l[i]}\frac{\partial \bm{s}^l[i]}{\partial \bm{m}^l[i]} \prod_{j=t}^{i-1}\frac{\partial \bm{m}^l[j+1]}{\partial \bm{m}^l[j]}.
\end{align}
However, if few spike occurs at time-step $t$, $\frac{\partial \bm{s}^l[i]}{\partial \bm{m}^l[i]}$  becomes very sparse, leading to minimal gradients flowing back to $\bm{m}^l[t]$. This results in limited information being propagated back to  $\bm{x}^{l-1}[t]$.

To address this issue, our proposed self-prediction enhanced neuron model introducing a predictive auxiliary state $\bm{m}_p^l[t]$. The gradients for this enhanced model are given by:
\begin{align}
&\frac{\partial L}{\partial \bm{x}^{l-1}[t]} = \frac{\partial L}{\partial \bm{m}^l[t]}\frac{\partial \bm{m}^l[t]}{\partial \bm{x}^{l-1}[t]}+\frac{\partial L}{\partial \bm{m}_p^l[t]}\frac{\partial \bm{m_p}^l[t]}{\partial \bm{x}^{l-1}[t]},\\
&
\begin{aligned}
    \frac{\partial L}{\partial \bm{m}^l[t]} =& \frac{\partial L}{\partial \bm{s}^l[t]}\frac{\partial \bm{s}^l[t]}{\partial \bm{m}^l[t]} +\frac{\partial L}{\partial \bm{m}^l[t+1]}\frac{\partial \bm{m}^l[t+1]}{\partial \bm{m}^l[t]}
    \\&+\frac{\partial L}{\partial \bm{m}_p^l[t]}\frac{\partial \bm{m}_p^l[t]}{\partial \bm{s}^l[t]}\frac{\partial \bm{s}^l[t]}{\partial \bm{m}^l[t]},
\end{aligned}\\
&
\begin{aligned}
\frac{\partial L}{\partial \bm{m}_p^l[t]} =& \frac{\partial L}{\partial \bm{m}^l[t+1]}\frac{\partial \bm{m}^l[t+1]}{\partial \bm{m}_p^l[t]} \\
&+\frac{\partial L}{\partial \bm{m}_p^l[t+1]}\frac{\partial \bm{m}^l_p[t+1]}{\partial \bm{m}_p^l[t]}.
\end{aligned}
\end{align}

\begin{figure}[t]
\begin{center}
\centerline{\includegraphics[width=1.0\columnwidth, trim=0.0cm 0.0cm 0.0cm 0.0cm, clip]{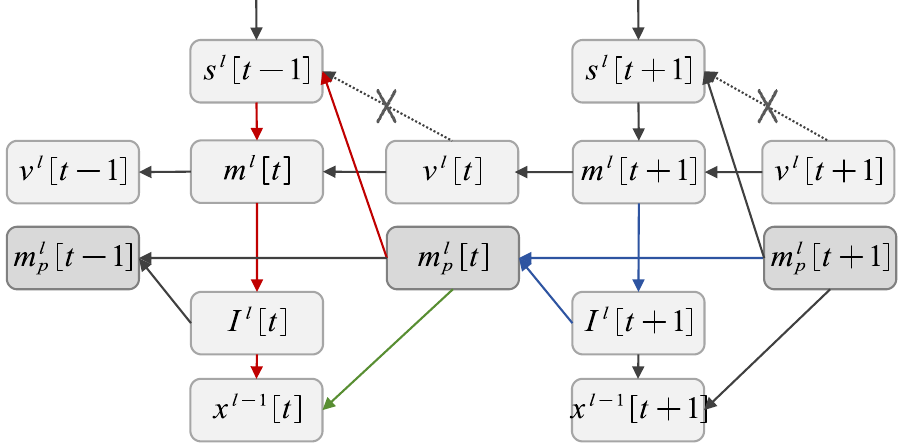}}
\caption{Backward propagation pathway of the self-prediction enhanced LIF neuron. The blue and green lines form the first additional gradient path, and the blue and red lines form the second additional gradient path.The dashed lines indicate paths that are detached from the computational graph.}
\label{pic:backward}
\end{center}
\vskip -0.2in
\end{figure}

Following \cite{neftci2019surrogate}, we remove the gradient path through the reset terms, but we keep the gradient path from $\bm{m}^l_p[t]$ to $\bm{s}^l[t]$.
Compared to the original gradient, $\frac{\partial L}{\partial \bm{x}^{l-1}[t]}$ now includes two additional pathways, as shown in Figure~\ref{pic:backward}:

\textbf{1. Direct pathway: via $\bm{m}_p^l[t] \rightarrow \bm{x}^{l-1}[t]$.}
This direct path bypasses the current-time spike $\bm{s}^l[t]$ and membrane potential $\bm{m}^l[t]$, thereby avoiding gradient fragmentation caused by the sparsity of spiking activity. At each time-step, $\bm{m}_p^l[t]$ can directly inherit gradients from both $\bm{m}_p^l[t+1]$ and $\bm{m}^l[t+1]$. Consequently, even if the neuron fires at time $t+1$, the gradient information is well preserved. By strengthening the influence of future-time gradients on the current time-step, this mechanism enhances the stability of prediction.

\textbf{2. Indirect pathway: $\bm{m}_p^l[t] \rightarrow \bm{s}^l[t] \rightarrow \bm{m}^l[t] \rightarrow \bm{x}^{l-1}[t]$.}
This indirect path leverages the predictive nature of the model to enable more comprehensive gradient flow. Notably, we do not decouple the computational graph between $\bm{m}_p^l[t]$ and $\bm{s}^l[t]$. Unlike the reset mechanism which discards historical information, we treat the auxiliary state $\bm{m}_p^l[t]$ as a continuous and smooth memory unit. As shown in Eq.~\eqref{eq:lif_m_p_raw}, although the update of $\bm{m}_p^l[t]$ involves the spike $\bm{s}^l[t]$, the prediction error is jointly regulated by both input and output. Gradients through this pathway are smoothly propagated via surrogate gradients, enabling the network to adaptively correct prediction errors.

As analyzed in Section~\ref{sec:pred_neuron}, the magnitude of $\bm{m}_p^l[t]$ remains small when there is no prediction error, but increases when a prediction error occurs. In such cases, the self-prediction–enhanced neuron actively learns from this error. By maintaining gradient pathways associated with $\bm{m}_p^l[t]$, the model better captures temporal dependencies and achieves improved overall performance.

\subsection{Biological Interpretability}




Our proposed self-prediction enhanced spiking neuron is not only inspired by the computational principles of predictive coding but also grounded in biological observations of real neuronal mechanisms. In particular, the model closely resembles the modulatory role of distal dendrites in pyramidal neurons.

In biological neural systems, distal dendritic compartments receive top-down inputs that typically do not directly elicit somatic spikes. Instead, they modulate the neuron’s response to feedforward inputs, which effectively implementing a local predictive mechanism. The prediction current $\bm{m}_p^l[t]$ in our model serves as a functional analog of this biological mechanism. It does not directly participate in spike generation but acts as a learnable, subthreshold modulatory signal. Based on the joint statistics of past inputs $\bm{x}^{l-1}[t]$ and outputs $\bm{s}^{l}[t]$, it continuously estimates the expected membrane change at the next time-step and accordingly fine-tunes the prediction current. Bring it closer to the threshold when a spike firing is predicted, and farther away when no spike is expected, thereby enabling local self-prediction.

Moreover, the error-driven update rule for  $\bm{m}_p^l[t]$ aligns closely with synaptic plasticity. In biological synapses, prediction errors, such as strong feedforward input failing to stimulate an expected somatic spike or spontaneous spiking in the absence of supportive input, are thought to trigger calcium-dependent plasticity mechanisms in dendrites that adjust synaptic efficacy to minimize future prediction errors.
In our model, the bottom-up drive is simplified to a local prediction error $\bm{x}^{l-1}[t] - \bm{s}^l[t]/\tau^l$, which error effectively reflects the prediction of the membrane potential changes. The total predictive drive $\bm{m}_p^l[t]$ is then updated slowly and adaptively with a time constant $\tau^l_p$, the dynamic process that closely mirrors the low-pass filtering properties exhibited by dendritic compartments. Thus, our model provides a functionally plausible and concise abstraction of how spiking neurons can leverage simplified dendritic compartmentalization and implicit local learning rules to implement predictive coding.

In summary, the proposed self-prediction enhanced neuron implements local predictive coding in SNNs through a concise mechanism that mimics dendritic modulation and error-driven plasticity.

\begin{table*}[t]
\centering
\caption{Ablation Study on CIFAR10 Classification Tasks. S-neuron stands for self-prediction enhanced neuron. eg. S-IF is the self-prediction enhanced IF neuron.}
\label{tab:ablation_classification_cifar10}
\begin{center}
\begin{small}
\begin{tabular}{cccccccc}
\toprule
Network & Neuron & T & Accuracy & Network & Neuron & T & Accuracy\\
\midrule
Spiking CIFAR10Net&IF&4,8&94.33,94.21
&Spiking CIFAR10Net&PLIF&4,8&93.85,94.05\\
Spiking CIFAR10Net&S-IF&4,8&94.24$\color{blue}{\downarrow}$,94.84$\color{red}{\uparrow}$
&Spiking CIFAR10Net&S-PLIF&4,8&94.39$\color{red}{\uparrow}$,94.85$\color{red}{\uparrow}$\\
\cmidrule(l){2-4}\cmidrule(l){6-8}
Spiking CIFAR10Net&LIF&4,8&93.54,93.60&Spiking CIFAR10Net&CLIF&4,8&94.20,94.23\\
Spiking CIFAR10Net&S-LIF&4,8&93.84$\color{red}{\uparrow}$,94.85$\color{red}{\uparrow}$
&Spiking CIFAR10Net&S-CLIF&4,8&94.70$\color{red}{\uparrow}$,94.38$\color{red}{\uparrow}$\\
\cmidrule(l){1-8}
SEW ResNet18&IF&4,8&94.53, 94.50
&SEW ResNet34&IF&4,8&95.49,95.58\\
SEW ResNet18&S-IF&4,8&94.20$\color{blue}{\downarrow}$,94.78$\color{red}{\uparrow}$
&SEW ResNet34&S-IF&4,8&95.59$\color{red}{\uparrow}$,95.73$\color{red}{\uparrow}$\\
\cmidrule(l){2-4}\cmidrule(l){6-8}
SEW ResNet18&LIF&4,8&94.99, 95.04
&SEW ResNet34&LIF&4,8&95.42,94.54\\
SEW ResNet18&S-LIF&4,8&95.04$\color{red}{\uparrow}$,95.56$\color{red}{\uparrow}$
&SEW ResNet34&S-LIF&4,8&95.86$\color{red}{\uparrow}$,96.16$\color{red}{\uparrow}$\\
\cmidrule(l){2-4}\cmidrule(l){6-8}
SEW ResNet18&PLIF&4,8&95.21,95.04
&SEW ResNet34&PLIF&4,8&95.57,93.49\\
SEW ResNet18&S-PLIF&4,8&95.25$\color{red}{\uparrow}$,95.73$\color{red}{\uparrow}$
&SEW ResNet34&S-PLIF&4,8&95.87$\color{red}{\uparrow}$,96.33$\color{red}{\uparrow}$\\
\cmidrule(l){2-4}\cmidrule(l){6-8}
SEW ResNet18&CLIF&4,8&94.92,95.31
&SEW ResNet34&CLIF&4,8&95.54,94.87\\
SEW ResNet18&S-CLIF&4,8&95.42$\color{red}{\uparrow}$,95.26$\color{blue}{\downarrow}$
&SEW ResNet34&S-CLIF&4,8&96.11$\color{red}{\uparrow}$,95.99$\color{red}{\uparrow}$\\
\cmidrule(l){1-8}
Spiking ResNet18&IF&4,8&94.93,95.16
&Spiking ResNet34&IF&4,8&94.89, 94.96\\
Spiking ResNet18&S-IF&4,8&94.98$\color{red}{\uparrow}$,95.42$\color{red}{\uparrow}$
&Spiking ResNet34&S-IF&4,8&94.67$\color{blue}{\downarrow}$,94.96\\
\cmidrule(l){2-4}\cmidrule(l){6-8}
Spiking ResNet18&LIF&4,8&93.43,94.21
&Spiking ResNet34&LIF&4,8&91.45,87.71\\
Spiking ResNet18&S-LIF&4,8&95.05$\color{red}{\uparrow}$,95.41$\color{red}{\uparrow}$
&Spiking ResNet34&S-LIF&4,8&92.43$\color{red}{\uparrow}$,93.82$\color{red}{\uparrow}$\\
\cmidrule(l){2-4}\cmidrule(l){6-8}
Spiking ResNet18&PLIF&4,8&94.26,94.30
&Spiking ResNet34&PLIF&4,8&93.16,88.93\\
Spiking ResNet18&S-PLIF&4,8&95.05$\color{red}{\uparrow}$,95.43$\color{red}{\uparrow}$
&Spiking ResNet34&S-PLIF&4,8&93.48$\color{red}{\uparrow}$,92.76$\color{red}{\uparrow}$\\
\cmidrule(l){2-4}\cmidrule(l){6-8}
Spiking ResNet18&CLIF&4,8&95.16,95.52
&Spiking ResNet34&CLIF&4,8&94.58,94.57\\
Spiking ResNet18&S-CLIF&4,8&95.29$\color{red}{\uparrow}$,95.67$\color{red}{\uparrow}$
&Spiking ResNet34&S-CLIF&4,8&95.04$\color{red}{\uparrow}$,94.60$\color{red}{\uparrow}$\\
\bottomrule
\end{tabular}
\end{small}
\end{center}
\vskip -0.2in
\end{table*}

\begin{table}[t]
\centering
\caption{Ablation Study on ImageNet100 and ImageNet1k Classification Tasks.}
\label{tab:ablation_classification_imagenet}
\begin{center}
\begin{small}
\begin{tabular}{ccccccc}
\toprule
Dataset & Network & Neuron & T & Accuracy\\
\midrule
\multirow{5}{*}{ImageNet100}
&SEW ResNet34&LIF&4&65.80\\
&SEW ResNet34&S-LIF&4&69.06$\color{red}{\uparrow}$\\
\cmidrule(l){3-5}
&SEW ResNet34&PLIF&4&68.30\\
&SEW ResNet34&S-PLIF&4&68.94$\color{red}{\uparrow}$\\
\cmidrule(l){1-5}
\multirow{2}{*}{ImageNet1k}
&SEW ResNet18&PLIF&4&64.25\\
&SEW ResNet18&S-PLIF&4&64.45$\color{red}{\uparrow}$\\
\bottomrule
\end{tabular}
\end{small}
\end{center}
\vskip -0.2in
\end{table}

\section{Experimental Results}
To validate the effectiveness of our proposed self-prediction enhanced spiking neuron method, we first conducted a series of ablation studies on the CIFAR-10 dataset. These experiments systematically evaluated performance under various neuron types, network architectures, and time-step configurations, thoroughly demonstrating the method's generality and effectiveness. Subsequently, we evaluated our approach on larger-scale datasets ImageNet-100 and ImageNet-1K, further confirming its strong generalization capability in complex scenarios. Additionally, we applied our method to the sequential CIFAR-10 task, showcasing its enhanced temporal modeling capacity. Finally, we tested the method in reinforcement learning tasks, verifying its broad applicability and robustness across diverse task settings.


\subsection{Ablation on Classification Tasks}


We first validate the effectiveness of our proposed method on the CIFAR-10 dataset. The experiments employ five network architectures, including Spiking CIFAR10Net, SEW ResNet-18, SEW ResNet-34, Spiking ResNet-18, and Spiking ResNet-34; and four spiking neuron models, including IF, LIF, PLIF, and CLIF neuron as detail in Section~\ref{sec:neurons_in_snn}, trained under time-steps $T=4$ and $T=8$. 

For the SEW ResNet and Spiking ResNet families which is originally designed for ImageNet, when adapting them to CIFAR-10, we modify the first convolutional layer parameters, changing kernel size, stride, and padding from 7, 2, 3 to 3, 1, 1, respectively, and replace the initial max pooling layer with an identity mapping. For the spiking CIFAR10Net, all Conv2d layers use a kernel size of 3, stride of 1, and padding of 1, with 256 output channels. Each Conv2d layer is followed by a batch normalization layer and a spiking neuron layer. The final feature map is passed through two fully connected layers for classification. Detailed architecture configuration is provided in the appendix.

  
Notably, we employ the learnable parameter $\tau_p^l$ in Eq.\eqref{eq:m_p_raw} per layer. For instance, in ResNet-18, this introduces only 18 additional parameters, which is negligible compared to the total number of network parameters.

The overall ablation results are presented in Table~\ref{tab:ablation_classification_cifar10}.
The results show that our self-prediction enhancement yields significant performance improvements with LIF, PLIF, and CLIF neurons, and also achieves gains in most configurations with IF neurons. For the few cases where performance slightly degrades, we attribute this to two main factors: first, the IF neuron lacks an input decay mechanism, with its time constant fixed at $ \tau=1 $ , making it difficult to effectively balance the presynaptic input $ \bm{x}^{l-1}[t] $ and the postsynaptic spike output $ \bm{s}^l[t] $ in the predict error term $ \bm{x}^{l-1}[t] - \bm{s}^l[t]/\tau $ of Equation~\eqref{eq:pred_error_term}; this insight also suggests a promising direction for future refinement. Second, the performance gain of our method becomes more consistent and stable as the number of simulation time steps increases.

\begin{figure}[htb]
\begin{center}
\centerline{\includegraphics[width=0.5\columnwidth, trim=1.5cm 1.5cm 1.5cm 1.5cm, clip]{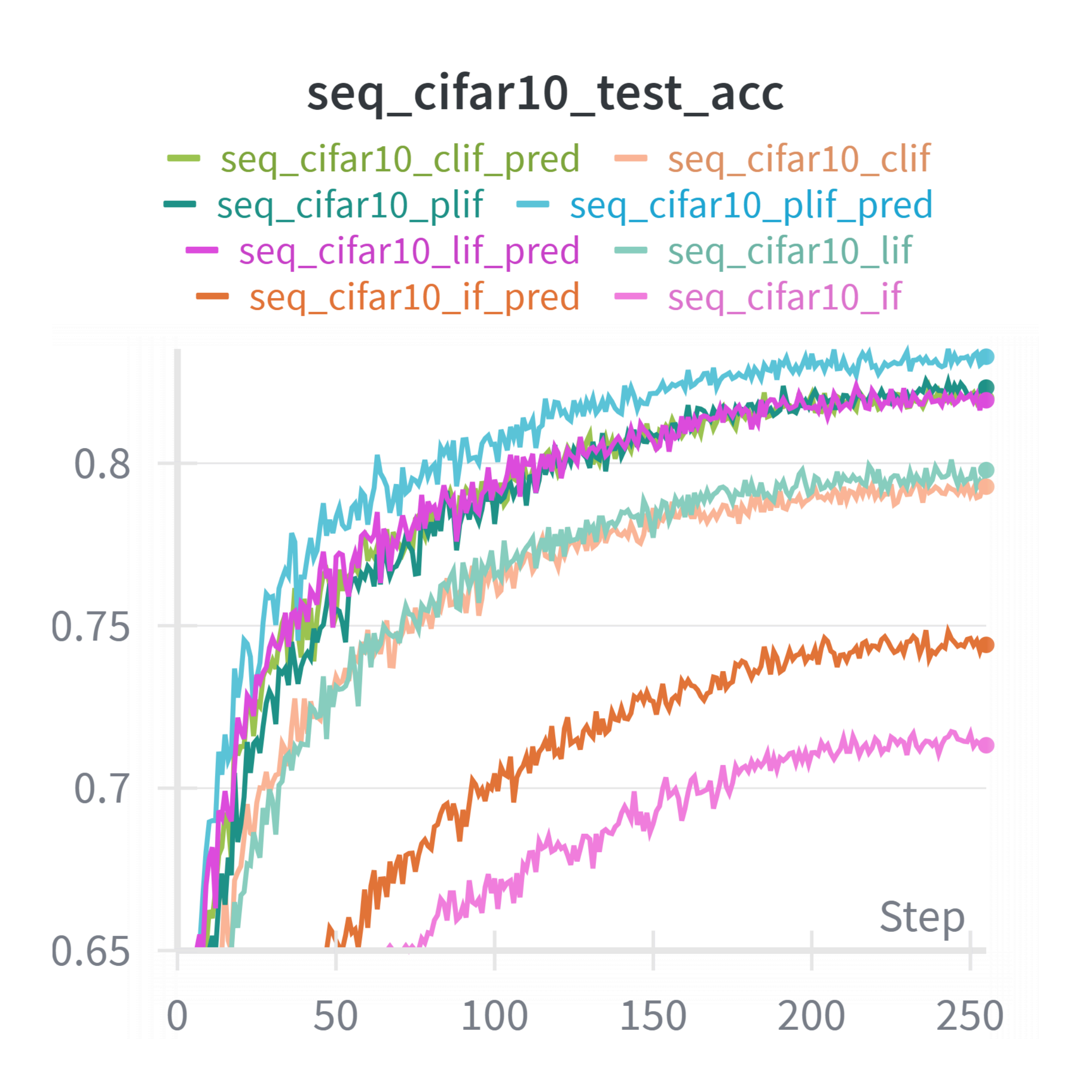}\includegraphics[width=0.5\columnwidth, trim=1.5cm 1.5cm 1.5cm 1.5cm, clip]{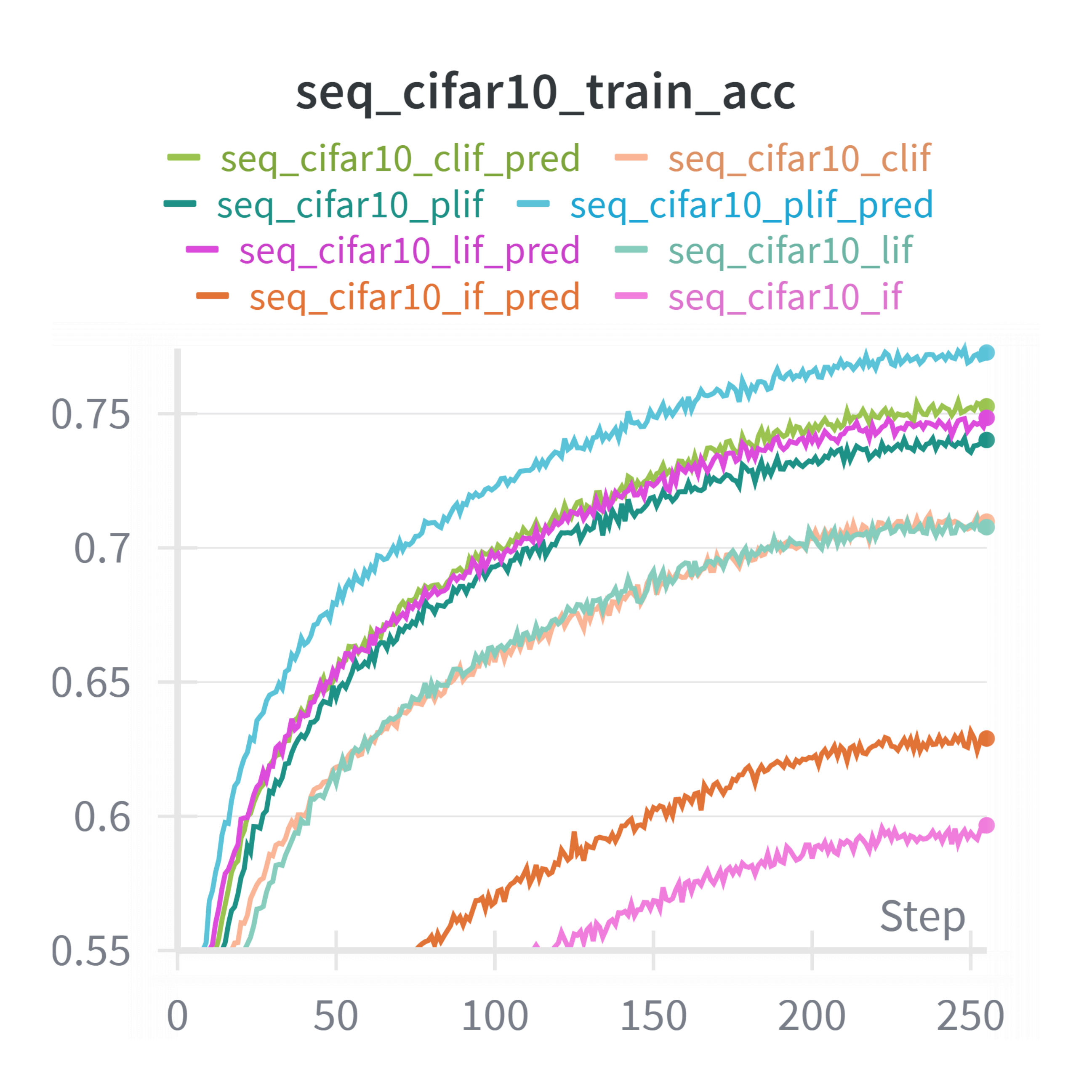}}
\caption{The training and testing accuracy curves of four spiking neuron models, IF, LIF, PLIF, and CLIF, along with their variants enhanced by self-prediction mechanisms.}
\label{pic:seq}
\end{center}
\vskip -0.2in
\end{figure}


Subsequently, comprehensive considering training cost, we evaluate our method on the larger-scale datasets ImageNet-100 with SEW ResNet34 model and ImageNet-1K with SEW ResNet18 model to further demonstrate its generalization capability in complex scenarios. 
The results are summarized in Table~\ref{tab:ablation_classification_imagenet}, it clearly shows that the self-prediction enhanced LIF and PLIF neurons consistently outperforms the original neuron across all settings.

\subsection{Ablation on Sequence Classification Tasks}

\begin{table}[htb]
\centering
\caption{Ablation on Sequential CIFAR10 Classification Tasks.}
\label{tab:ablation_seq_classification}
\begin{center}
\begin{small}
\begin{tabular}{ccccccc}
\toprule
Neuron/Acc. & base & self-pred(detach) & self-pred\\
\midrule
IF&71.74&69.08$\color{blue}{\downarrow}$&74.86$\color{red}{\uparrow}$\\
\cmidrule(l){1-4}
LIF&80.11&81.43$\color{red}{\uparrow}$&82.38$\color{red}{\uparrow}$\\
\cmidrule(l){1-4}
PLIF&82.54&83.03$\color{red}{\uparrow}$&83.52$\color{red}{\uparrow}$\\
\cmidrule(l){1-4}
CLIF&79.55&81.08$\color{red}{\uparrow}$&82.32$\color{red}{\uparrow}$\\
\bottomrule
\end{tabular}
\end{small}
\end{center}
\vskip -0.2in
\end{table}

We believe our method is particularly effective for tasks involving long-term temporal dependencies. To verify this, we evaluate it on the Sequential CIFAR-10 task. In this setting, each time step feeds the network with one column of the input image, resulting in a total of  $ T = 32 $ time-steps that matches the image width. As shown in Figure~\ref{pic:seq}, we train models using IF, LIF, PLIF, and CLIF neurons for 256 epochs and plot their training and test accuracy curves. Our self-prediction enhancement consistently yields noticeable improvements throughout the entire training process. Detailed best accuracy results are reported in Table~\ref{tab:ablation_seq_classification}.

Furthermore, we conduct an ablation study on detaching the gradient of the spike signal $ \bm{s}^l[t] $ in the prediction error term  $ \bm{x}^{l-1}[t] - \bm{s}^l[t]/\tau $ when calculating $\bm{m}^l_p[t]$ . The results show that preserving the full computational graph across $\bm{m}^l_p[t]$ leads to better performance, which validates the rationale behind our gradient propagation design discussed in Section~\ref{sec:gradient}.

\subsection{Ablation on Reinforcement Learning Tasks}
\begin{figure}[htb]
\begin{center}
\centerline{\includegraphics[width=1.0\columnwidth, trim=0.0cm 0.0cm 0.0cm 0.0cm, clip]{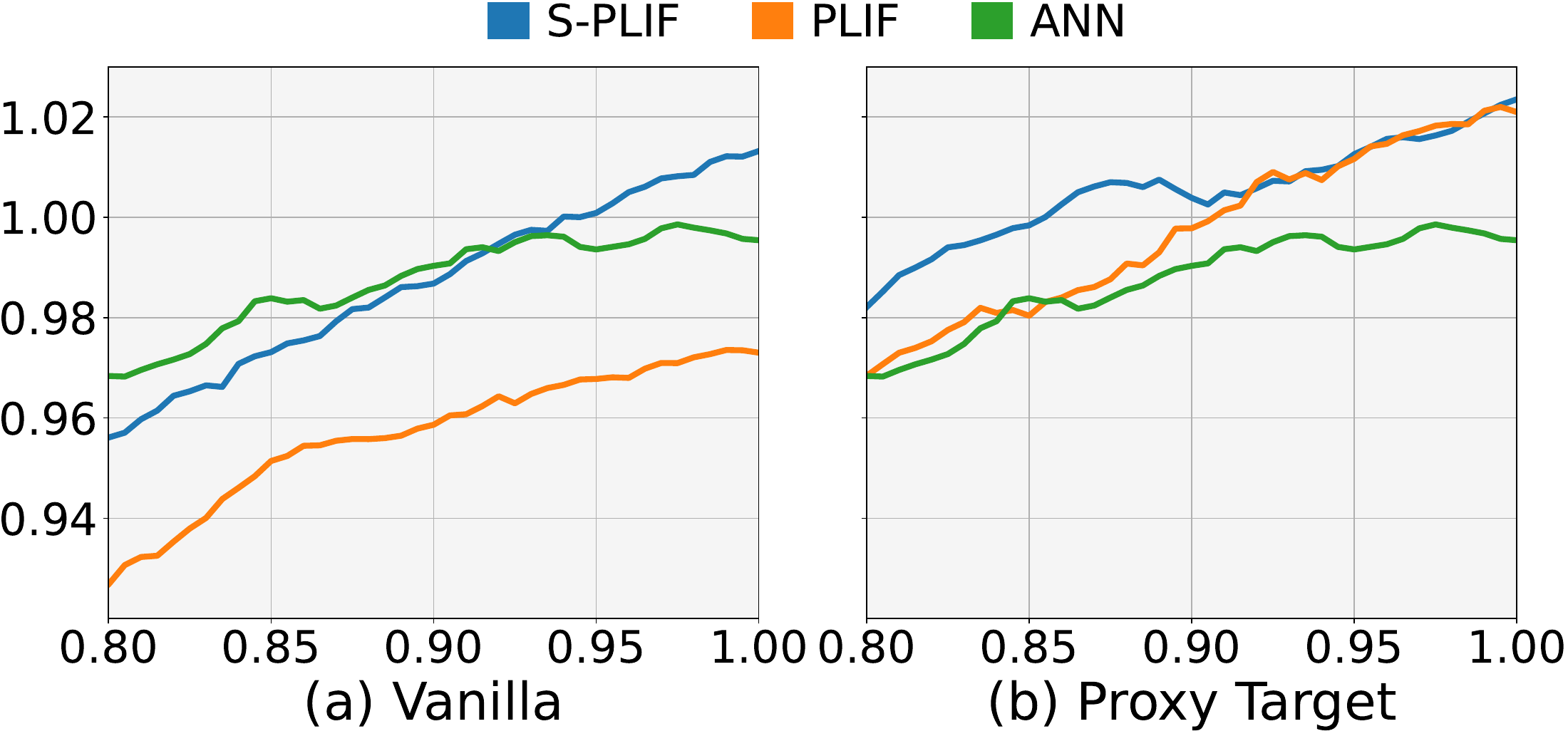}}
\caption{Normalized learning curves across all environments of the TD3 algorithm with different spiking neurons across all environments. The performance and training steps are normalized linearly based on ANN performance. Curves are uniformly smoothed for visual clarity.}
\label{pic:reinforcement_result}
\end{center}
\vskip -0.2in
\end{figure}
Moreeover, we conduct experiment on four changing reinforcement learning tasks from the MuJoCo \cite{mujoco1,mujoco2} continuous control benchmarks including Ant-v4, HalfCheetah-v4, Hopper-v4, and Walker2d-v4. We combined PLIF and self-prediction enchanced PLIF(S-PLIF) neuron with the vanilla hybrid frameworks of the spiking actor networks \cite{popSAN}, and the SOTA proxy target (PT) framework \cite{xu2025proxy}. All experiments result are reimplemented results using the RL algorithm of Twin-delayed Deep Determination Policy Gradient (TD3) \cite{TD3} and we choose the hyper-parameters as suggested in \cite{xu2025proxy}. 

Figure~\ref{pic:reinforcement_result} shows the normalized learning curves across all 4 environments, and Table~\ref{tab:ablation_reinforcement} reports the expected returns for each environment. We also report the average performance gain, defined as the average percentile improvement over ANNs, following the metric in \citet{xu2025care}. Our self-prediction enhancement consistently improves SNN performance in reinforcement learning (RL) tasks, even when integrated with state-of-the-art SNN-RL approaches such as proxy target frameworks. Notably, with this enhancement, directly trained SNNs can achieve and sometimes even surpass ANN performances in RL tasks.

\begin{table}[t]
\centering
\caption{Max average returns over $5$ random seeds with PLIF and S-PLIF spiking neurons, and the average performance gain against ANN baseline, where $\pm$ denotes one standard deviation.}
\label{tab:ablation_reinforcement}
\begin{center}
\begin{small}
\begin{tabular}{cccc}
\toprule
Environment & Frame & Neuron & Performance\\
\midrule
\multirow{6}{*}{Ant-v4}&\multirow{1}{*}{ANN}& ReLU & $4932 \pm 1287$\\
\cmidrule(l){2-4}
&\multirow{2}{*}{Vanilla}&PLIF&$4912 \pm 1009$\\
&&S-PLIF&$4981 \pm 607$\\
\cmidrule(l){2-4}
&\multirow{2}{*}{PT}&PLIF&$5259 \pm 326$\\
&&S-PLIF&$5527 \pm 151$\\
\cmidrule(l){1-4}
\multirow{6}{*}{HalfCheetah-v4}&\multirow{1}{*}{ANN}& ReLU & $10554 \pm 561$\\
\cmidrule(l){2-4}
&\multirow{2}{*}{Vanilla}&PLIF&$9252 \pm 520$\\
&&S-PLIF&$9828 \pm 594$\\
\cmidrule(l){2-4}
&\multirow{2}{*}{PT}&PLIF&$9219 \pm 216$\\
&&S-PLIF&$9405 \pm 597$\\
\cmidrule(l){1-4}
\multirow{6}{*}{Hopper-v4}&\multirow{1}{*}{ANN}& ReLU & $3349 \pm 151$\\
\cmidrule(l){2-4}
&\multirow{2}{*}{Vanilla}&PLIF&$3414 \pm 122$\\
&&S-PLIF&$3462 \pm 128$\\
\cmidrule(l){2-4}
&\multirow{2}{*}{PT}&PLIF&$3384 \pm 100$\\
&&S-PLIF&$3380 \pm 105$\\
\cmidrule(l){1-4}
\multirow{6}{*}{Walker2d-v4}&\multirow{1}{*}{ANN}& ReLU & $4050 \pm 323$\\
\cmidrule(l){2-4}
&\multirow{2}{*}{Vanilla}&PLIF&$4003 \pm 274$\\
&&S-PLIF&$4271 \pm 485$\\
\cmidrule(l){2-4}
&\multirow{2}{*}{PT}&PLIF&$4445 \pm 217$\\
&&S-PLIF&$4497 \pm 391$\\
\cmidrule(l){1-4}
\multirow{6}{*}{Overall}&\multirow{1}{*}{ANN}& ReLU & $0.00\%$\\
\cmidrule(l){2-4}
&\multirow{2}{*}{Vanilla}&PLIF&$-2.99\%$\\
&&S-PLIF&$0.74\%$\\
\cmidrule(l){2-4}
&\multirow{2}{*}{PT}&PLIF&$1.20\%$\\
&&S-PLIF&$3.28\%$\\
\bottomrule
\end{tabular}
\end{small}
\end{center}
\vskip -0.1in
\end{table}

\section{Conclusion}
In this work, we proposed a self-prediction enhanced spiking neuron method, drawing inspiration from the brain’s predictive coding mechanisms. By enabling each neuron to generate an internal prediction current based on its own input-output history, our approach offers a simple yet effective way to modulate membrane potential dynamics. This design not only creates a continuous gradient path that mitigates vanishing gradients and significantly boosts training accuracy across diverse SNN architectures, neuron models, and tasks, but also aligns with biological principles of dendritic modulation and error-driven synaptic plasticity. Our comprehensive experiments on image classification, sequential classification, and reinforcement learning tasks consistently demonstrate the broad applicability and robustness of the proposed method, establishing it as a promising general enhancement for Spiking Neural Networks.

\clearpage
\section*{Impact Statement}
This paper presents work whose goal is to advance the field of Machine Learning. There are many potential societal consequences of our work, none which we feel must be specifically highlighted here.

\bibliography{example_paper}
\bibliographystyle{icml2026}

\newpage
\appendix
\onecolumn
\section{Results on CIFAR10 Dataset}
\subsection{Results on CIFAR10Net}
\begin{figure}[htb]
\begin{center}
\vskip -0.15in
\centerline{\includegraphics[width=0.25\columnwidth, trim=0.0cm 0.0cm 0.0cm 0.0cm, clip]{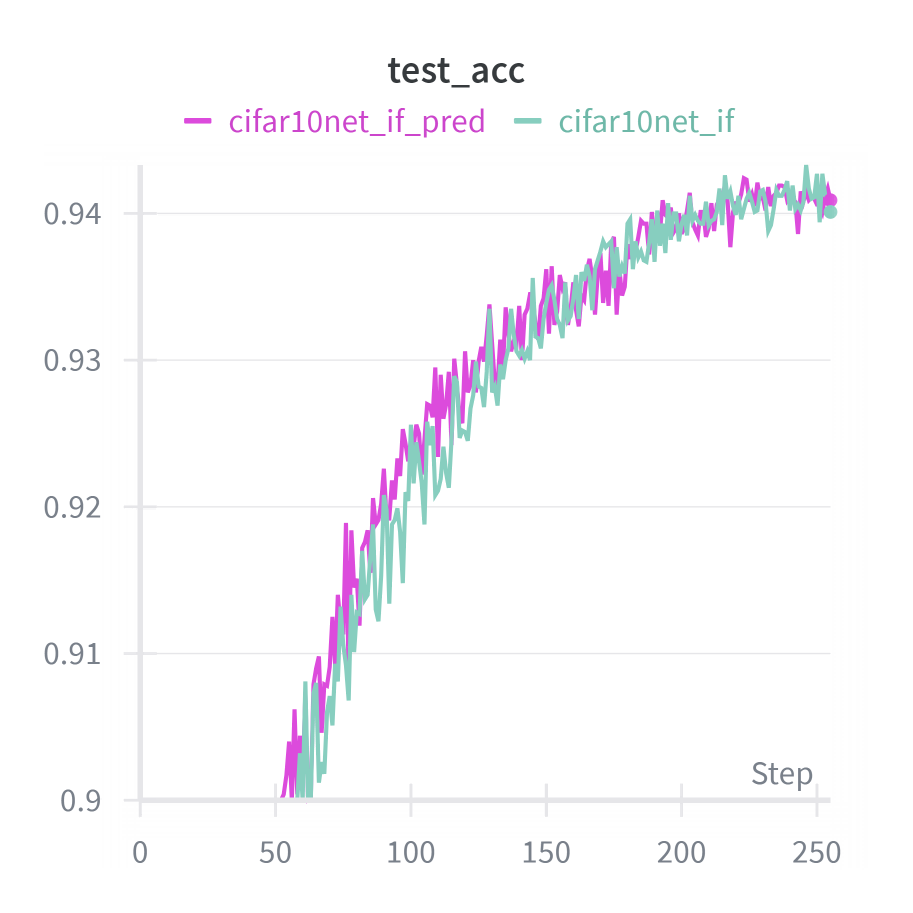}\includegraphics[width=0.25\columnwidth, trim=0.0cm 0.0cm 0.0cm 0.0cm, clip]{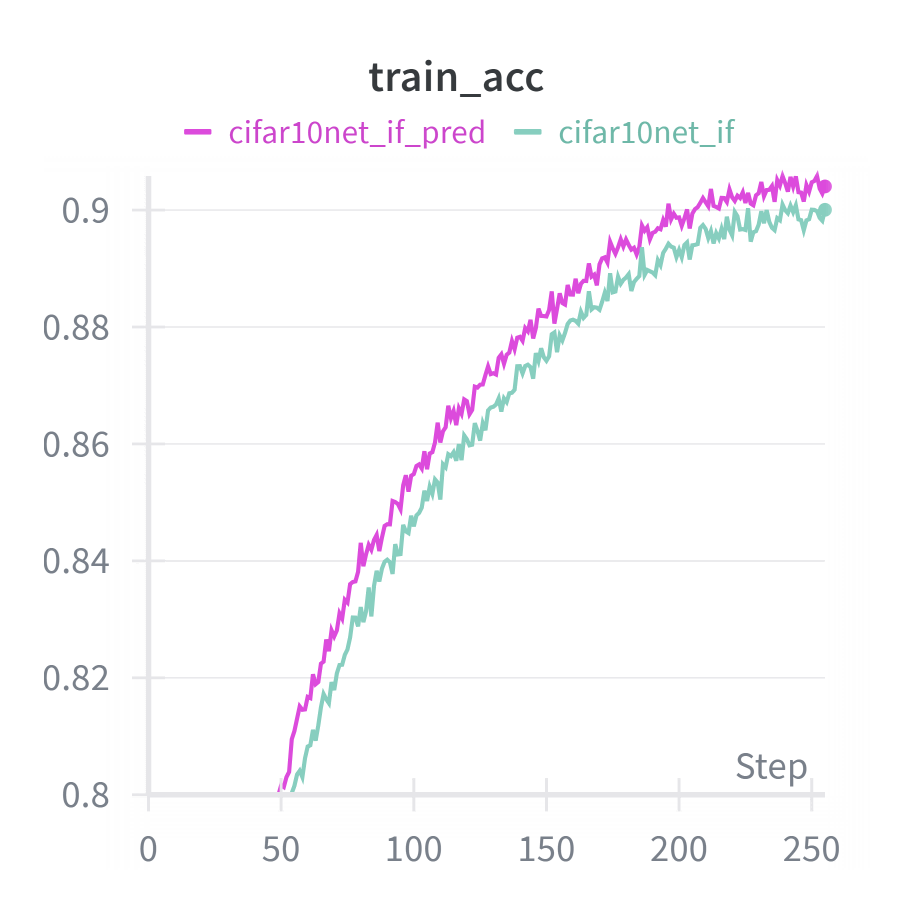}
\includegraphics[width=0.25\columnwidth, trim=0.0cm 0.0cm 0.0cm 0.0cm, clip]{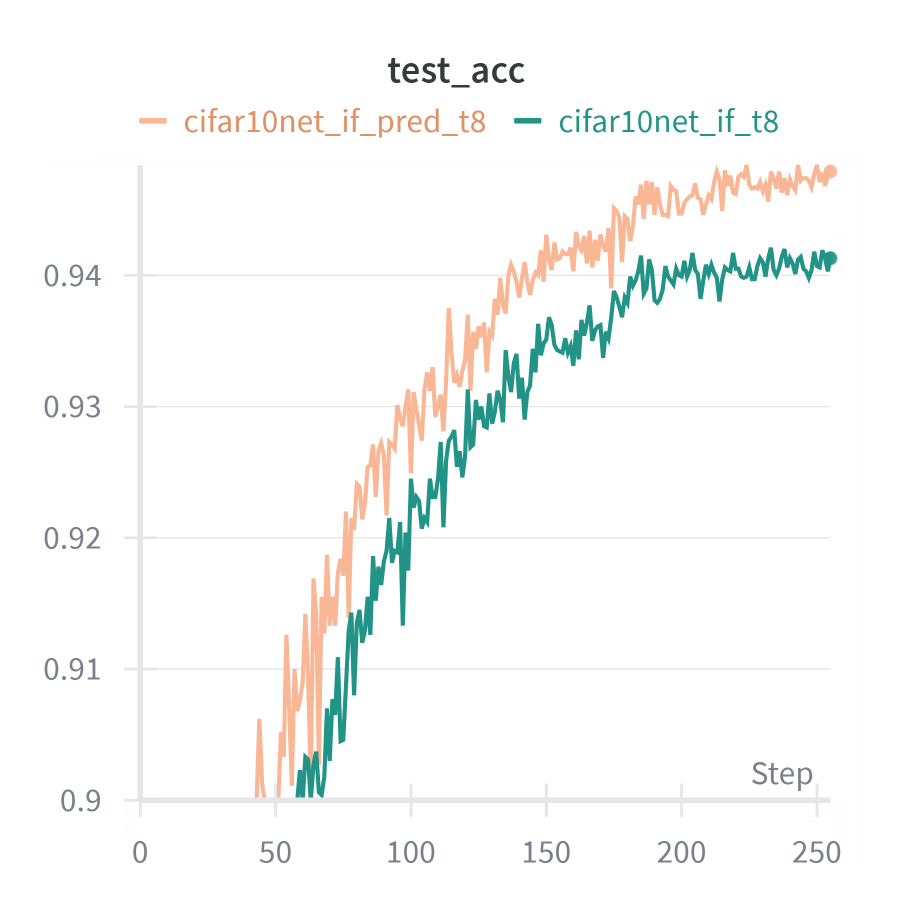}\includegraphics[width=0.25\columnwidth, trim=0.0cm 0.0cm 0.0cm 0.0cm, clip]{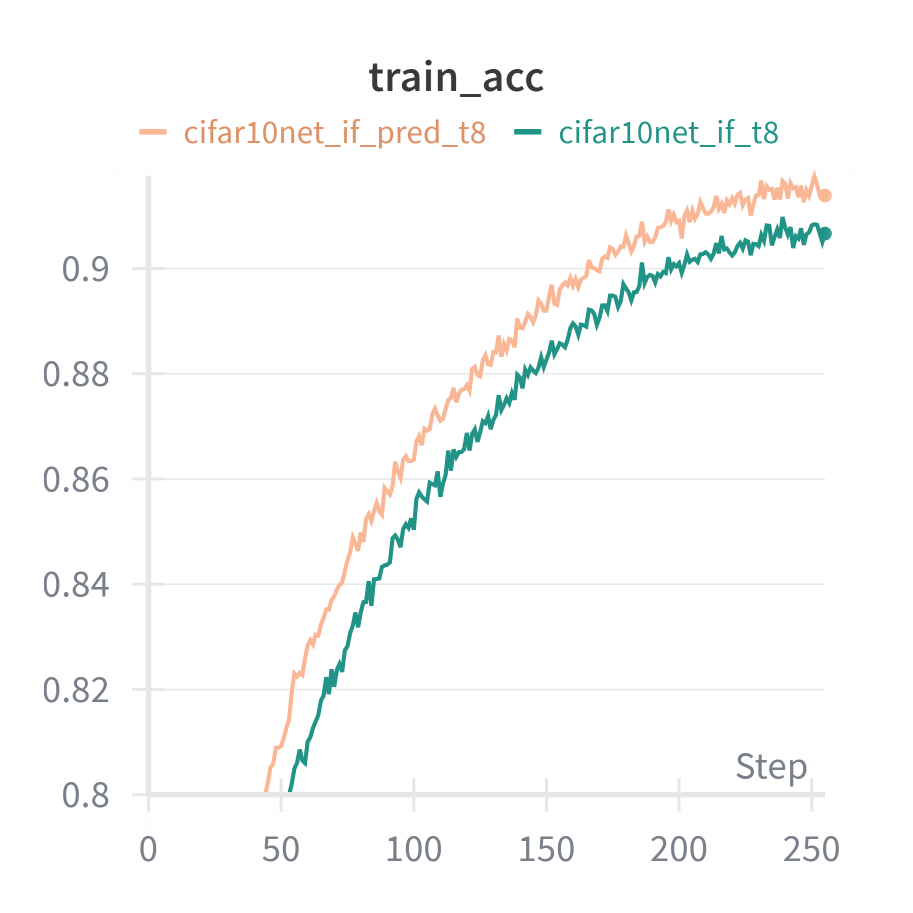}
}
\vskip -0.05in
\centerline{\includegraphics[width=0.25\columnwidth, trim=0.0cm 0.0cm 0.0cm 0.0cm, clip]{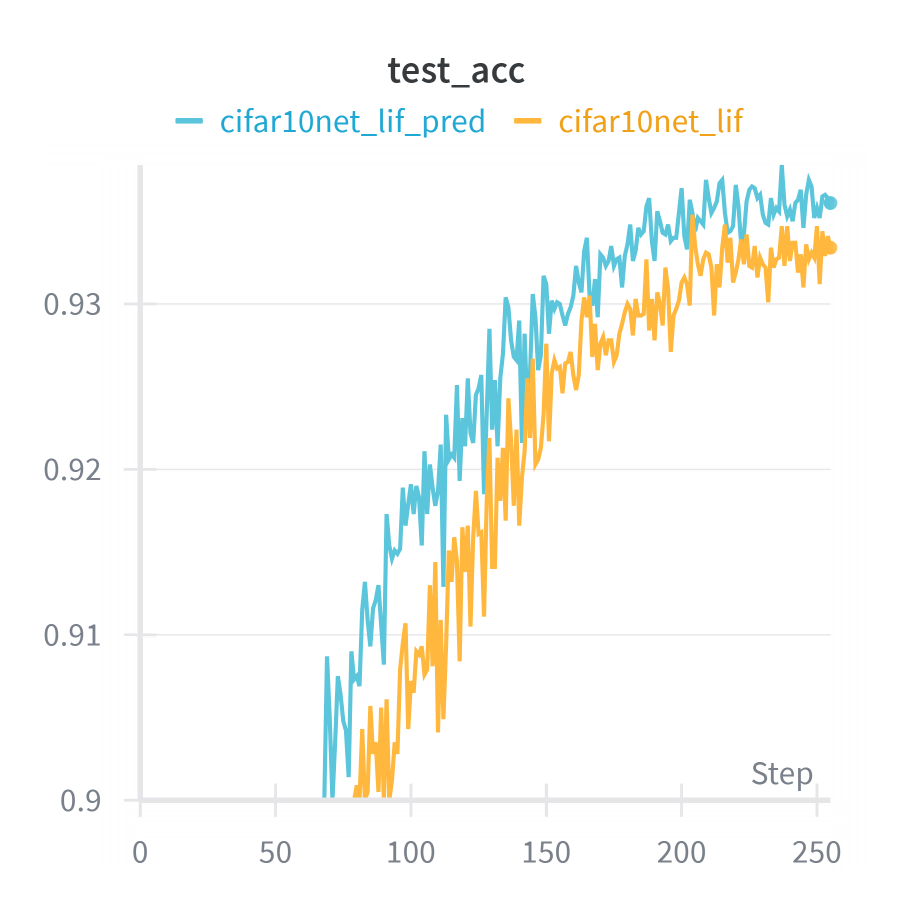}\includegraphics[width=0.25\columnwidth, trim=0.0cm 0.0cm 0.0cm 0.0cm, clip]{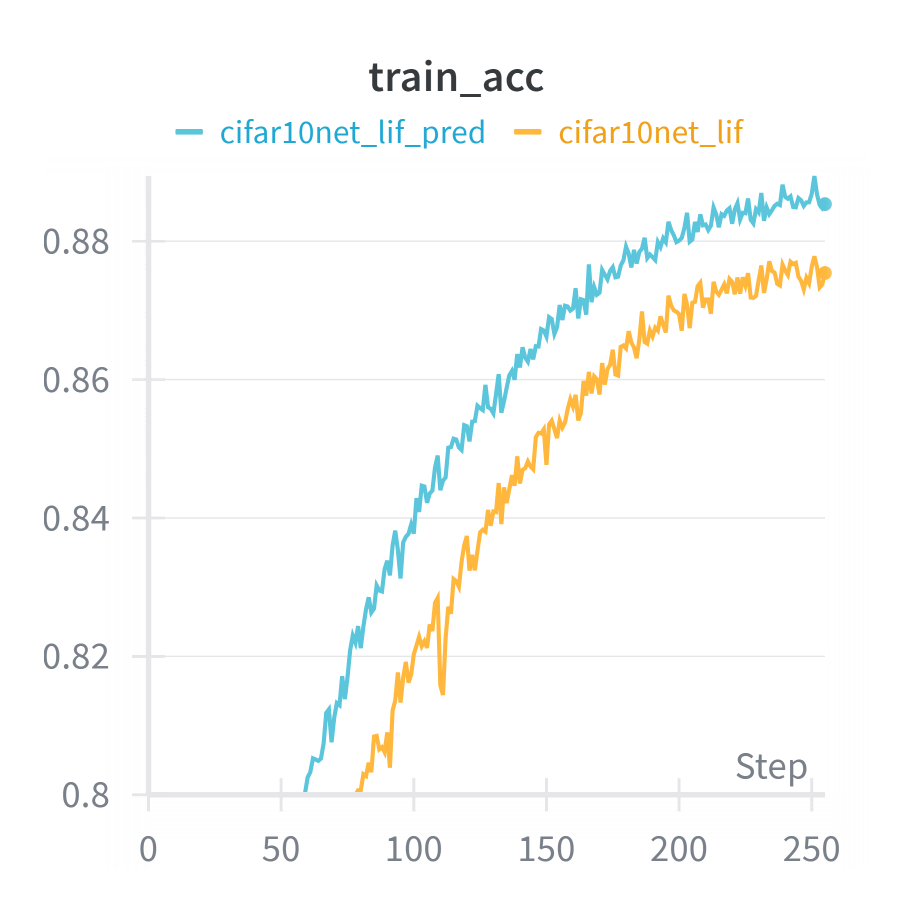}
\includegraphics[width=0.25\columnwidth, trim=0.0cm 0.0cm 0.0cm 0.0cm, clip]{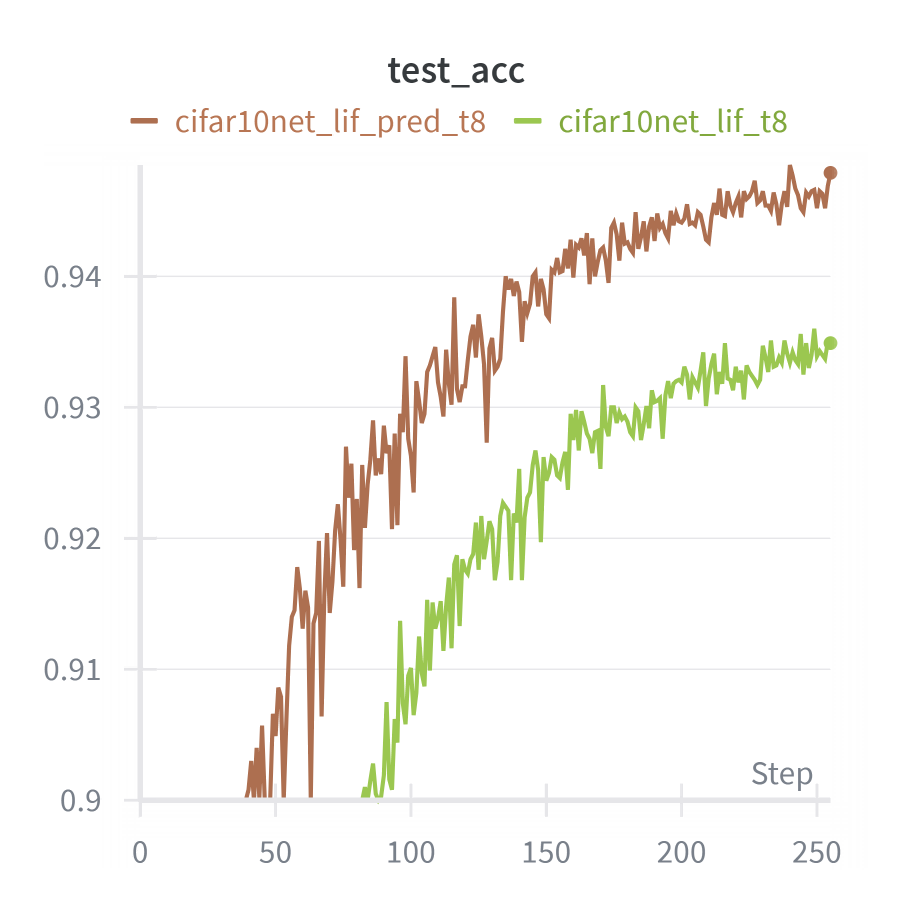}\includegraphics[width=0.25\columnwidth, trim=0.0cm 0.0cm 0.0cm 0.0cm, clip]{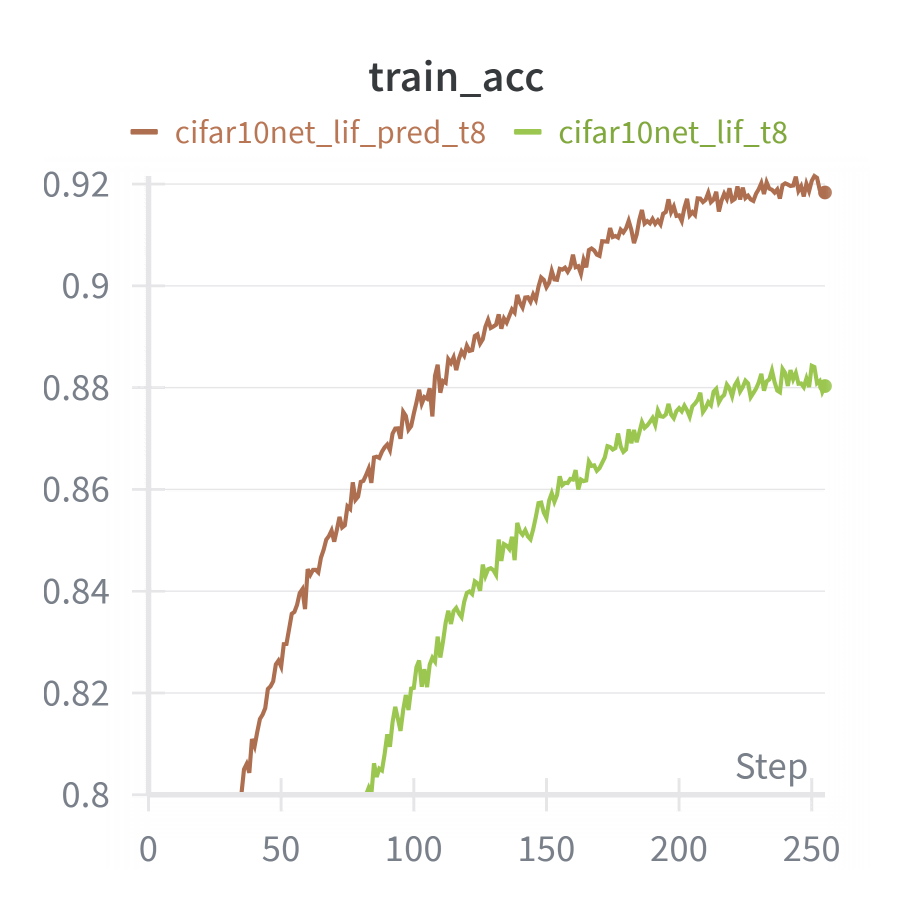}
}
\vskip -0.05in
\centerline{\includegraphics[width=0.25\columnwidth, trim=0.0cm 0.0cm 0.0cm 0.0cm, clip]{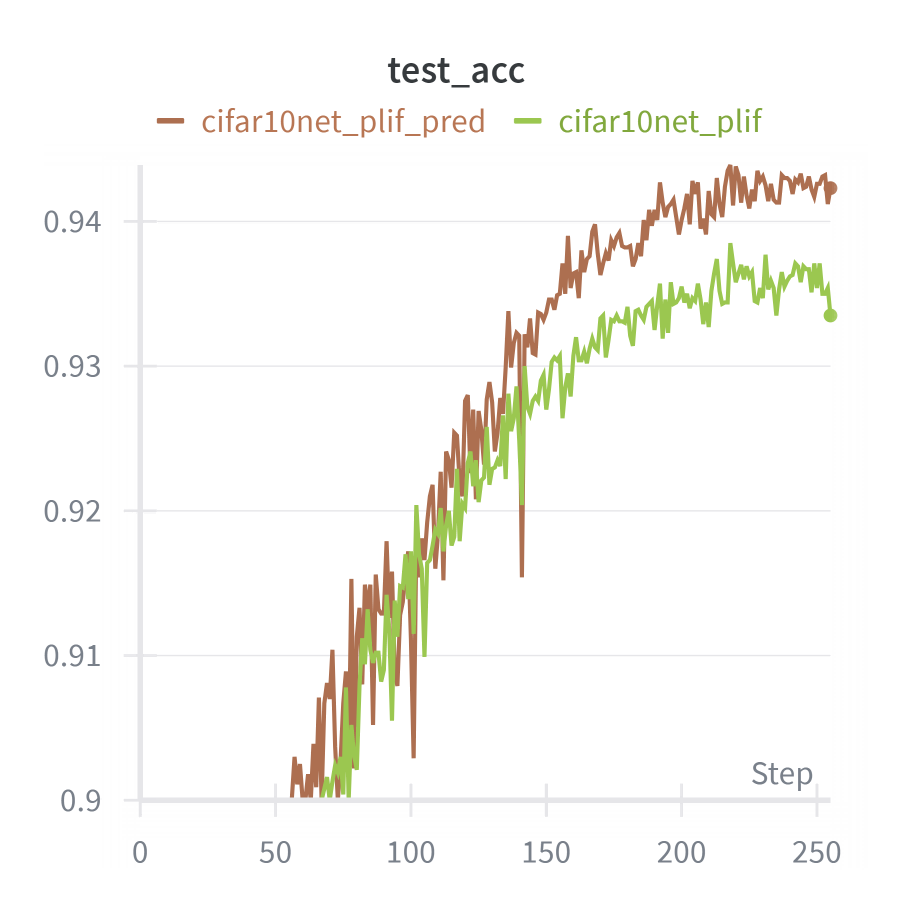}\includegraphics[width=0.25\columnwidth, trim=0.0cm 0.0cm 0.0cm 0.0cm, clip]{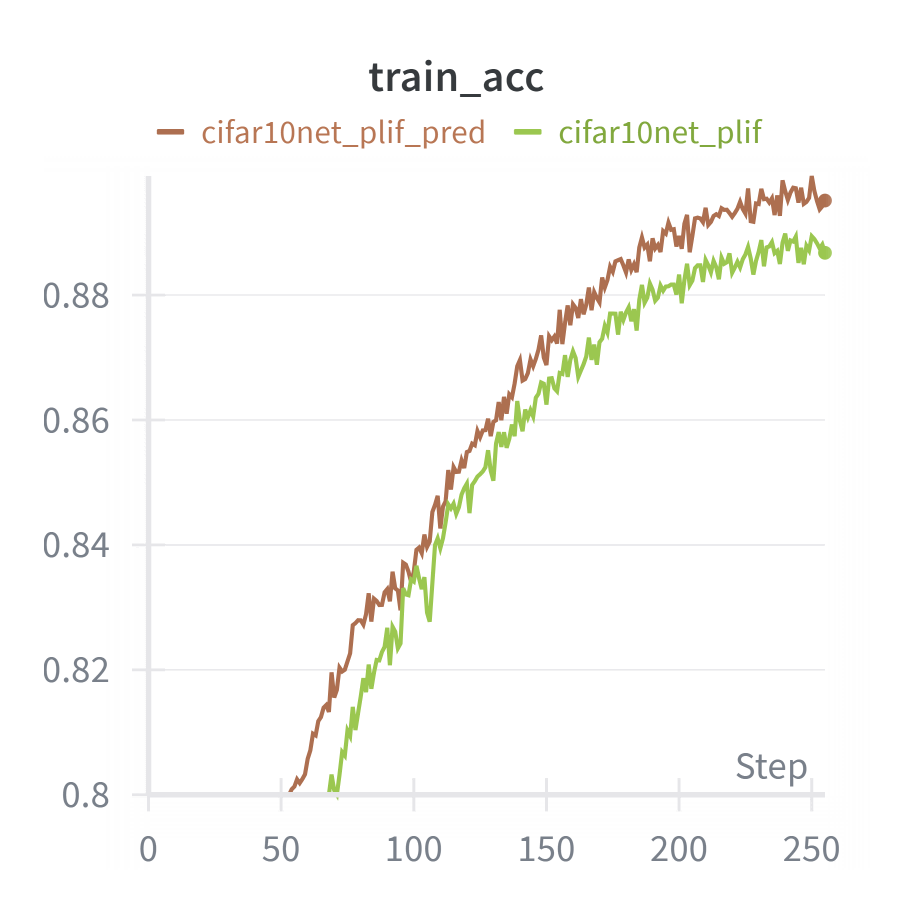}
\includegraphics[width=0.25\columnwidth, trim=0.0cm 0.0cm 0.0cm 0.0cm, clip]{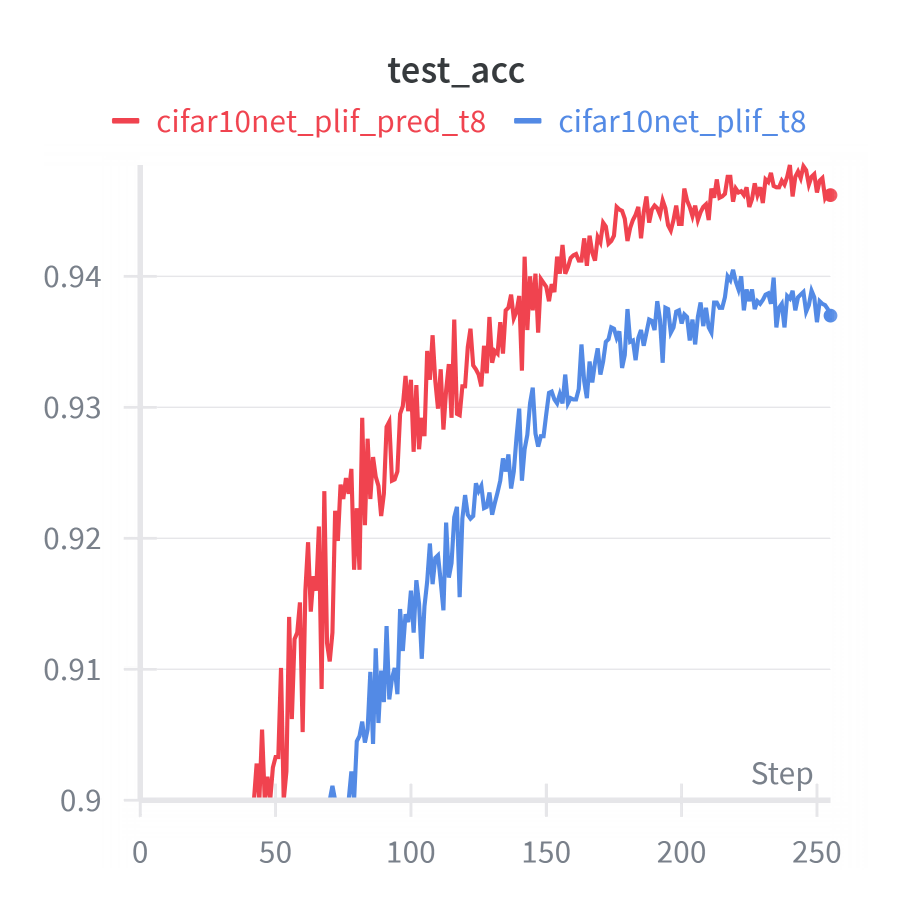}\includegraphics[width=0.25\columnwidth, trim=0.0cm 0.0cm 0.0cm 0.0cm, clip]{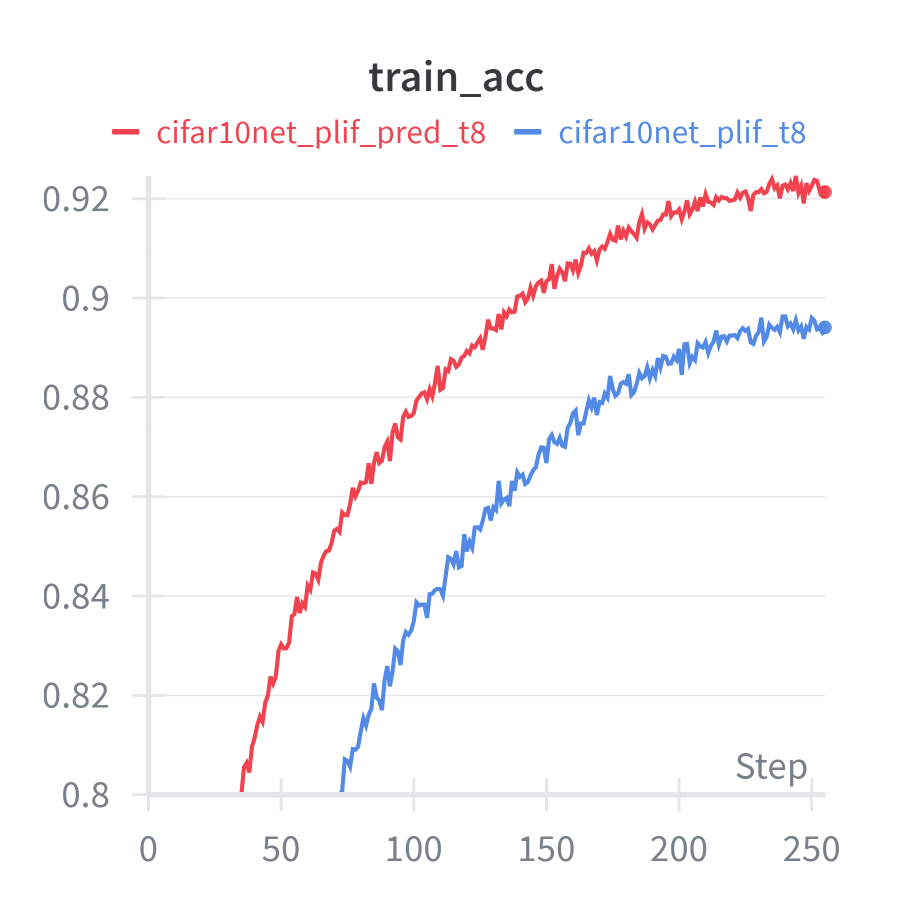}
}
\vskip -0.05in
\centerline{\includegraphics[width=0.25\columnwidth, trim=0.0cm 0.0cm 0.0cm 0.0cm, clip]{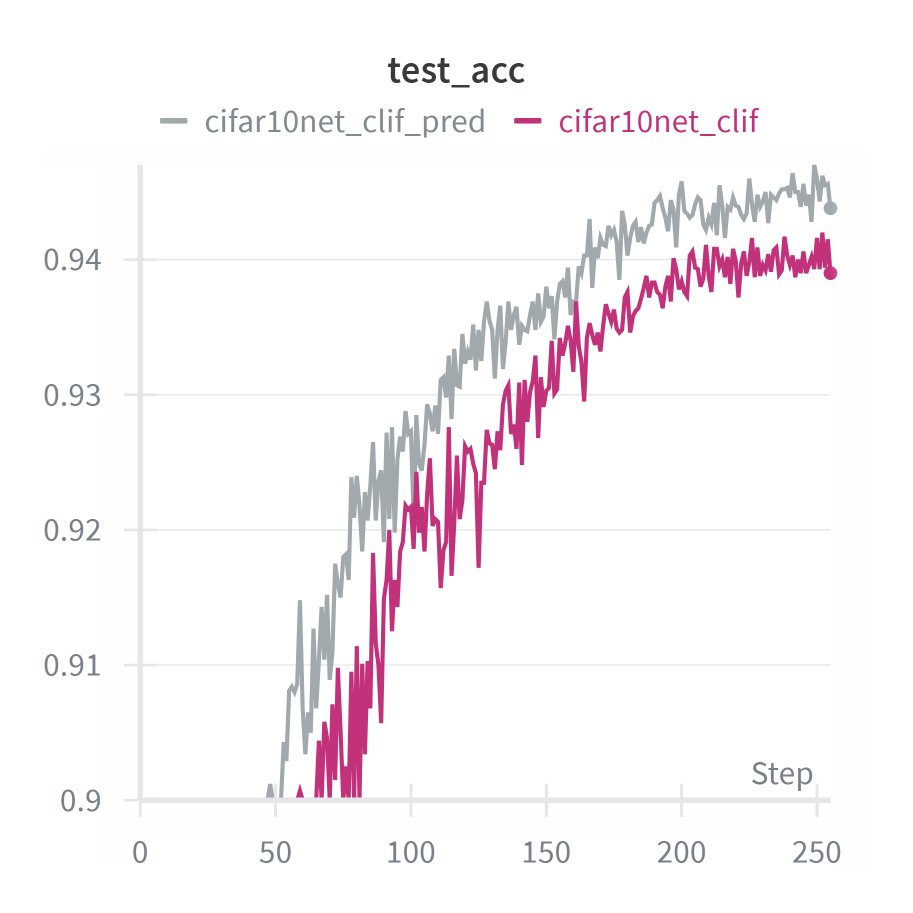}\includegraphics[width=0.25\columnwidth, trim=0.0cm 0.0cm 0.0cm 0.0cm, clip]{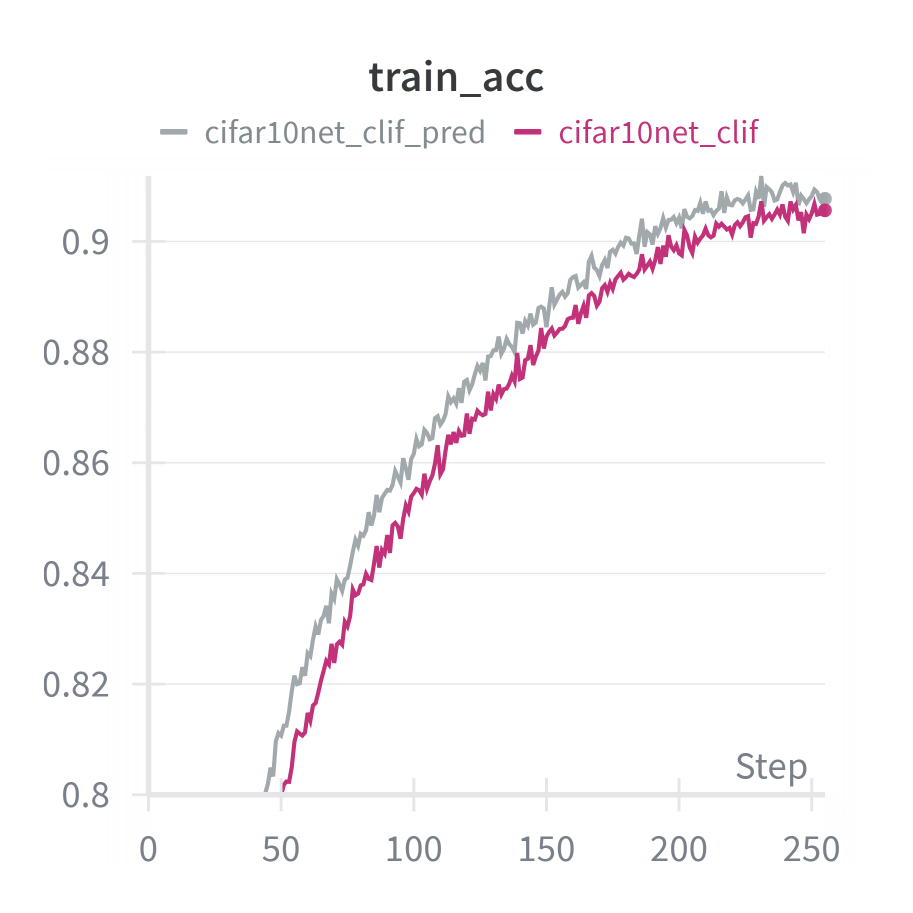}
\includegraphics[width=0.25\columnwidth, trim=0.0cm 0.0cm 0.0cm 0.0cm, clip]{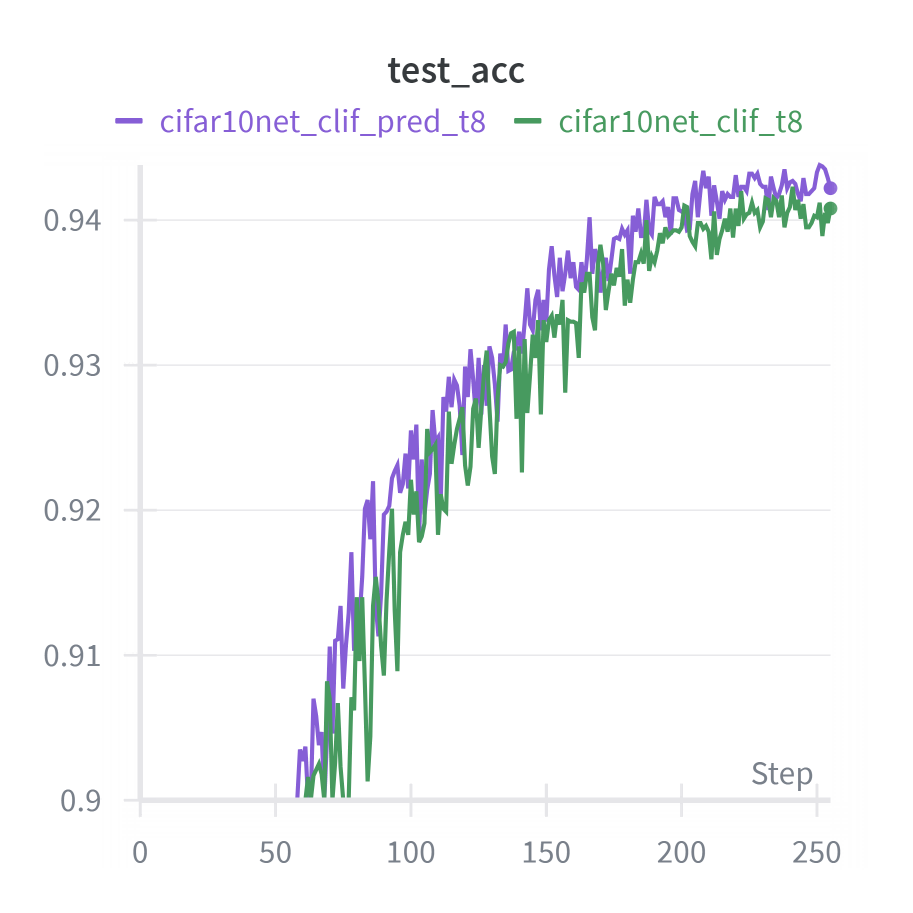}\includegraphics[width=0.25\columnwidth, trim=0.0cm 0.0cm 0.0cm 0.0cm, clip]{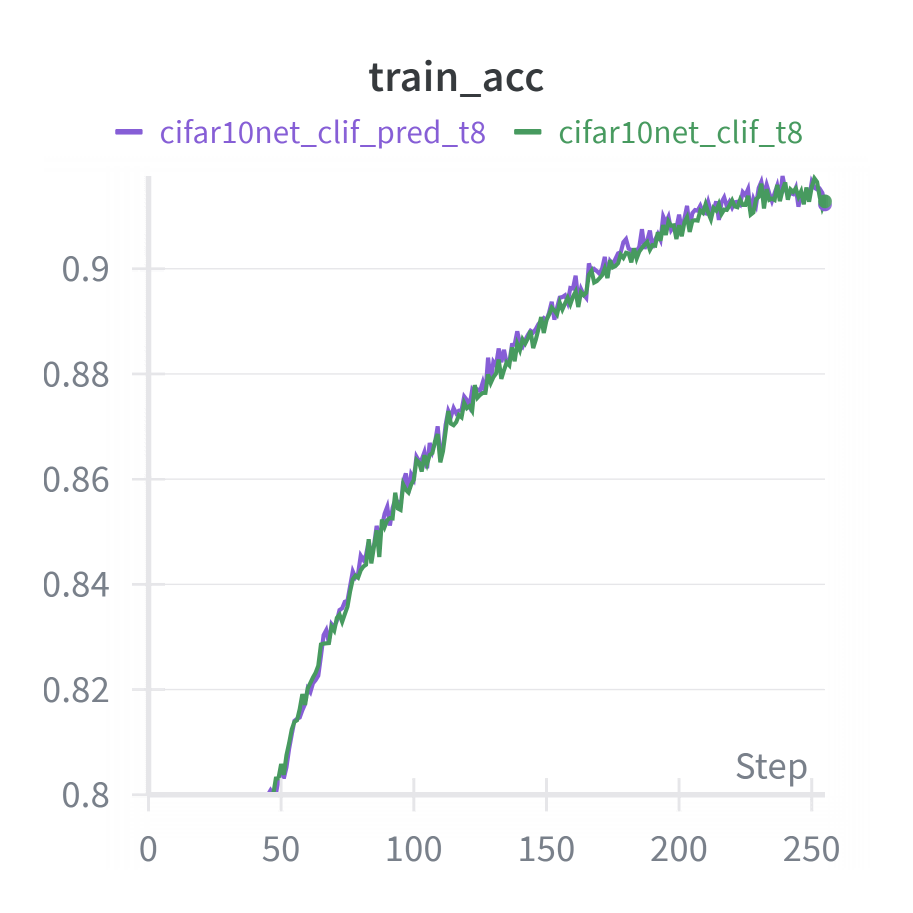}
}
\vskip -0.1in
\caption{The training and testing accuracy curves of the CIFAR10Net trained with IF,LIF,PLIF and CLIF neurons along with their variants enhanced by self-prediction mechanisms at time-steps T=4 and T=8.}
\label{pic:cifar10net}
\end{center}
\end{figure}

This Cifar10Net spiking neural network architecture consists of the following components in sequence: an initial convolutional block (Conv2d(3, 128, kernel size=3, padding=1) + BatchNorm2d(128) + spiking neuron), followed by MaxPool2d(2,2); then two convolutional blocks—first Conv2d(128, 256, 3×3) and then Conv2d(256, 256, 3×3), each equipped with batch normalization and a spiking neuron, followed by another MaxPool2d(2,2); next, three identical deeper convolutional blocks (Conv2d(256, 512, 3×3) + BN + neuron); after flattening the feature maps, the data passes through two fully connected layers, the first being Linear(512×8×8, 512) with a neuron, and the second Linear(512, 10) with a neuron, to produce 10-class spiking outputs; finally, a global average pooling operation nn.AvgPool1d(10, 10) is applied along the temporal dimension to generate the final prediction. Training and testing curve in shown in Figure~\ref{pic:cifar10net}.

\subsection{Results on SEW ResNet}

\begin{figure}[h]
\begin{center}
\vskip -0.1in
\centerline{\includegraphics[width=0.25\columnwidth, trim=0.0cm 0.0cm 0.0cm 0.0cm, clip]{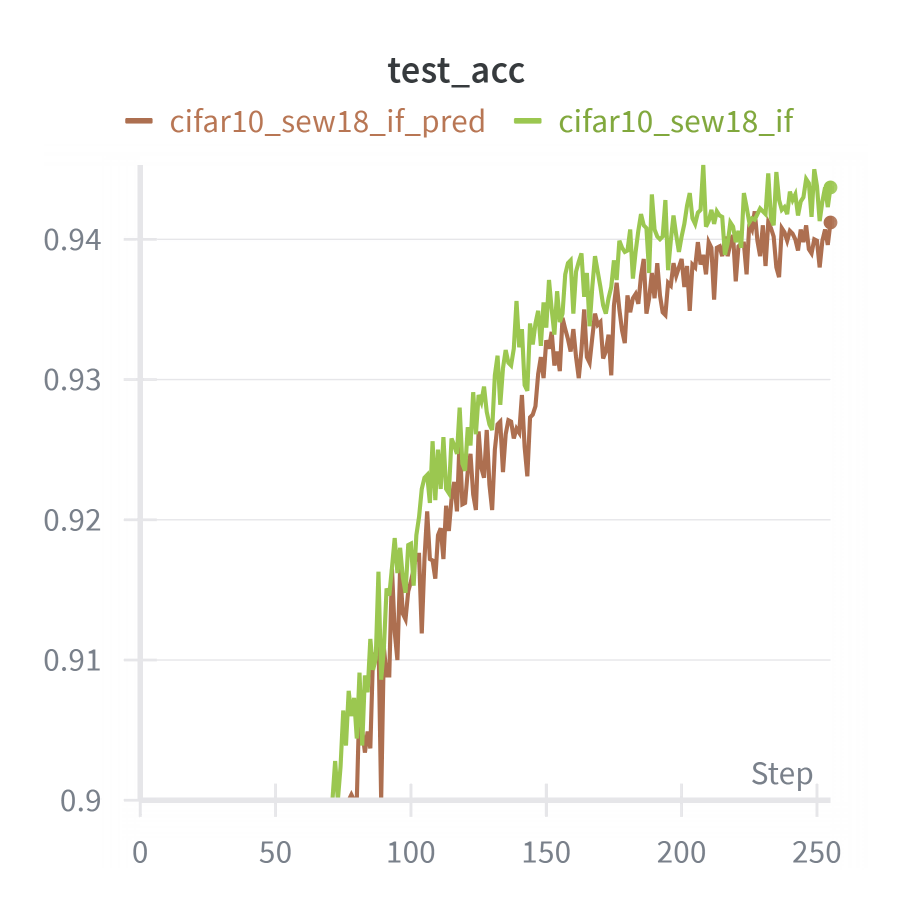}
\includegraphics[width=0.25\columnwidth, trim=0.0cm 0.0cm 0.0cm 0.0cm, clip]{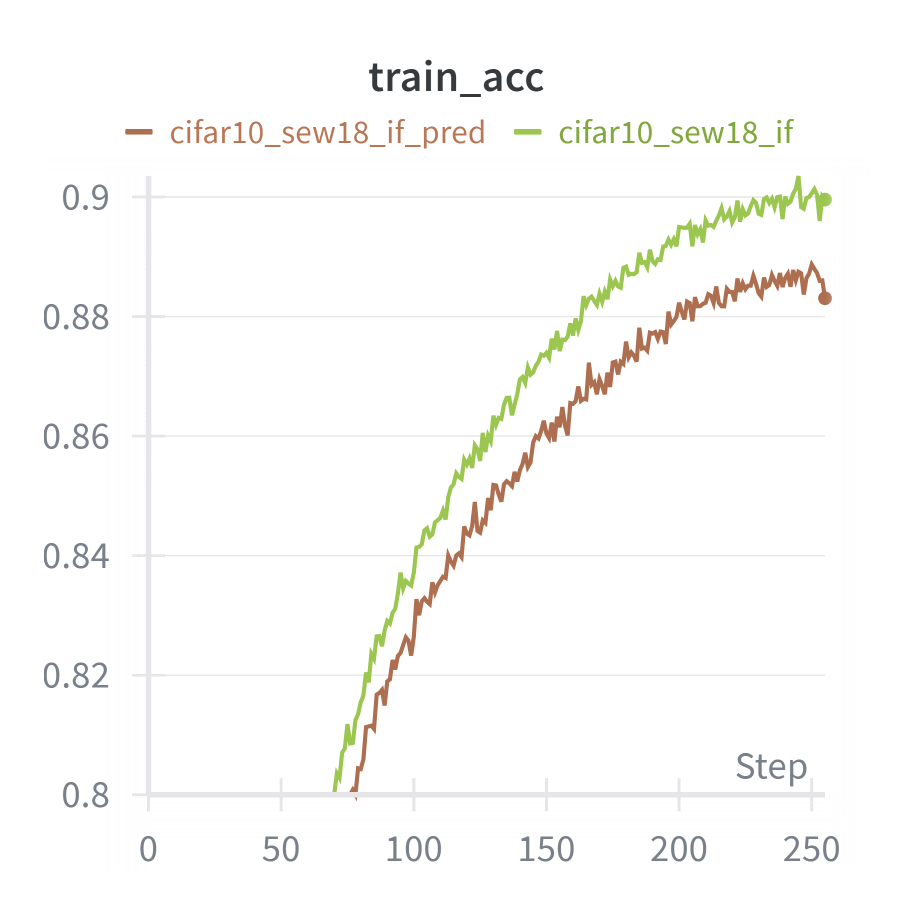}
\includegraphics[width=0.25\columnwidth, trim=0.0cm 0.0cm 0.0cm 0.0cm, clip]{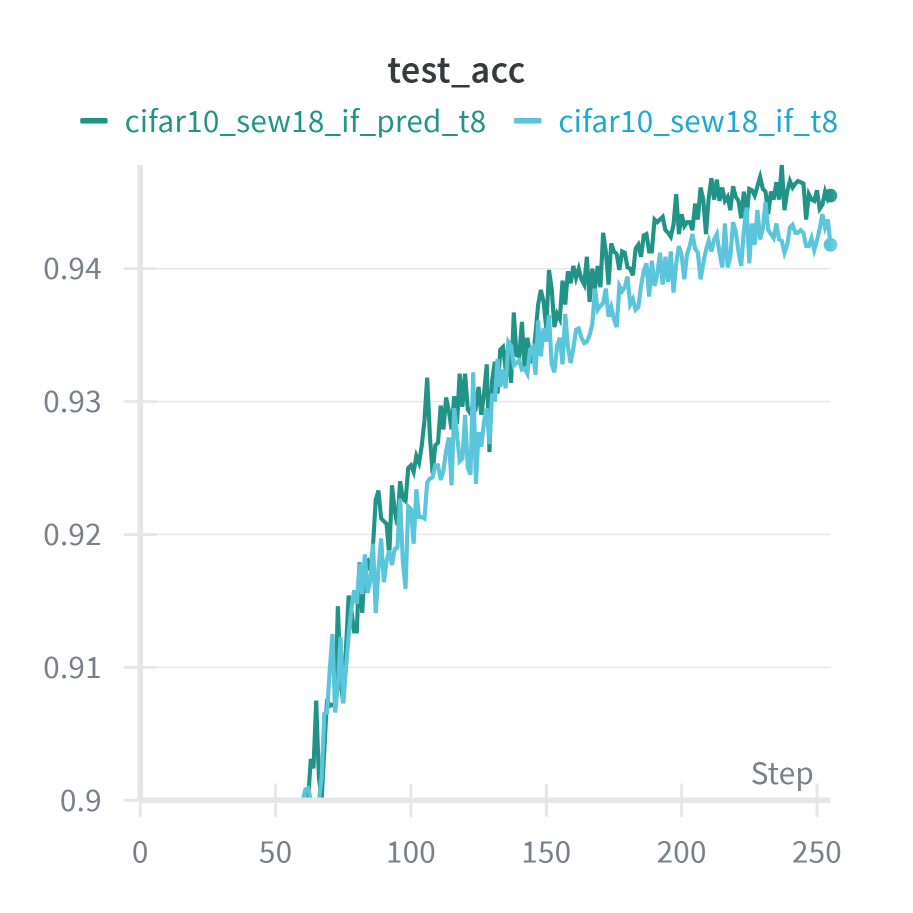}
\includegraphics[width=0.25\columnwidth, trim=0.0cm 0.0cm 0.0cm 0.0cm, clip]{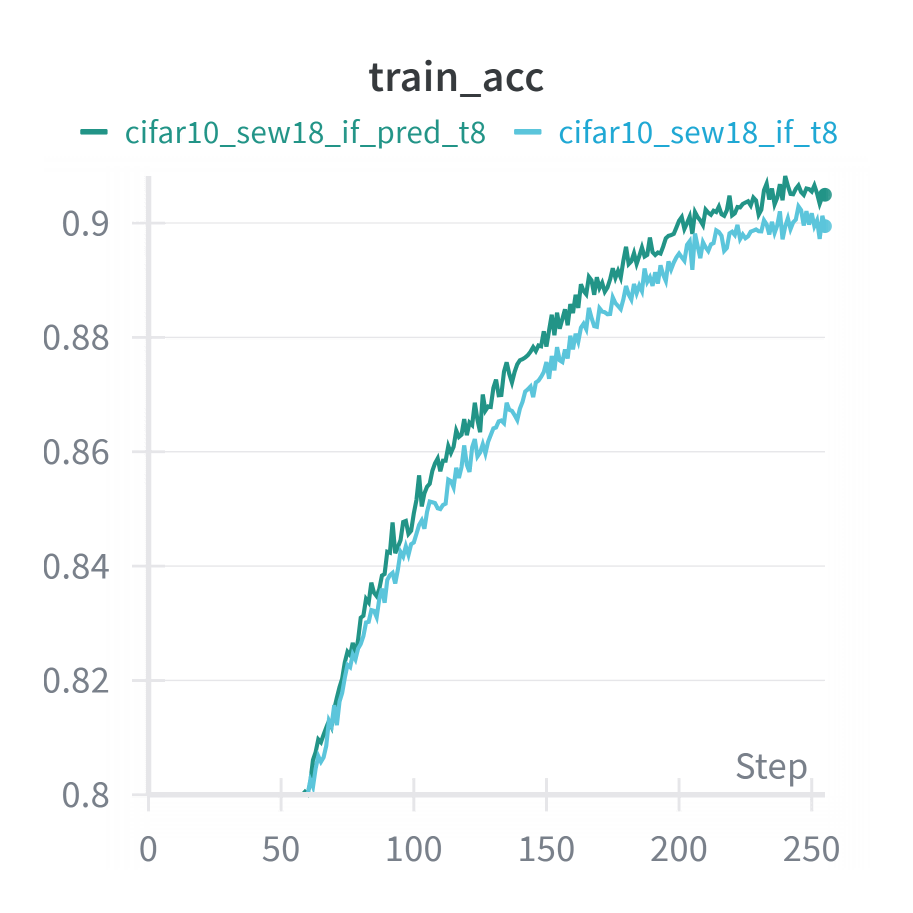}
}
\vskip -0.05in
\centerline{\includegraphics[width=0.25\columnwidth, trim=0.0cm 0.0cm 0.0cm 0.0cm, clip]{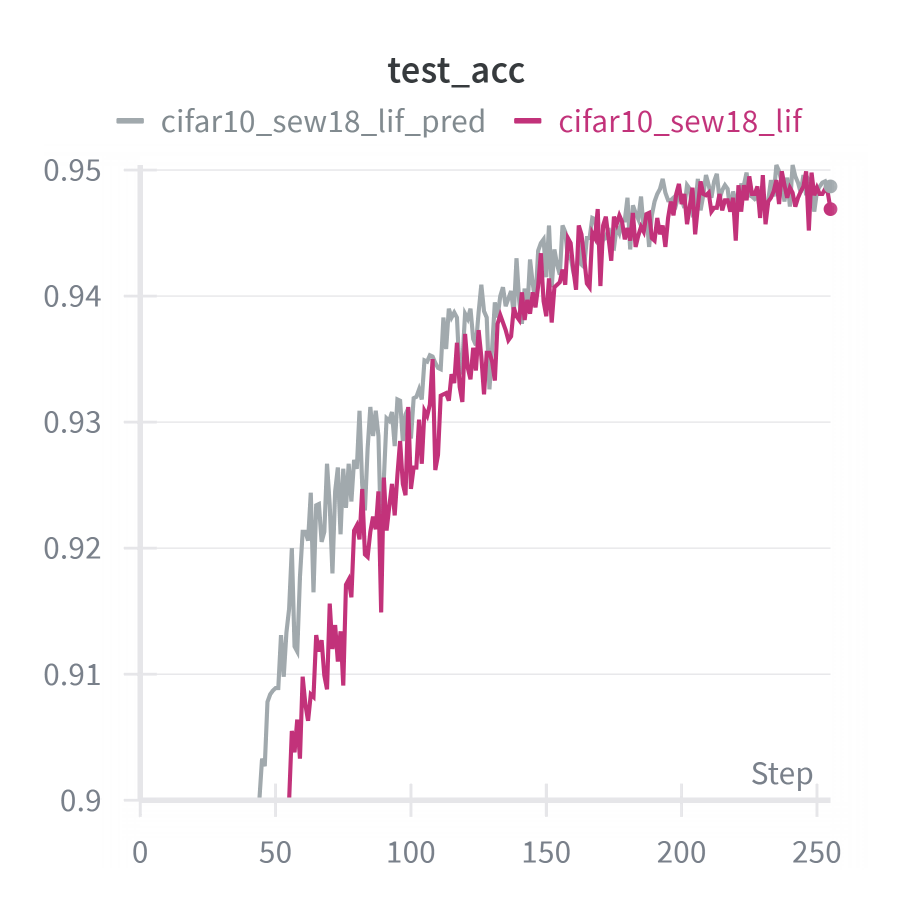}
\includegraphics[width=0.25\columnwidth, trim=0.0cm 0.0cm 0.0cm 0.0cm, clip]{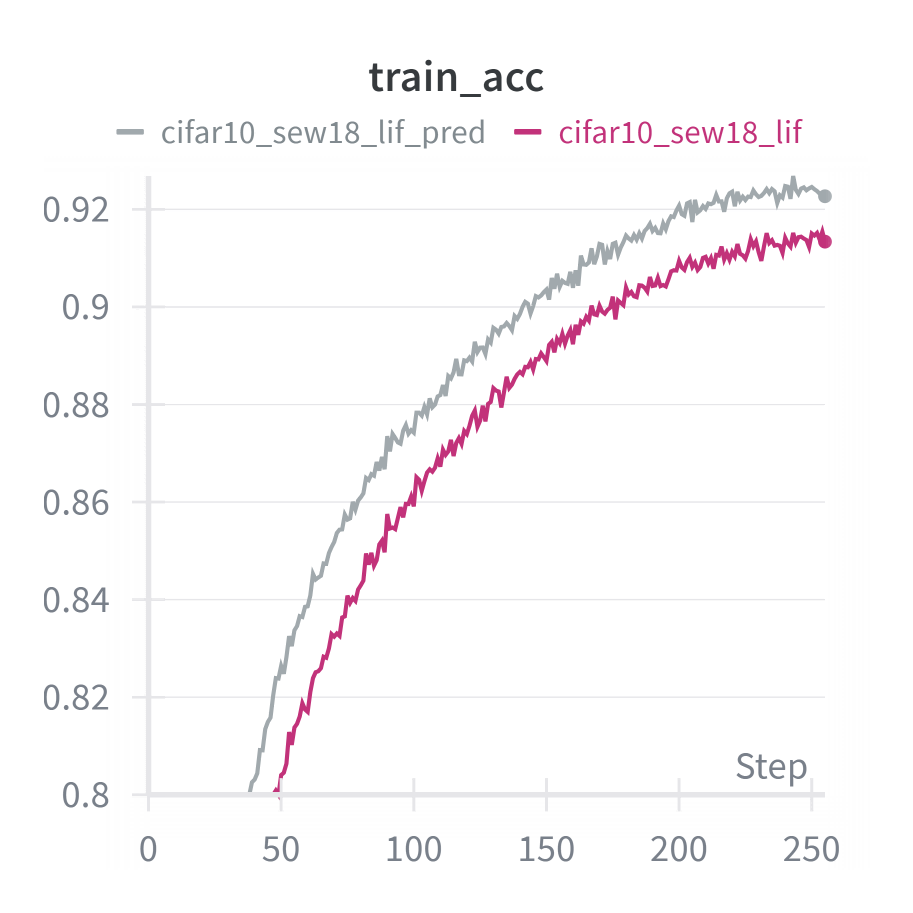}
\includegraphics[width=0.25\columnwidth, trim=0.0cm 0.0cm 0.0cm 0.0cm, clip]{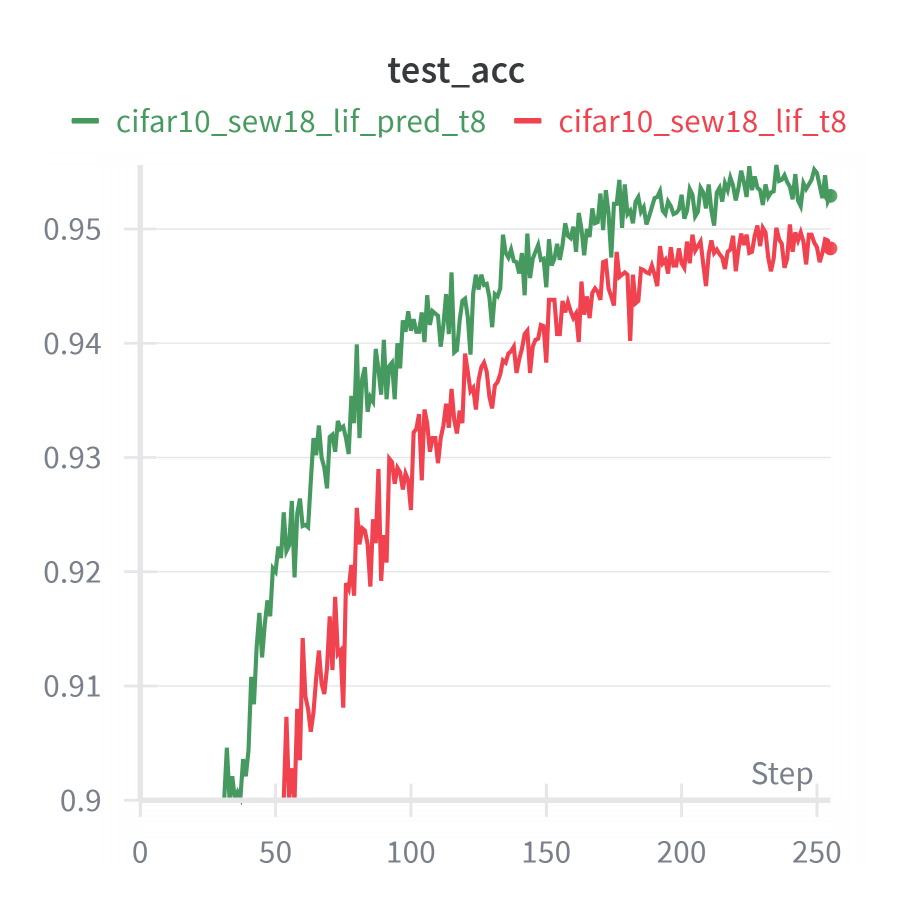}
\includegraphics[width=0.25\columnwidth, trim=0.0cm 0.0cm 0.0cm 0.0cm, clip]{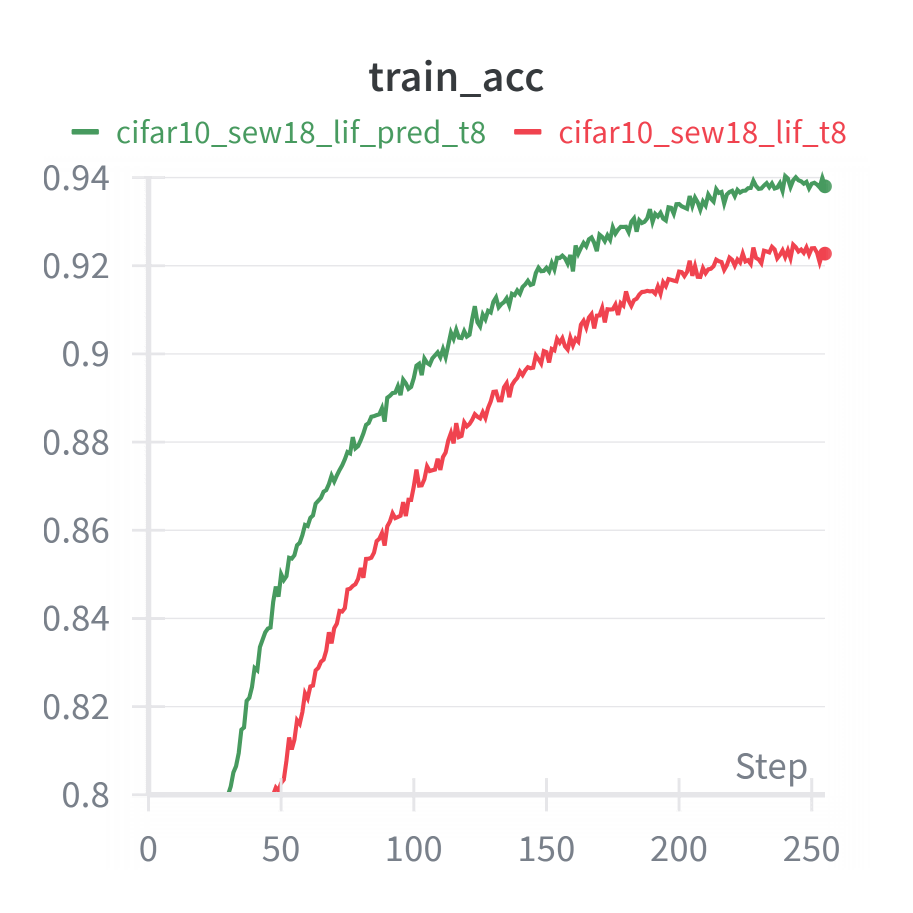}
}
\vskip -0.05in
\centerline{\includegraphics[width=0.25\columnwidth, trim=0.0cm 0.0cm 0.0cm 0.0cm, clip]{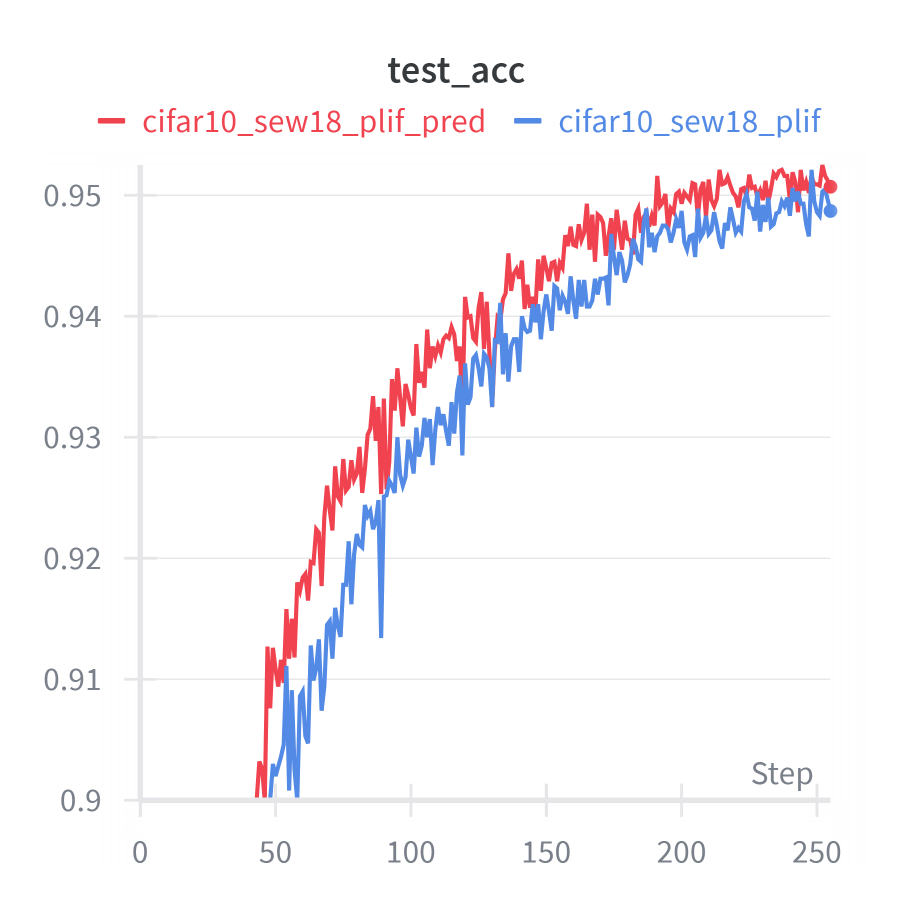}
\includegraphics[width=0.25\columnwidth, trim=0.0cm 0.0cm 0.0cm 0.0cm, clip]{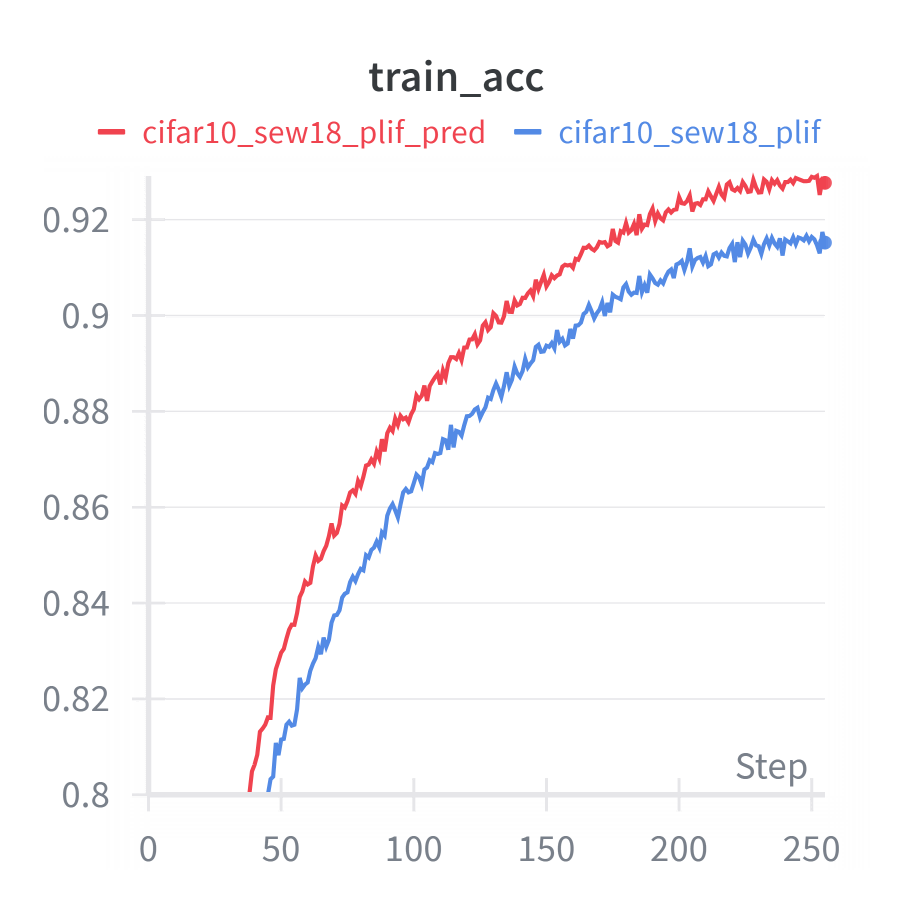}
\includegraphics[width=0.25\columnwidth, trim=0.0cm 0.0cm 0.0cm 0.0cm, clip]{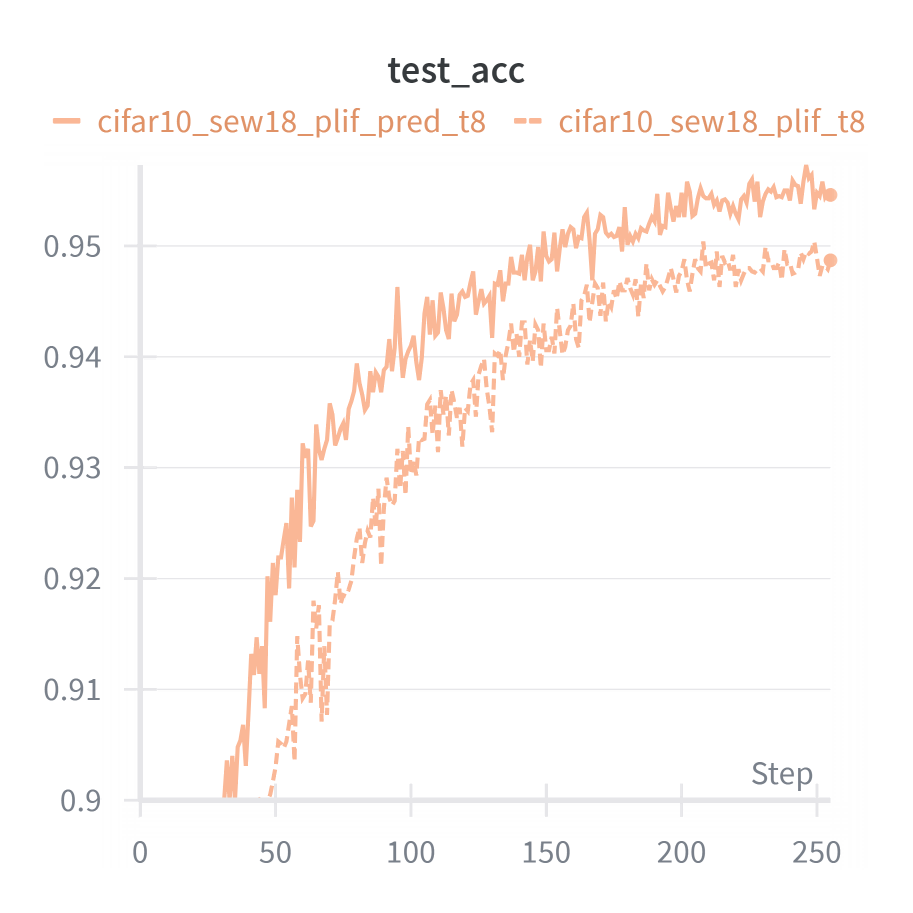}
\includegraphics[width=0.25\columnwidth, trim=0.0cm 0.0cm 0.0cm 0.0cm, clip]{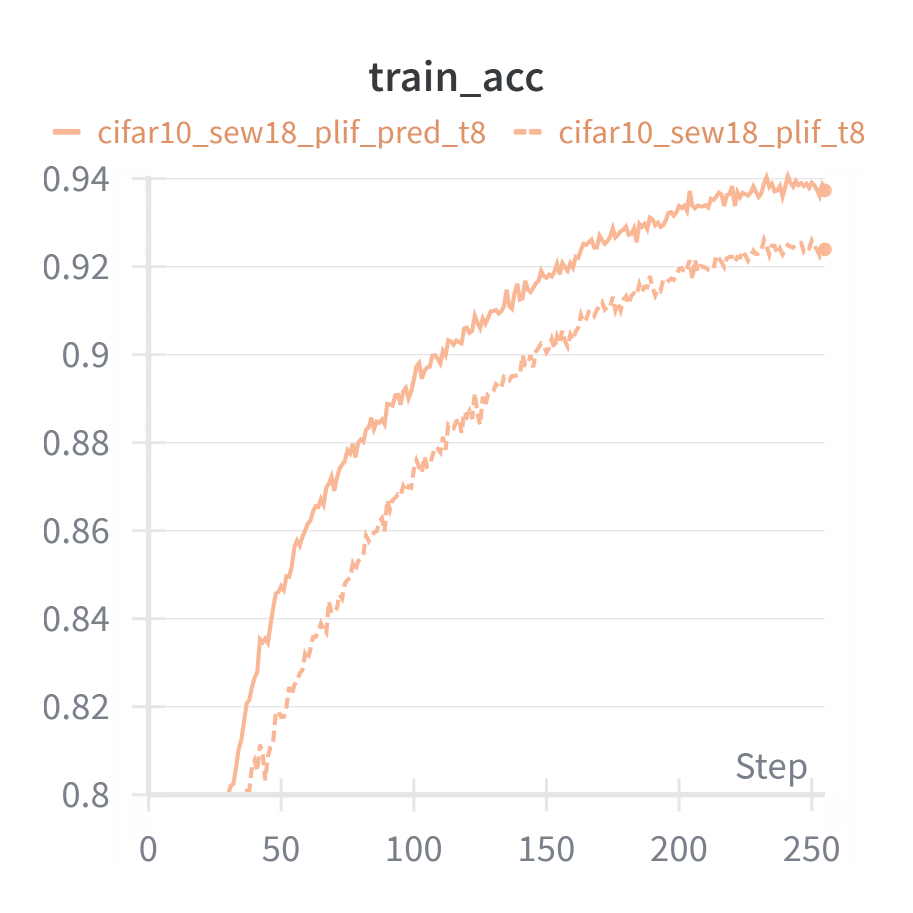}
}
\vskip -0.05in
\centerline{\includegraphics[width=0.25\columnwidth, trim=0.0cm 0.0cm 0.0cm 0.0cm, clip]{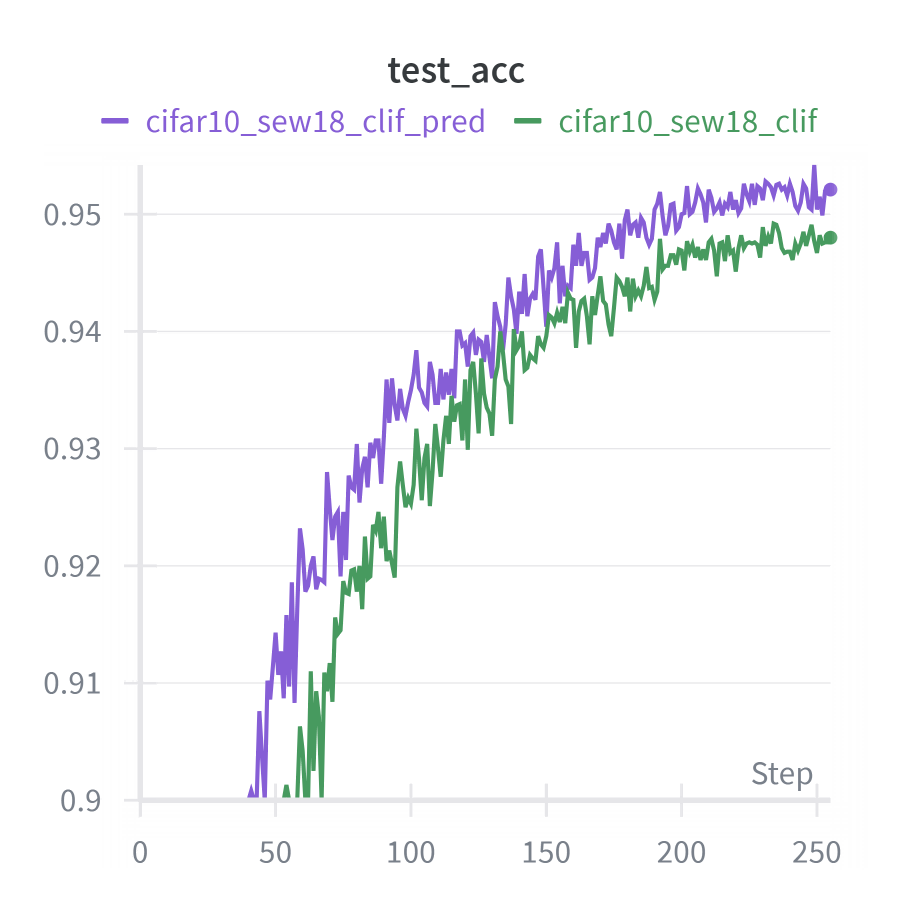}
\includegraphics[width=0.25\columnwidth, trim=0.0cm 0.0cm 0.0cm 0.0cm, clip]{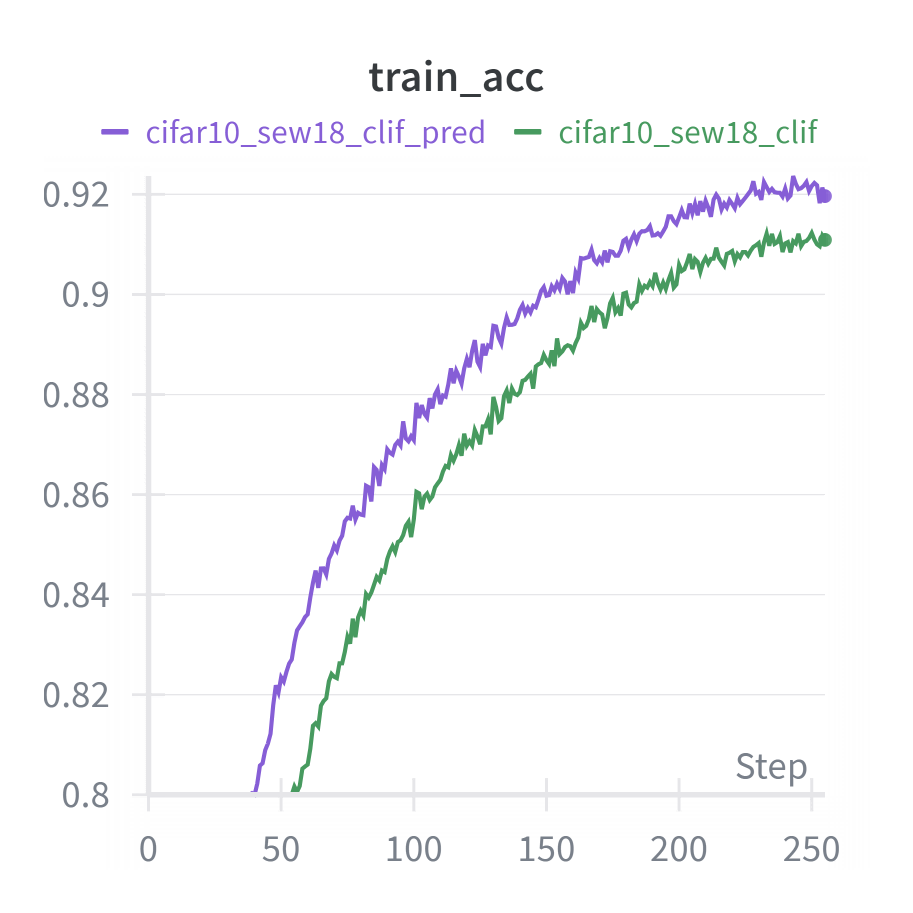}
\includegraphics[width=0.25\columnwidth, trim=0.0cm 0.0cm 0.0cm 0.0cm, clip]{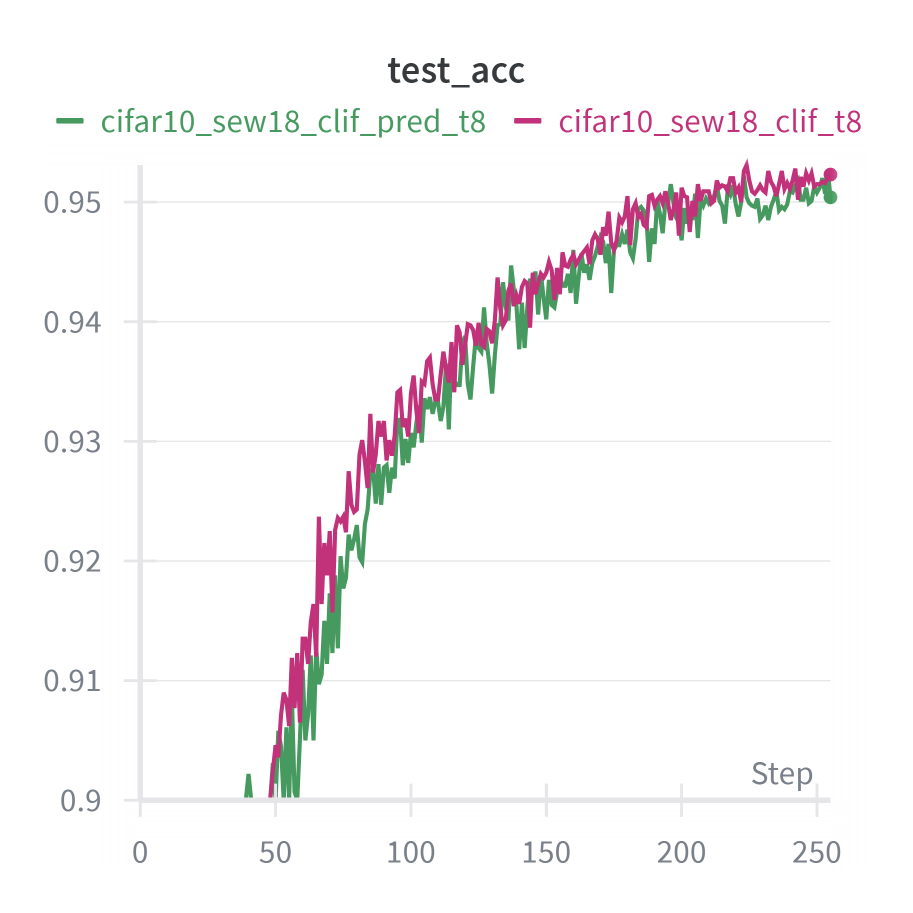}
\includegraphics[width=0.25\columnwidth, trim=0.0cm 0.0cm 0.0cm 0.0cm, clip]{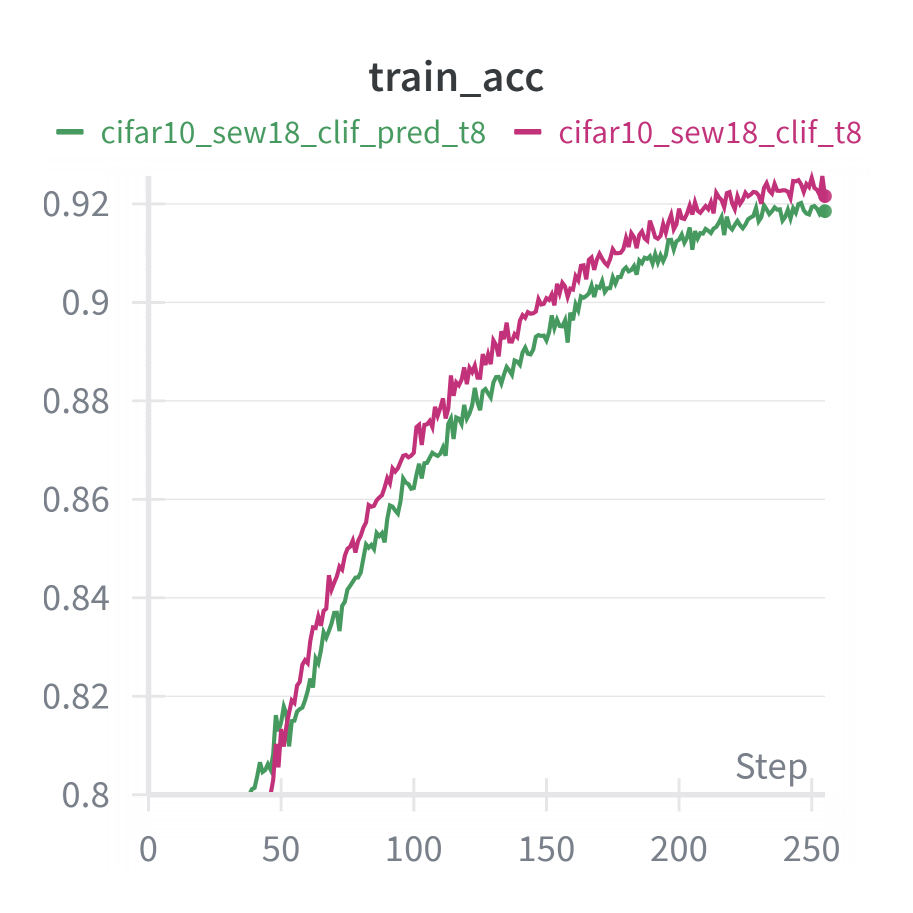}
}
\vskip -0.1in
\caption{The training and testing accuracy curves of the SEW ResNet18 trained on CIFAR10 dataset with IF,LIF,PLIF and CLIF neurons along with their variants enhanced by self-prediction mechanisms at time-steps T=4 and T=8.}
\label{pic:cifar10_sew18}
\end{center}
\end{figure}

\begin{figure}[h]
\begin{center}
\centerline{\includegraphics[width=0.25\columnwidth, trim=0.0cm 0.0cm 0.0cm 0.0cm, clip]{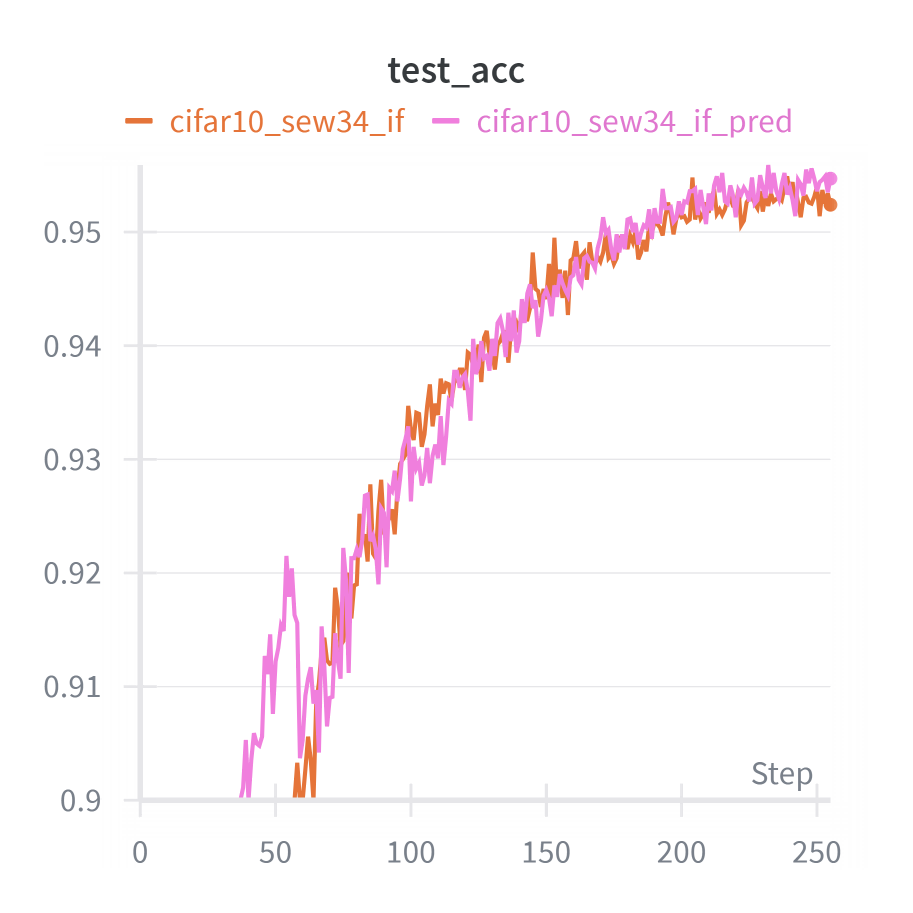}
\includegraphics[width=0.25\columnwidth, trim=0.0cm 0.0cm 0.0cm 0.0cm, clip]{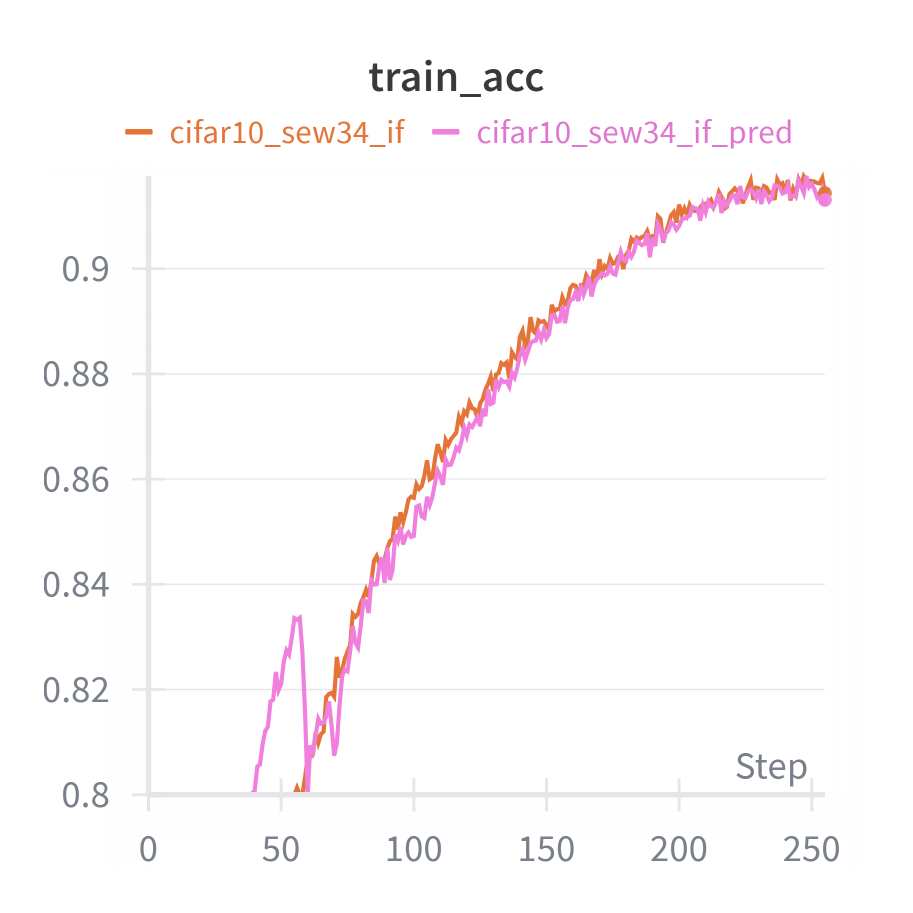}
\includegraphics[width=0.25\columnwidth, trim=0.0cm 0.0cm 0.0cm 0.0cm, clip]{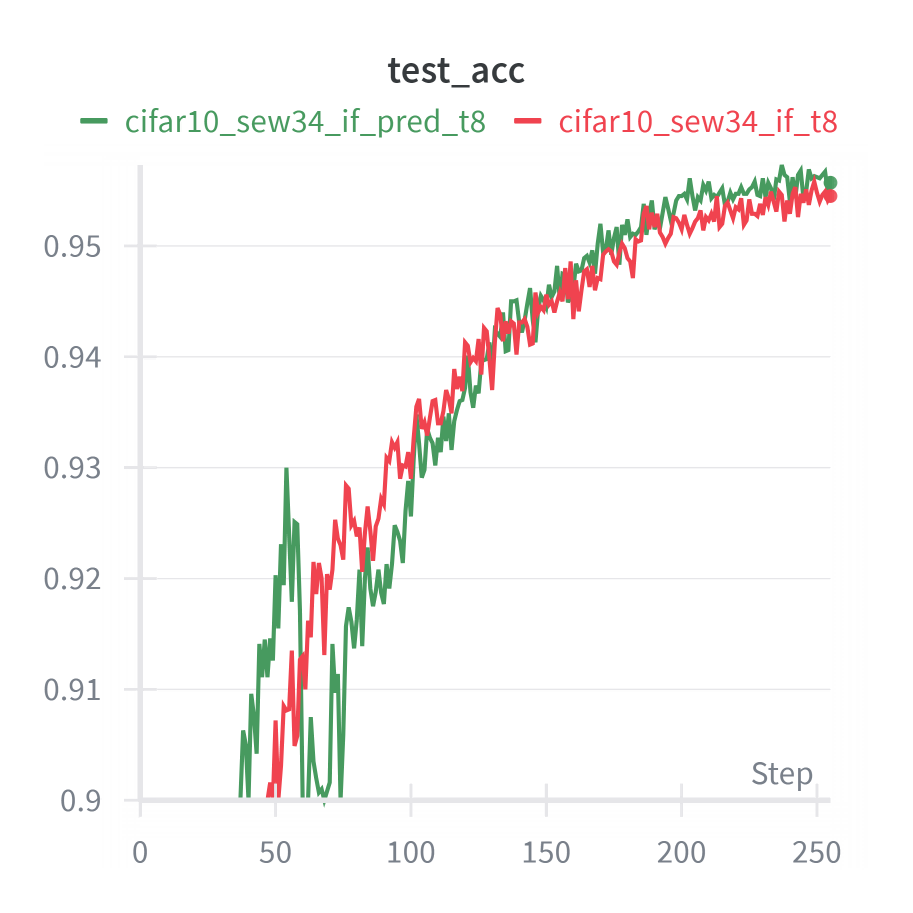}
\includegraphics[width=0.25\columnwidth, trim=0.0cm 0.0cm 0.0cm 0.0cm, clip]{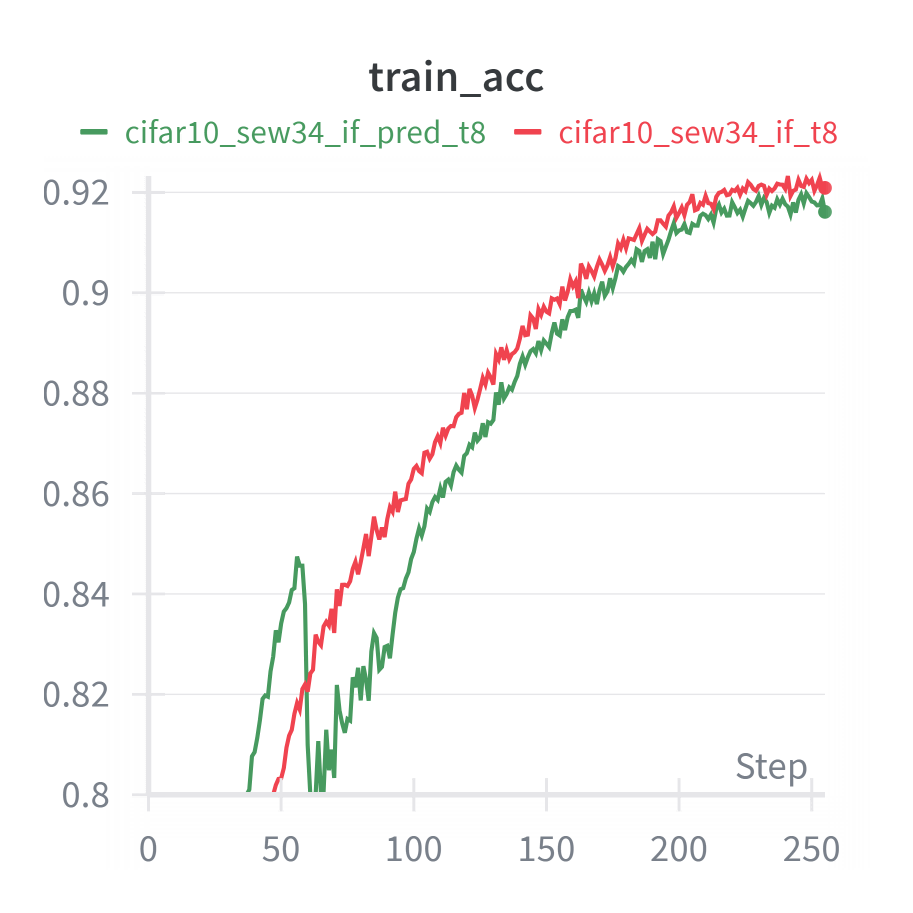}
}
\centerline{\includegraphics[width=0.25\columnwidth, trim=0.0cm 0.0cm 0.0cm 0.0cm, clip]{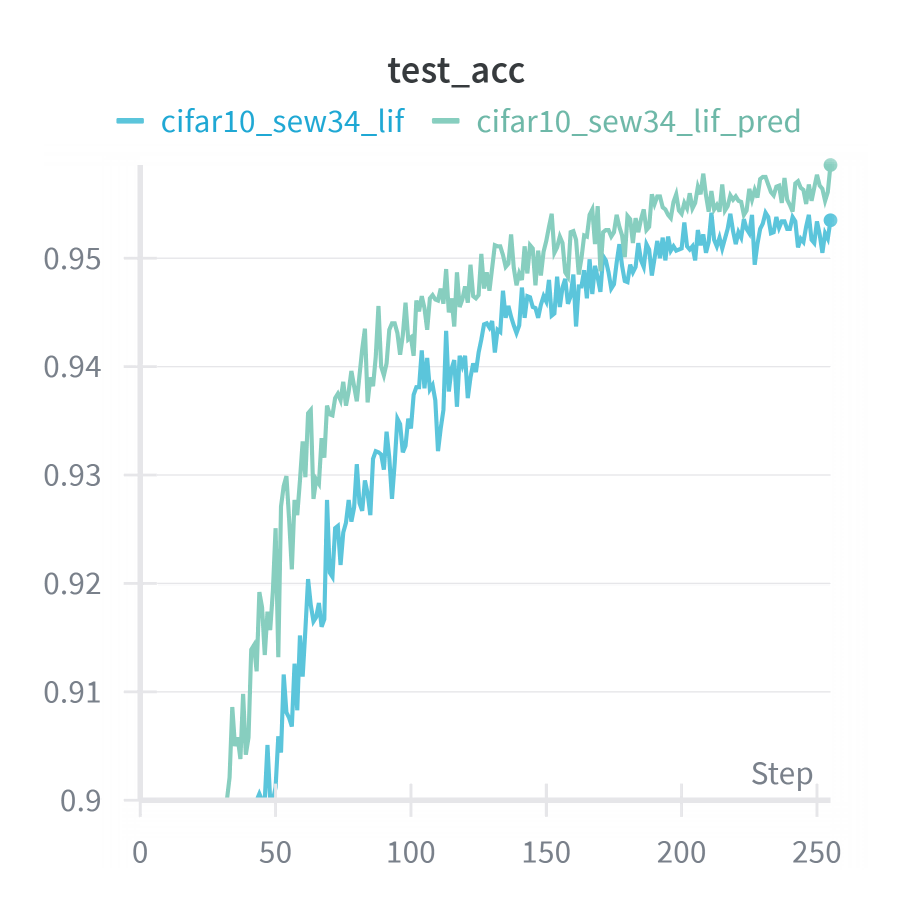}
\includegraphics[width=0.25\columnwidth, trim=0.0cm 0.0cm 0.0cm 0.0cm, clip]{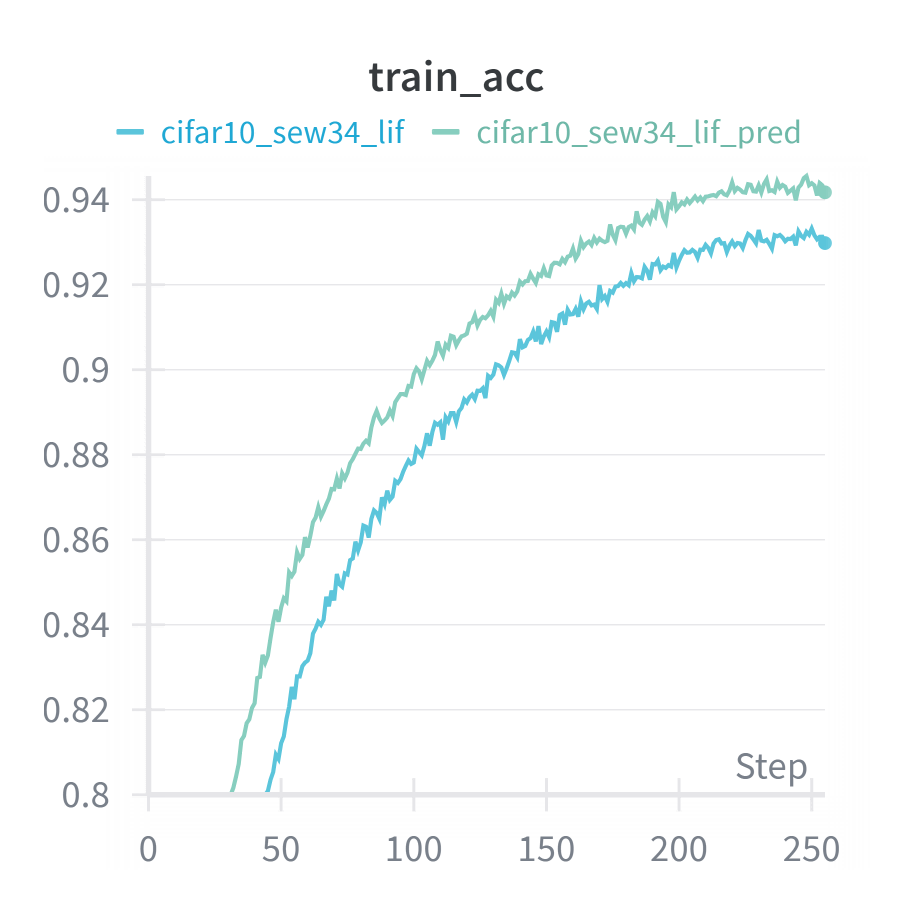}
\includegraphics[width=0.25\columnwidth, trim=0.0cm 0.0cm 0.0cm 0.0cm, clip]{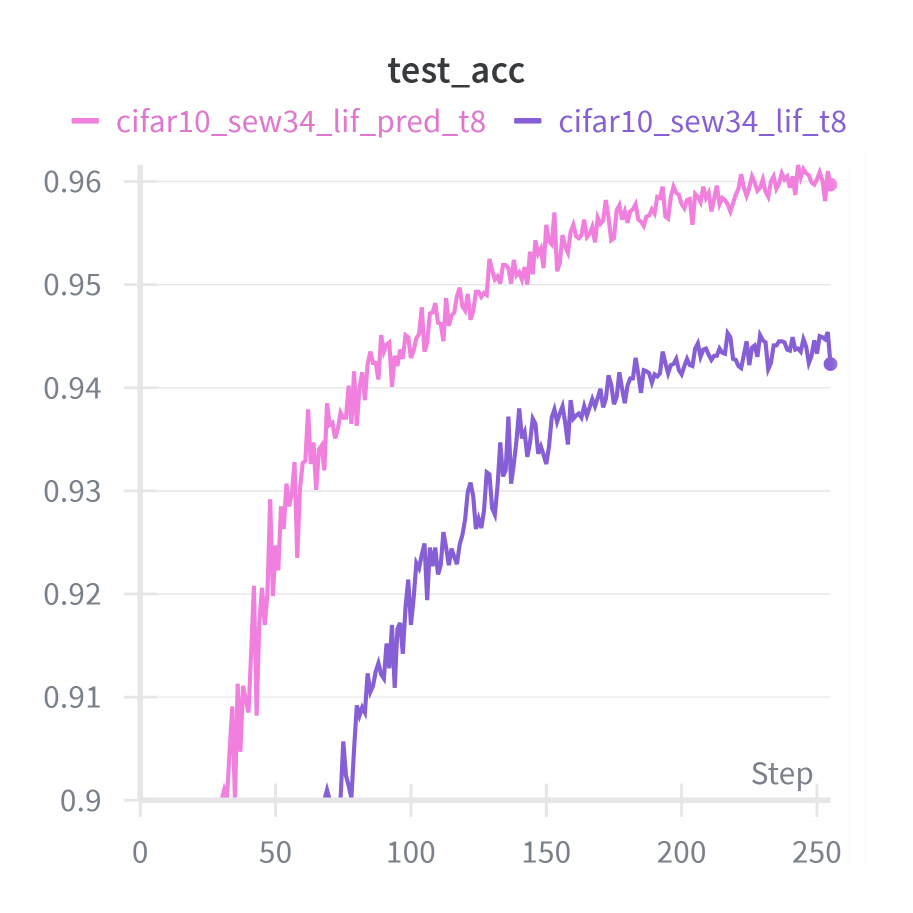}
\includegraphics[width=0.25\columnwidth, trim=0.0cm 0.0cm 0.0cm 0.0cm, clip]{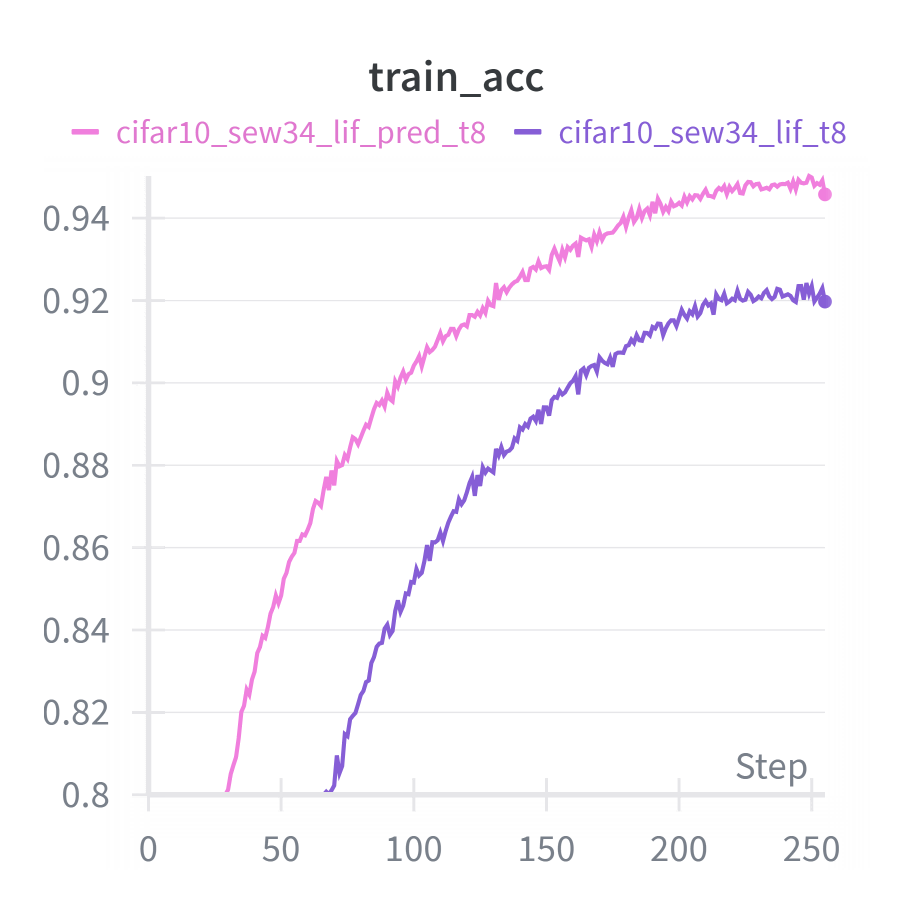}
}
\centerline{\includegraphics[width=0.25\columnwidth, trim=0.0cm 0.0cm 0.0cm 0.0cm, clip]{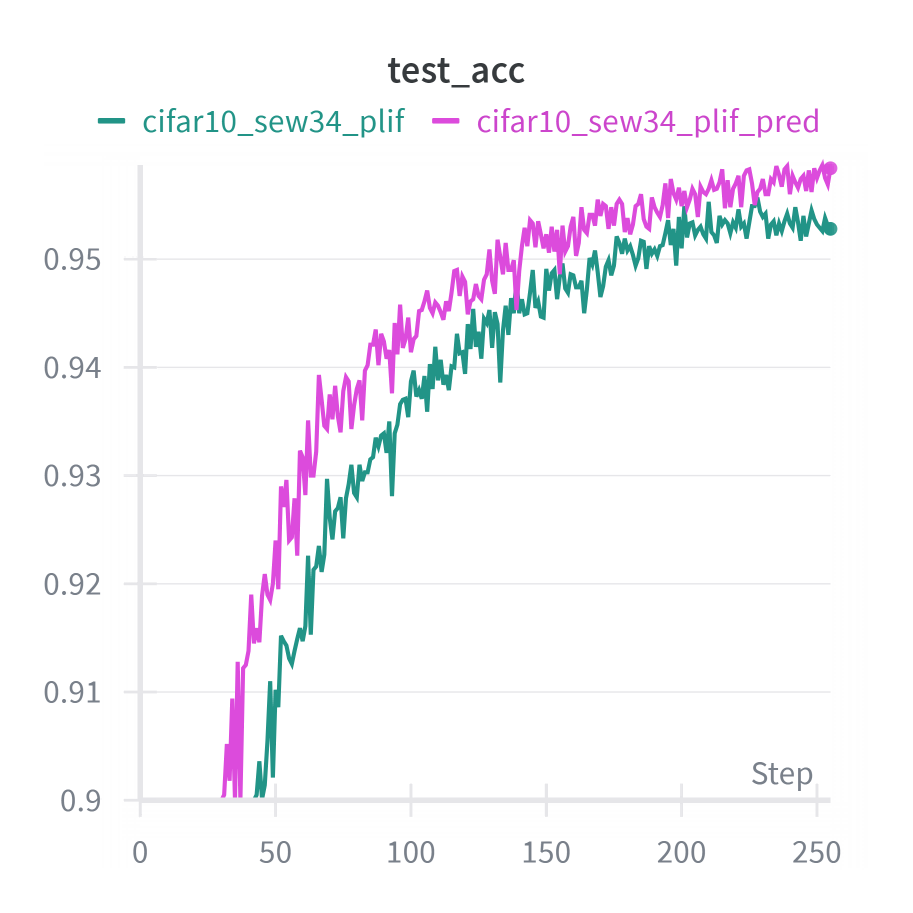}
\includegraphics[width=0.25\columnwidth, trim=0.0cm 0.0cm 0.0cm 0.0cm, clip]{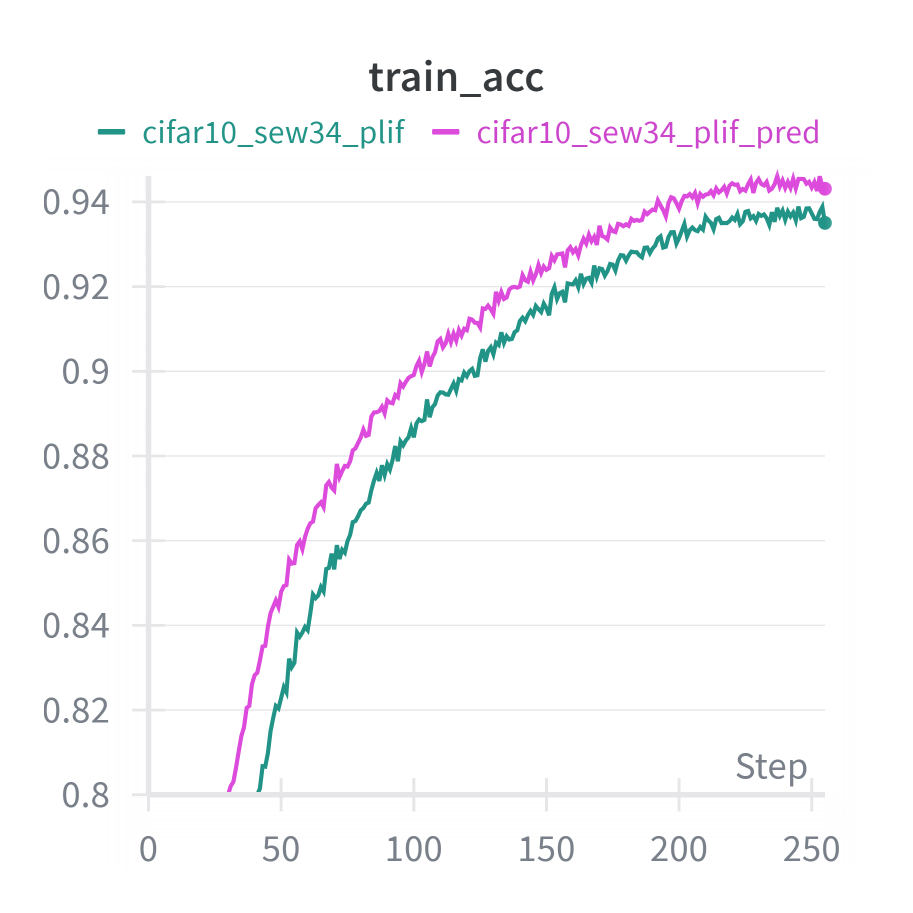}
\includegraphics[width=0.25\columnwidth, trim=0.0cm 0.0cm 0.0cm 0.0cm, clip]{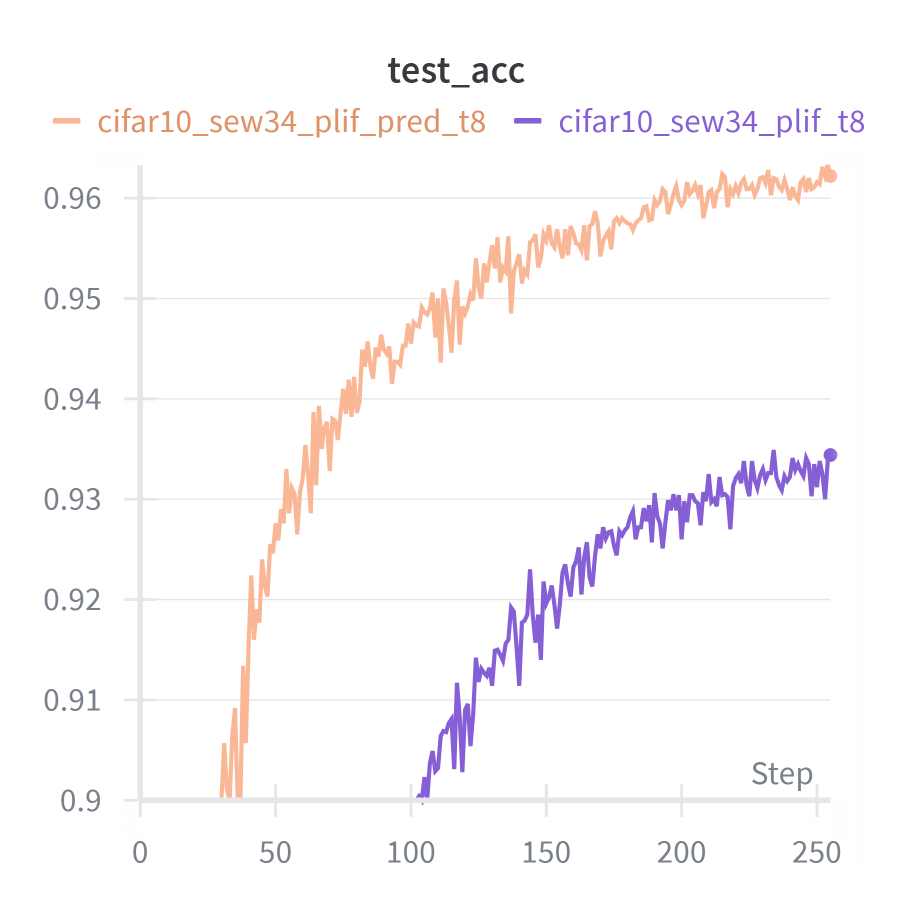}
\includegraphics[width=0.25\columnwidth, trim=0.0cm 0.0cm 0.0cm 0.0cm, clip]{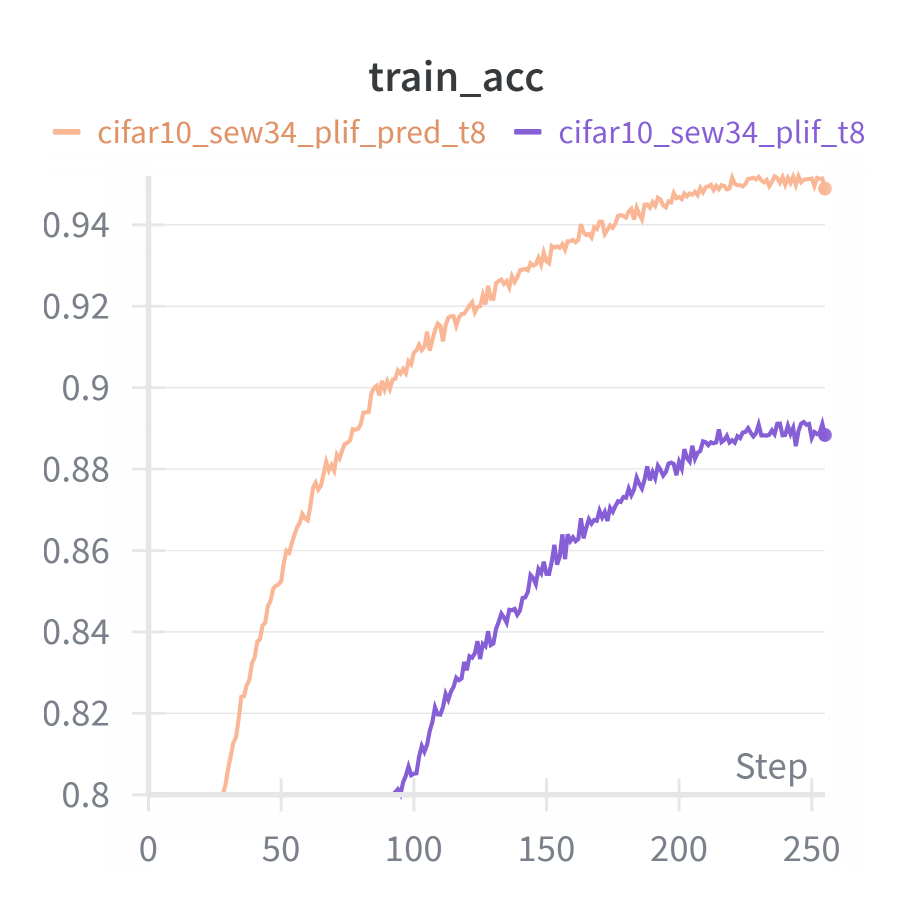}
}
\centerline{\includegraphics[width=0.25\columnwidth, trim=0.0cm 0.0cm 0.0cm 0.0cm, clip]{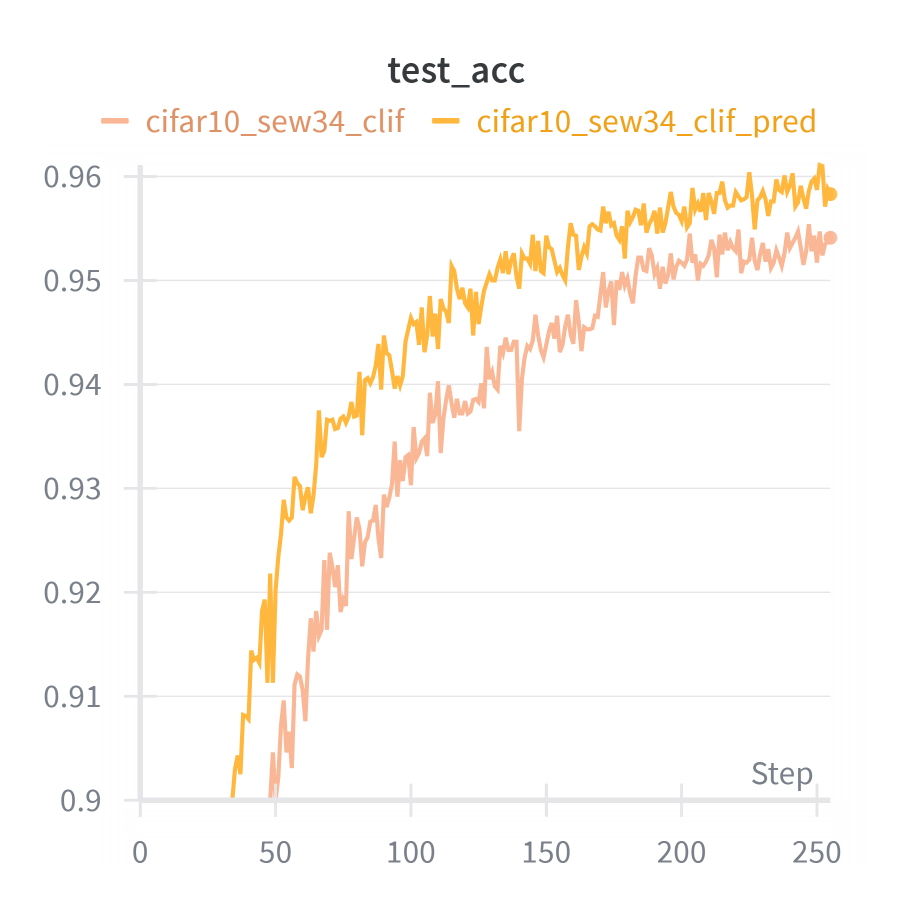}
\includegraphics[width=0.25\columnwidth, trim=0.0cm 0.0cm 0.0cm 0.0cm, clip]{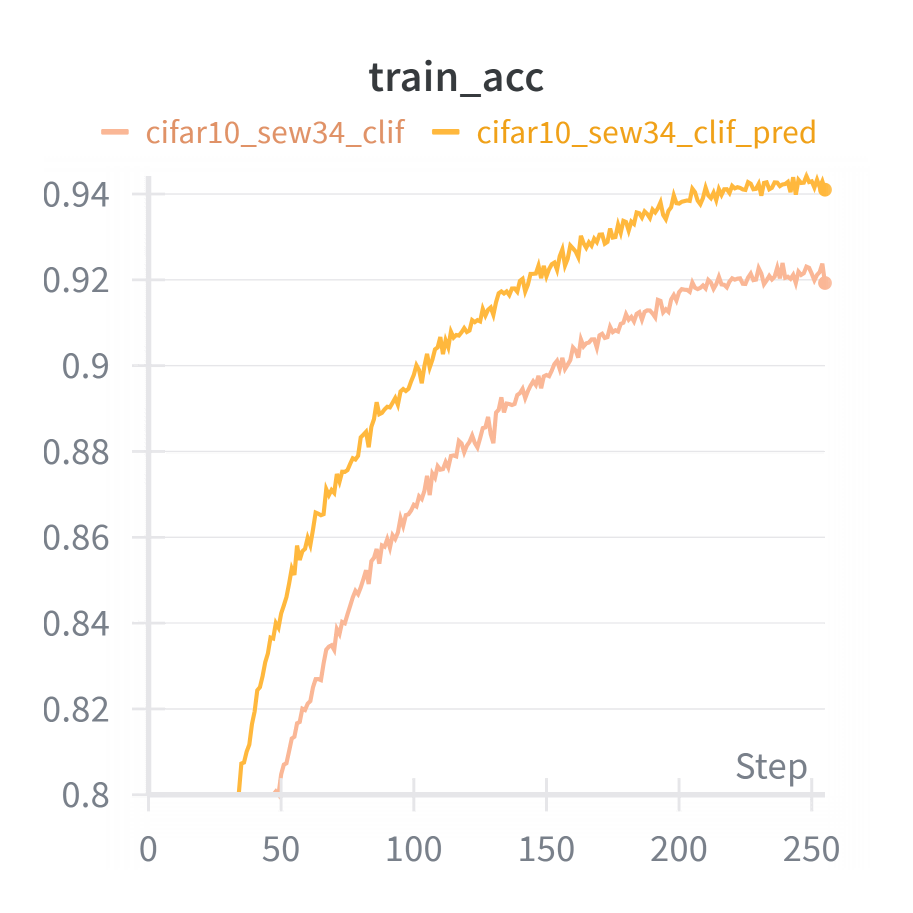}
\includegraphics[width=0.25\columnwidth, trim=0.0cm 0.0cm 0.0cm 0.0cm, clip]{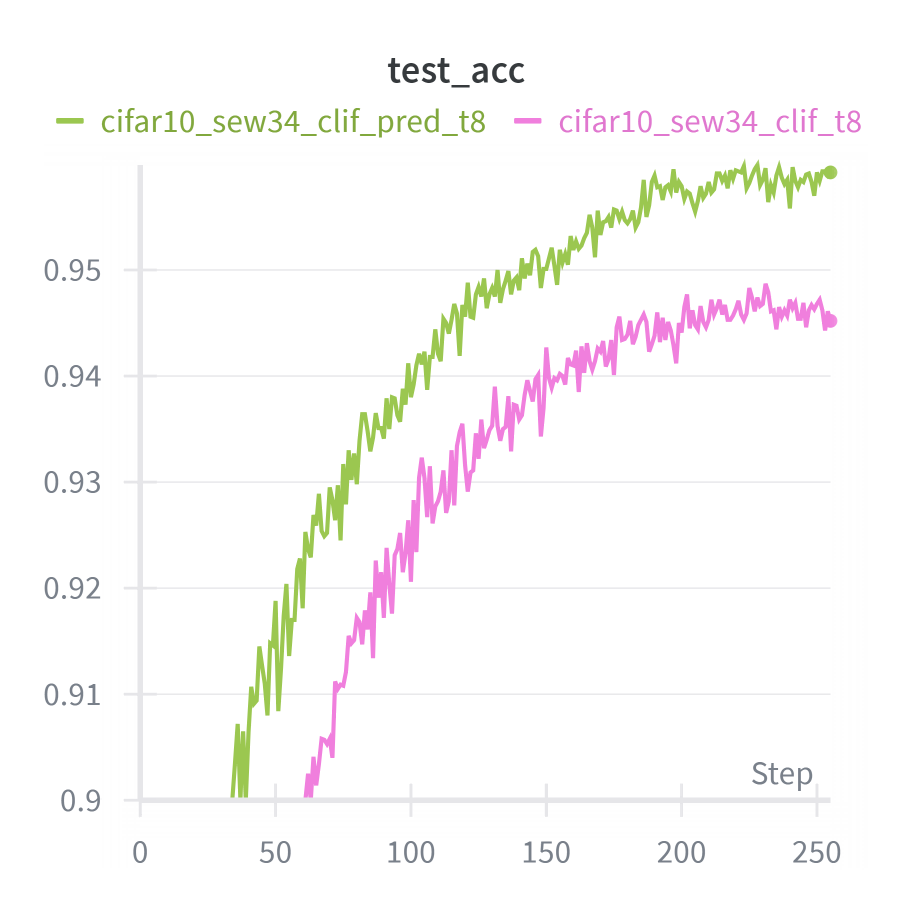}
\includegraphics[width=0.25\columnwidth, trim=0.0cm 0.0cm 0.0cm 0.0cm, clip]{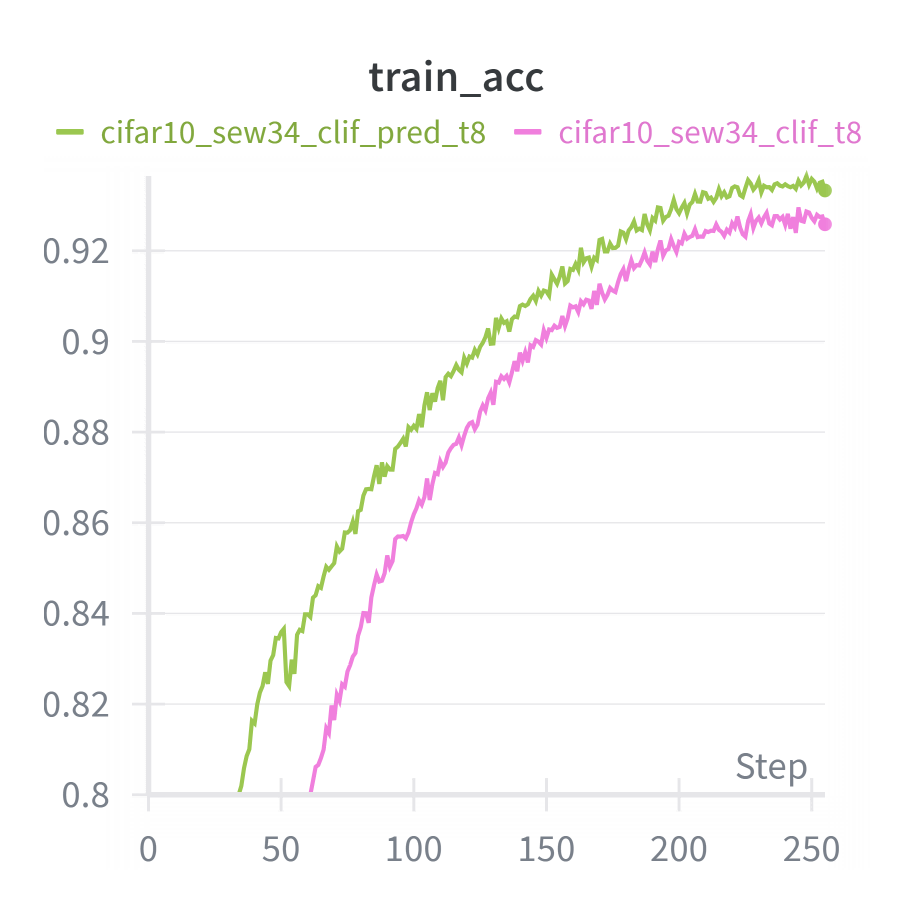}
}
\caption{The training and testing accuracy curves of the SEW ResNet34 trained on CIFAR10 dataset with IF,LIF,PLIF and CLIF neurons along with their variants enhanced by self-prediction mechanisms at time-steps T=4 and T=8.}
\label{pic:cifar10_sew34}
\end{center}
\end{figure}

For the SEW ResNet families which is originally designed for ImageNet, when adapting them to CIFAR-10, we modify the first convolutional layer parameters, changing kernel size, stride, and padding from 7, 2, 3 to 3, 1, 1, respectively, and replace the initial max pooling layer with an identity mapping. Figure~\ref{pic:cifar10_sew18} shows the training and testing accuracy curve on SEW ResNet18. Figure~\ref{pic:cifar10_sew34} shows the training and testing accuracy curve on SEW ResNet34.

\clearpage
\subsection{Results on Spiking ResNet}
\begin{figure}[h]
\begin{center}
\vskip -0.15in
\centerline{\includegraphics[width=0.25\columnwidth, trim=0.0cm 0.0cm 0.0cm 0.0cm, clip]{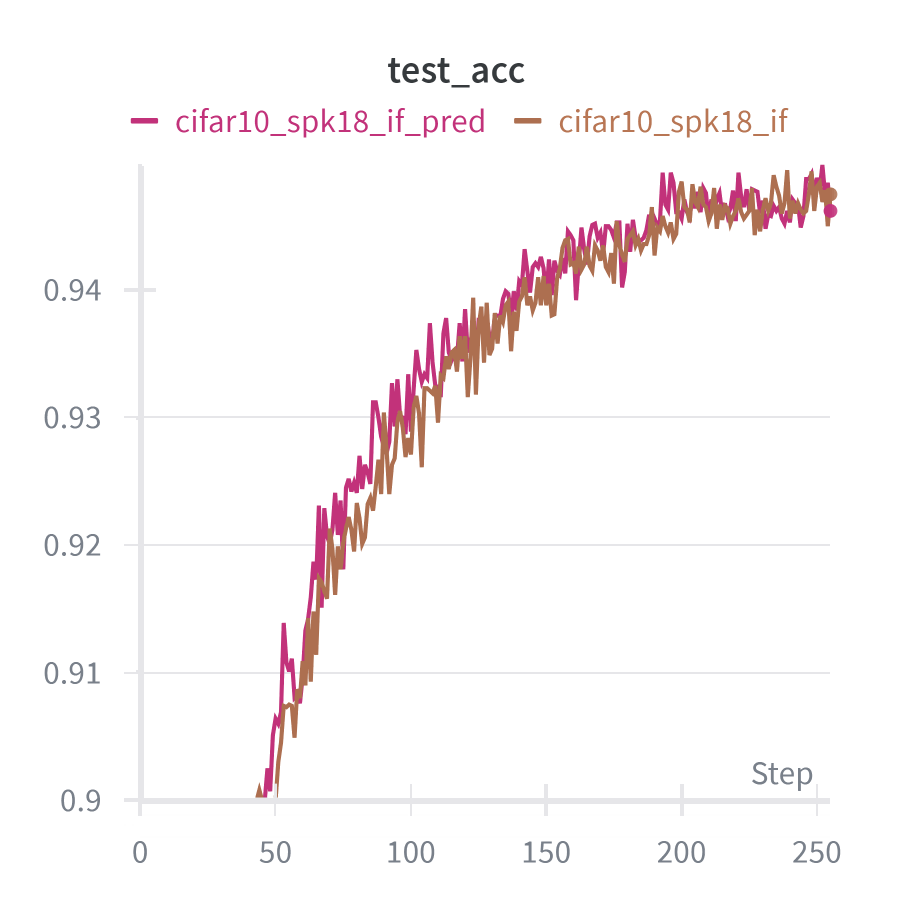}
\includegraphics[width=0.25\columnwidth, trim=0.0cm 0.0cm 0.0cm 0.0cm, clip]{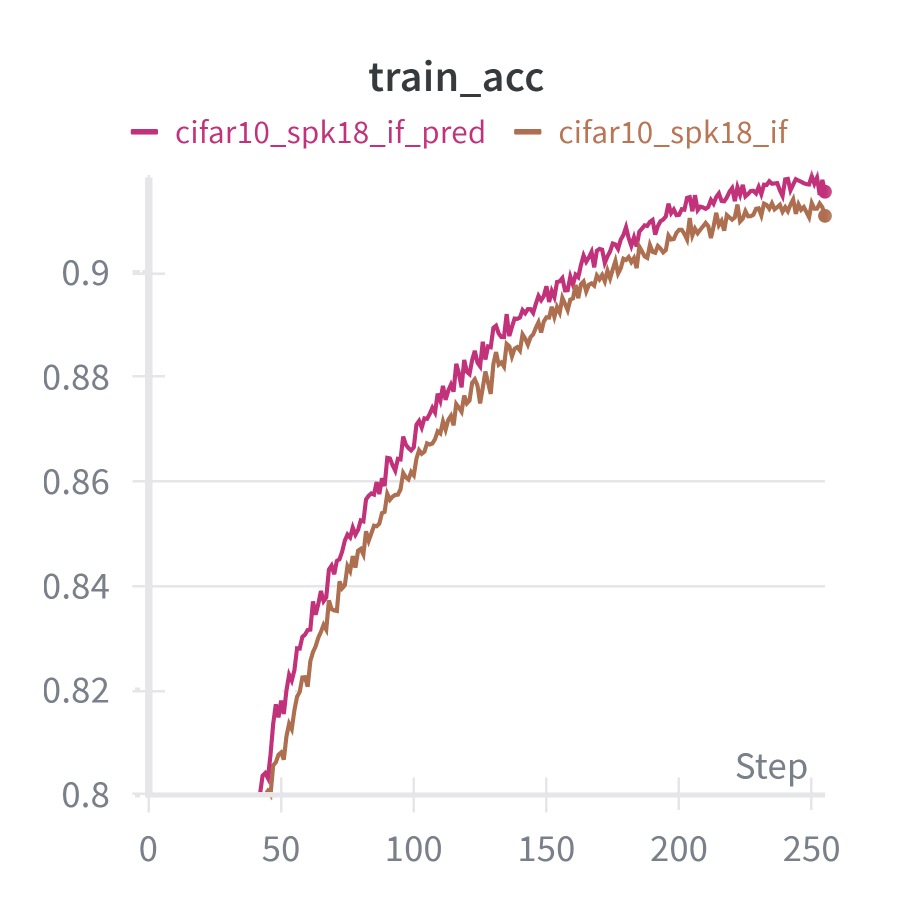}
\includegraphics[width=0.25\columnwidth, trim=0.0cm 0.0cm 0.0cm 0.0cm, clip]{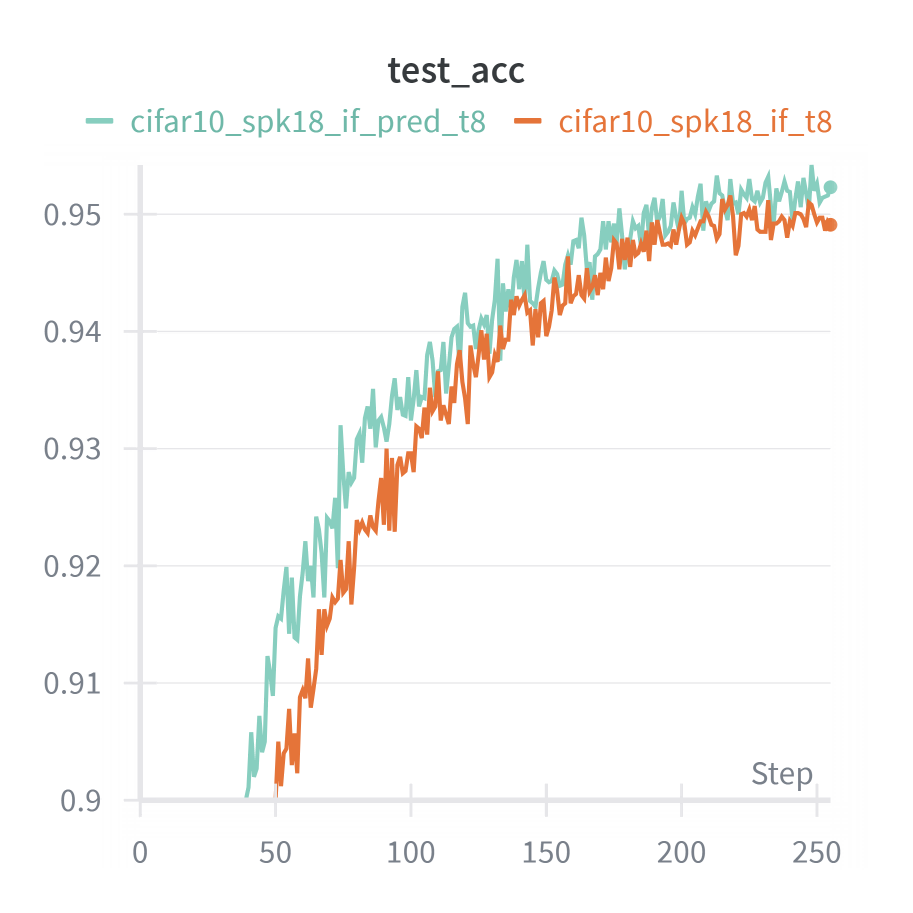}
\includegraphics[width=0.25\columnwidth, trim=0.0cm 0.0cm 0.0cm 0.0cm, clip]{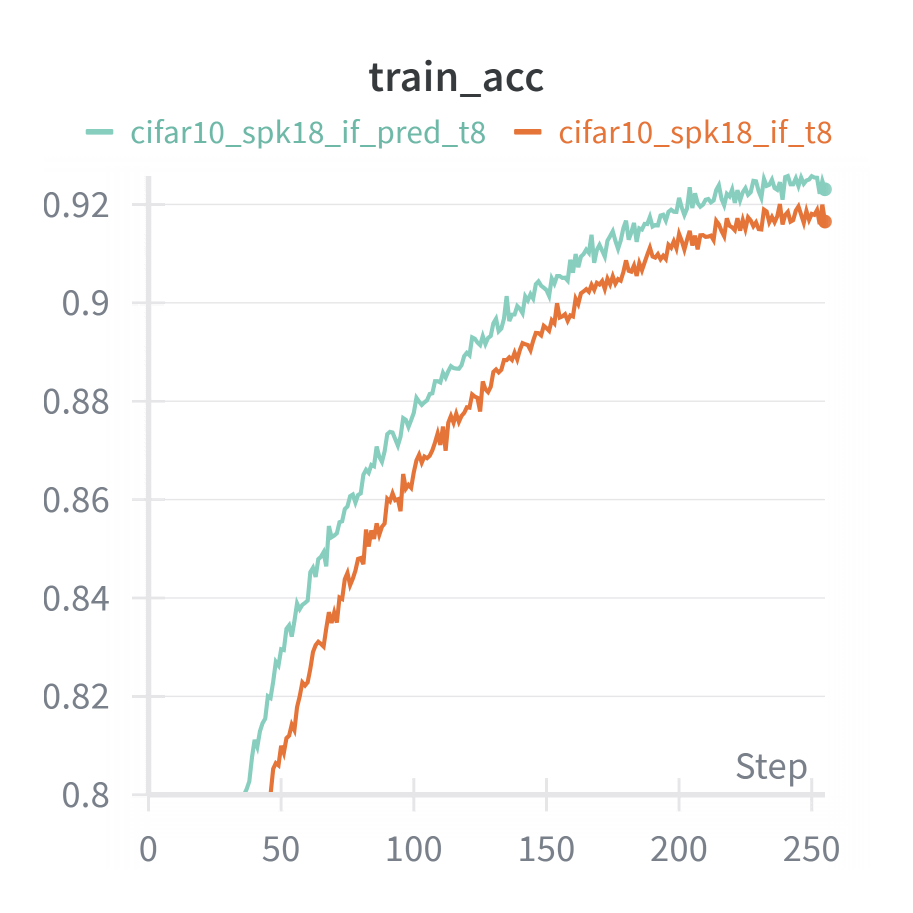}
}
\vskip -0.05in
\centerline{\includegraphics[width=0.25\columnwidth, trim=0.0cm 0.0cm 0.0cm 0.0cm, clip]{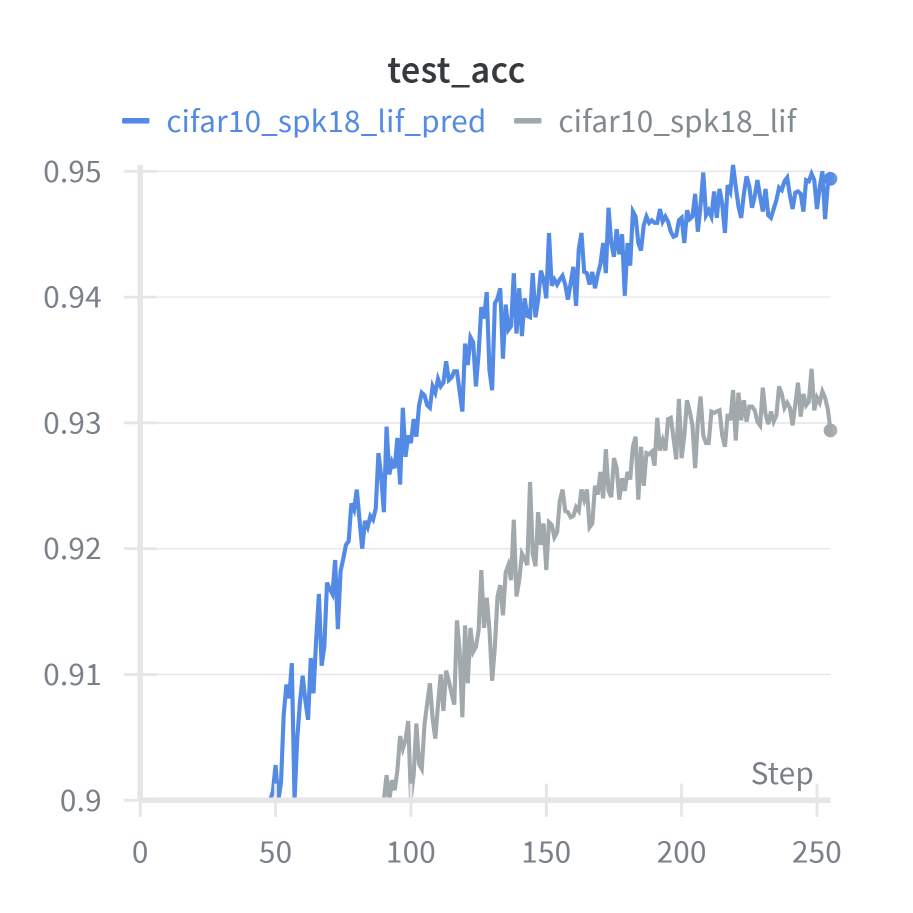}
\includegraphics[width=0.25\columnwidth, trim=0.0cm 0.0cm 0.0cm 0.0cm, clip]{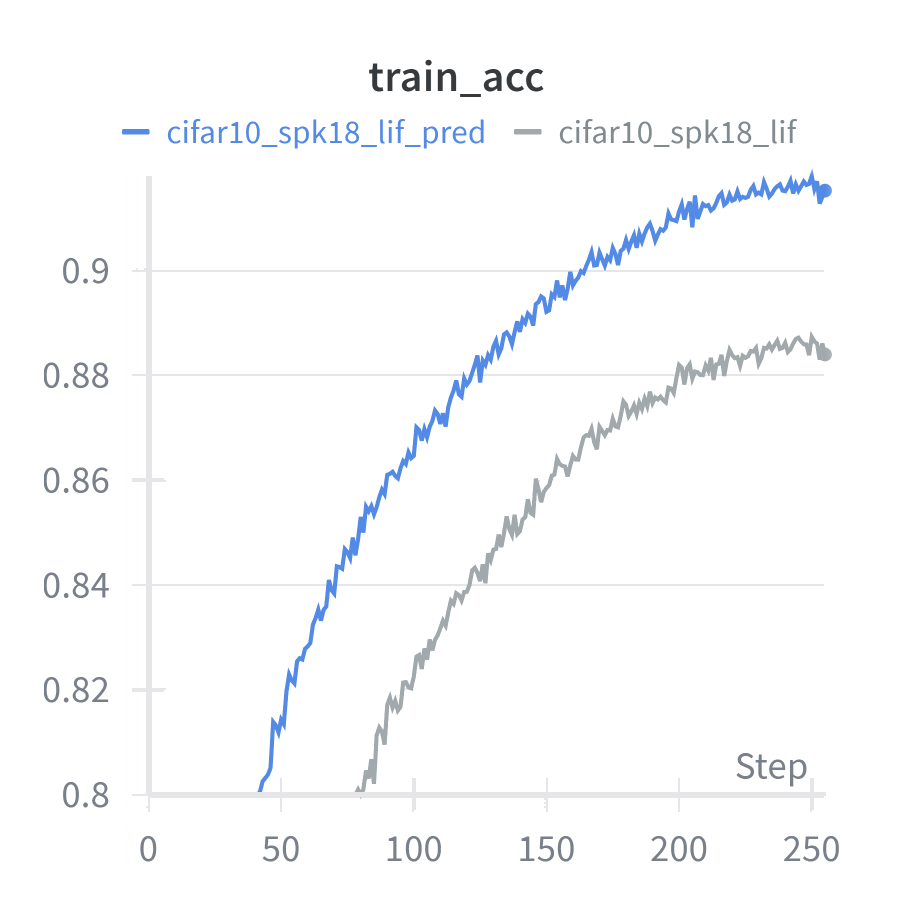}
\includegraphics[width=0.25\columnwidth, trim=0.0cm 0.0cm 0.0cm 0.0cm, clip]{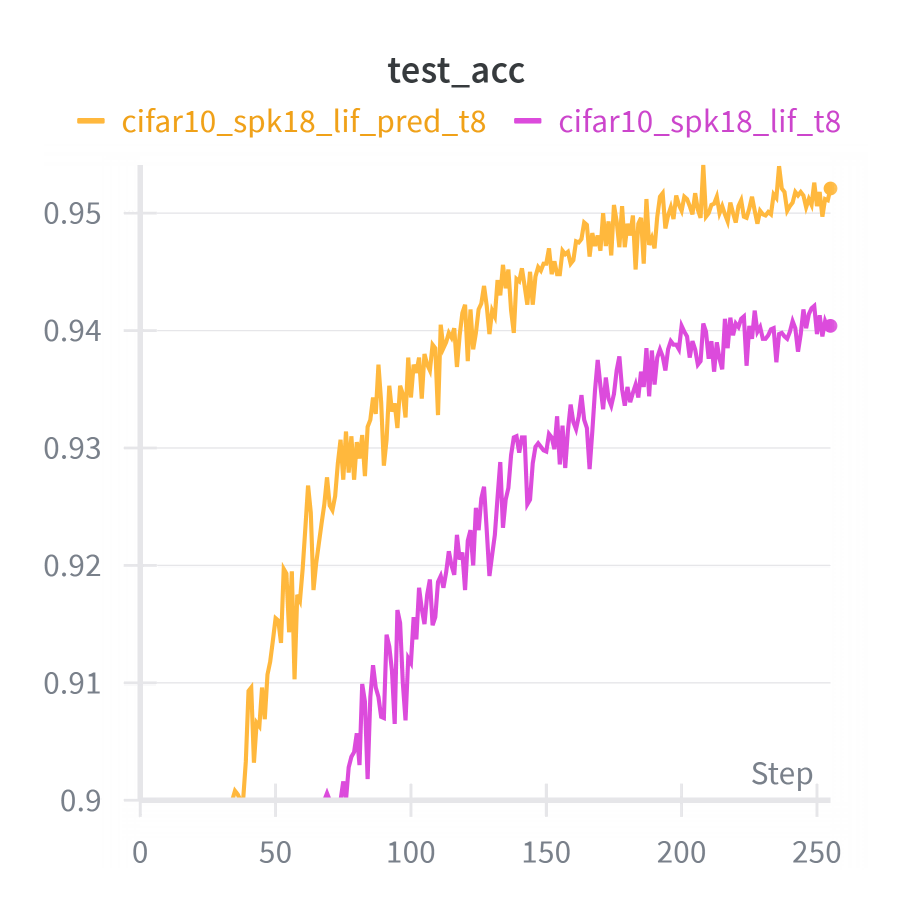}
\includegraphics[width=0.25\columnwidth, trim=0.0cm 0.0cm 0.0cm 0.0cm, clip]{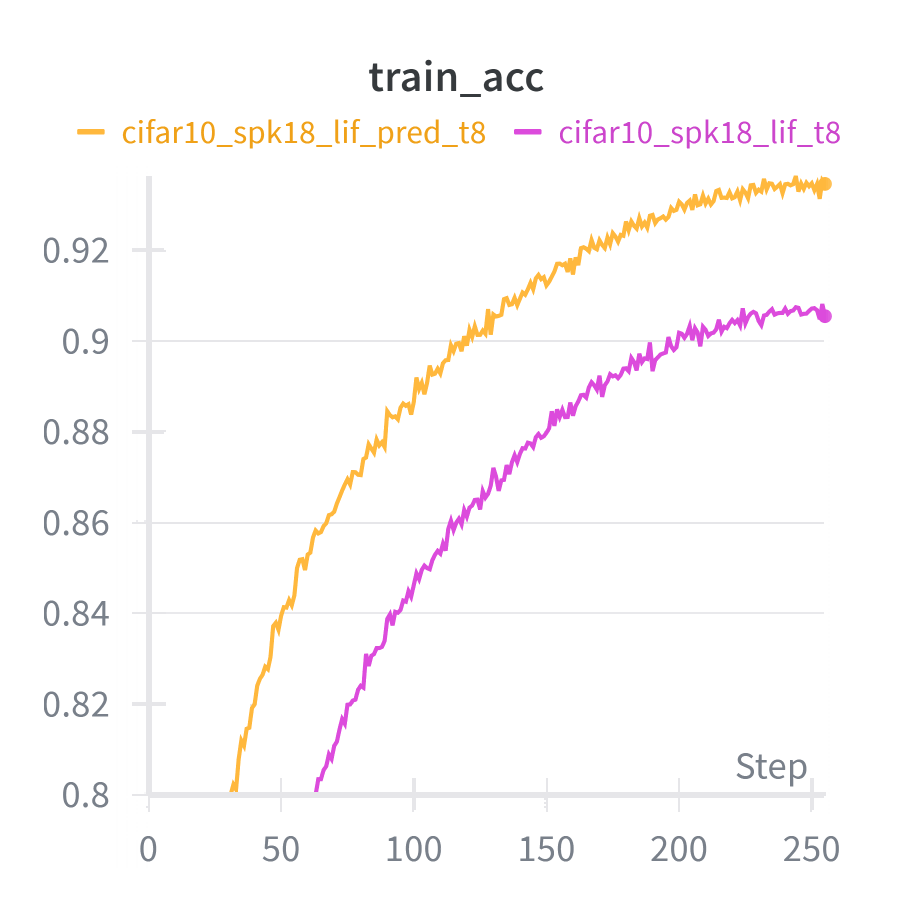}
}
\vskip -0.05in
\centerline{\includegraphics[width=0.25\columnwidth, trim=0.0cm 0.0cm 0.0cm 0.0cm, clip]{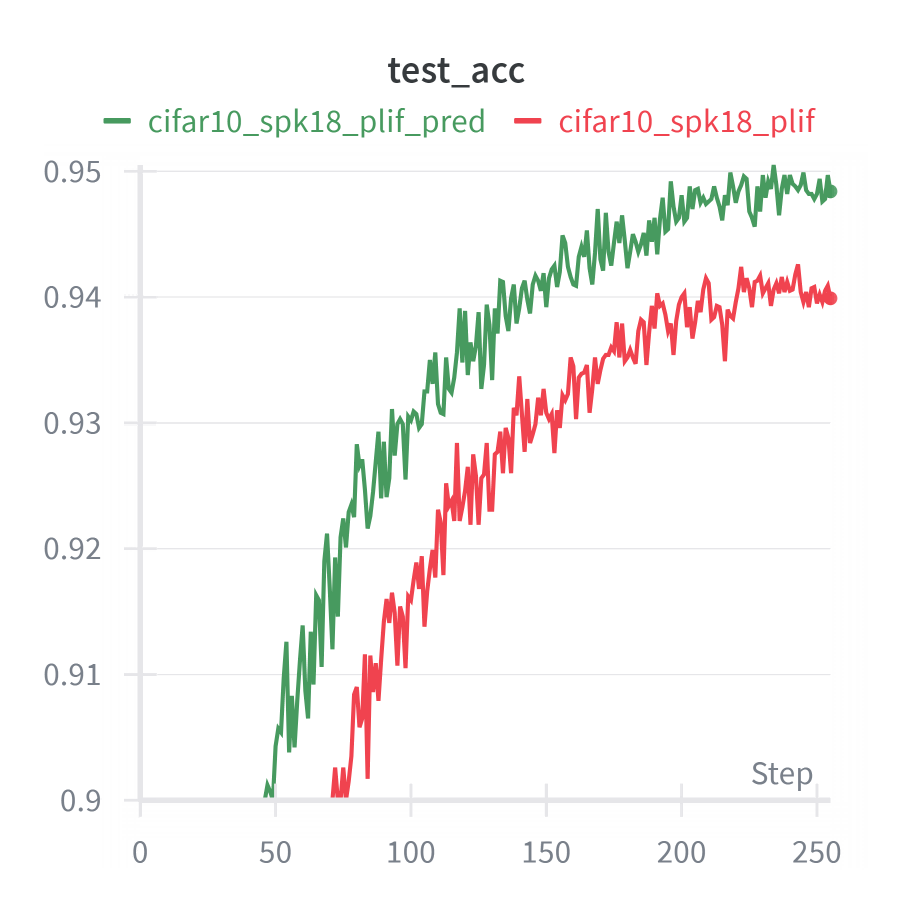}
\includegraphics[width=0.25\columnwidth, trim=0.0cm 0.0cm 0.0cm 0.0cm, clip]{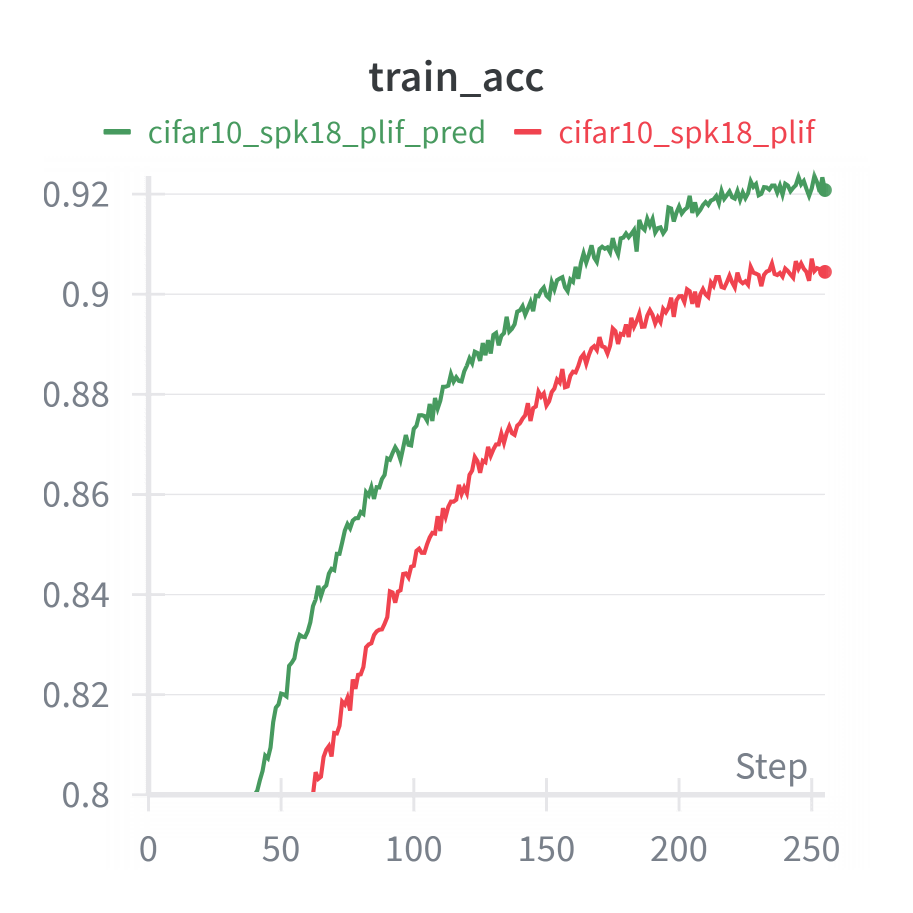}
\includegraphics[width=0.25\columnwidth, trim=0.0cm 0.0cm 0.0cm 0.0cm, clip]{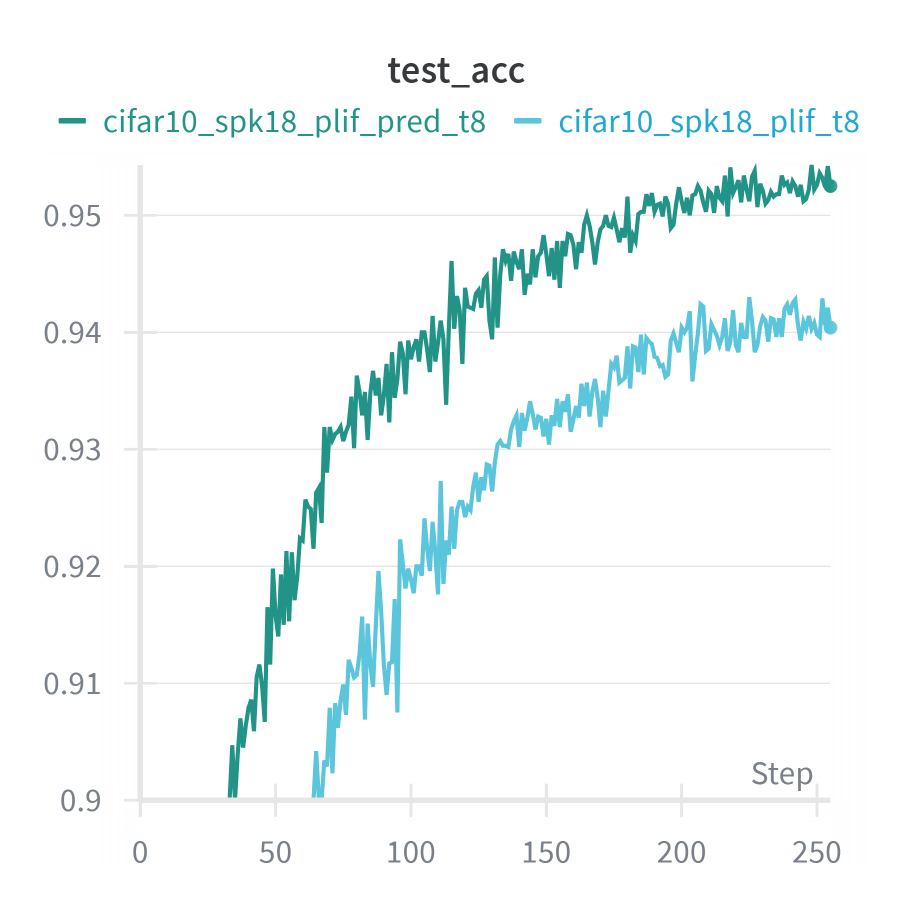}
\includegraphics[width=0.25\columnwidth, trim=0.0cm 0.0cm 0.0cm 0.0cm, clip]{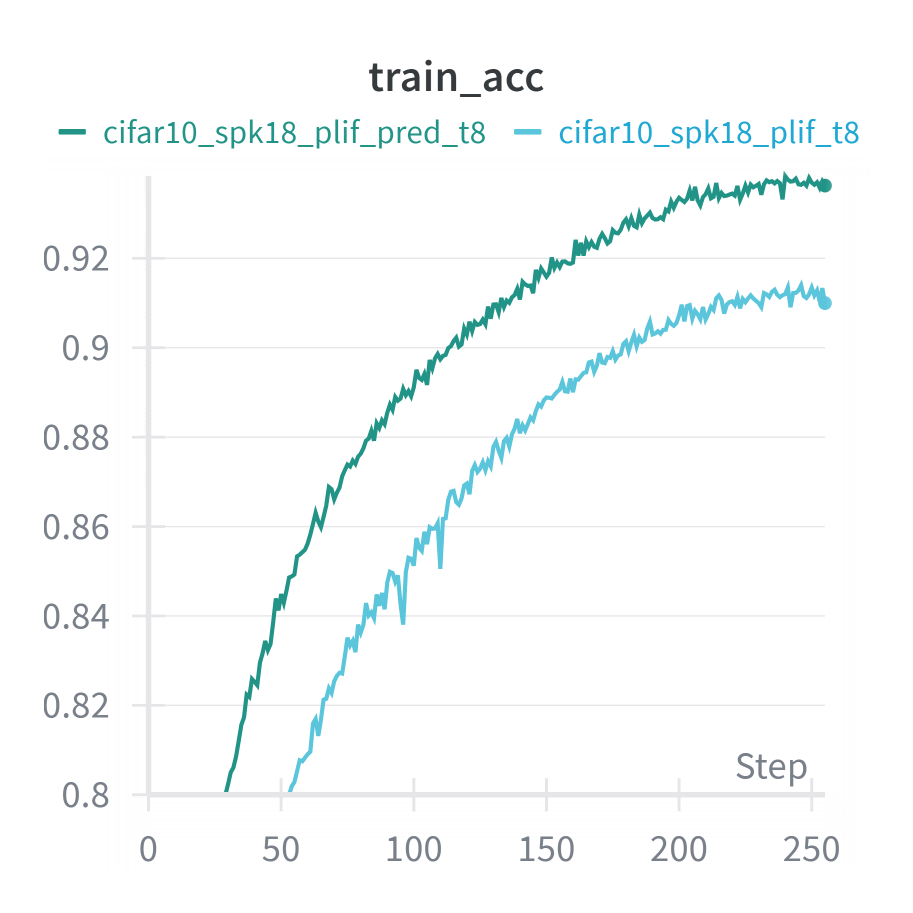}
}
\vskip -0.05in
\centerline{\includegraphics[width=0.25\columnwidth, trim=0.0cm 0.0cm 0.0cm 0.0cm, clip]{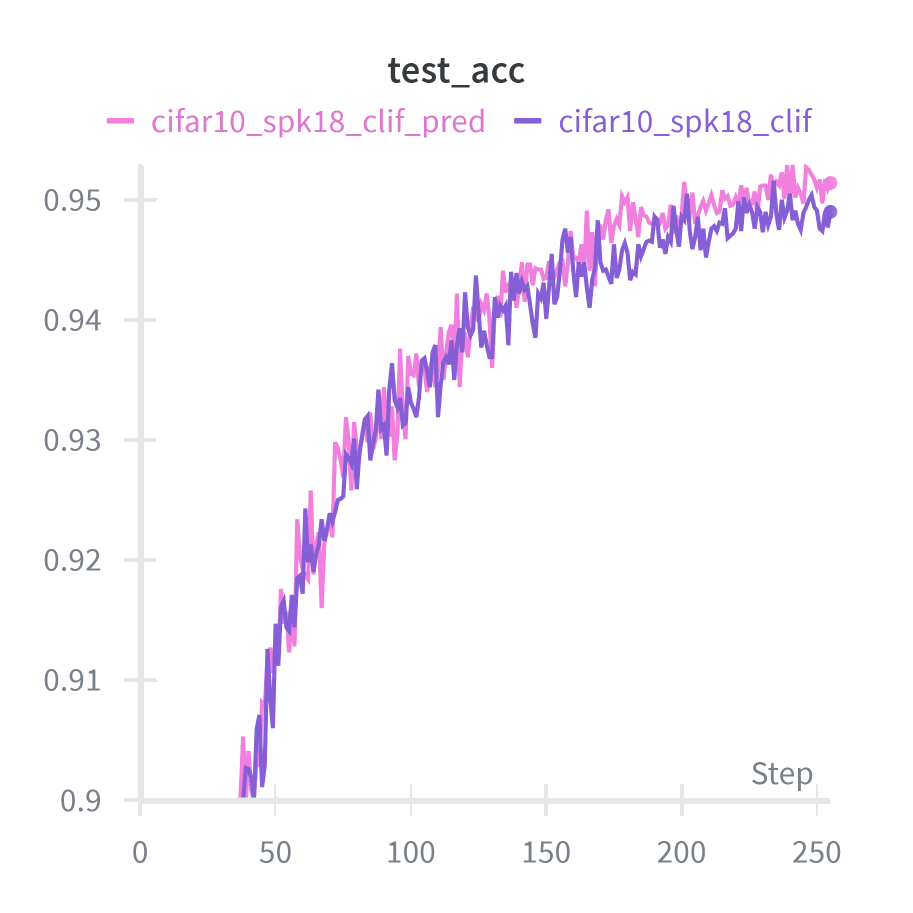}
\includegraphics[width=0.25\columnwidth, trim=0.0cm 0.0cm 0.0cm 0.0cm, clip]{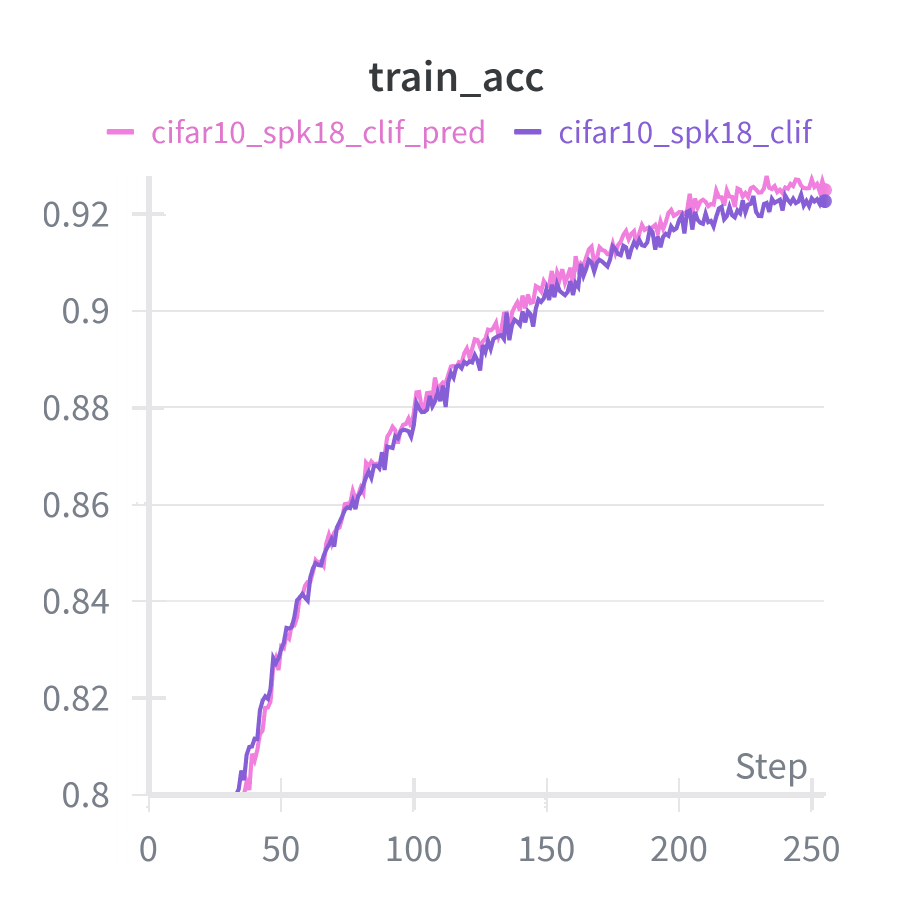}
\includegraphics[width=0.25\columnwidth, trim=0.0cm 0.0cm 0.0cm 0.0cm, clip]{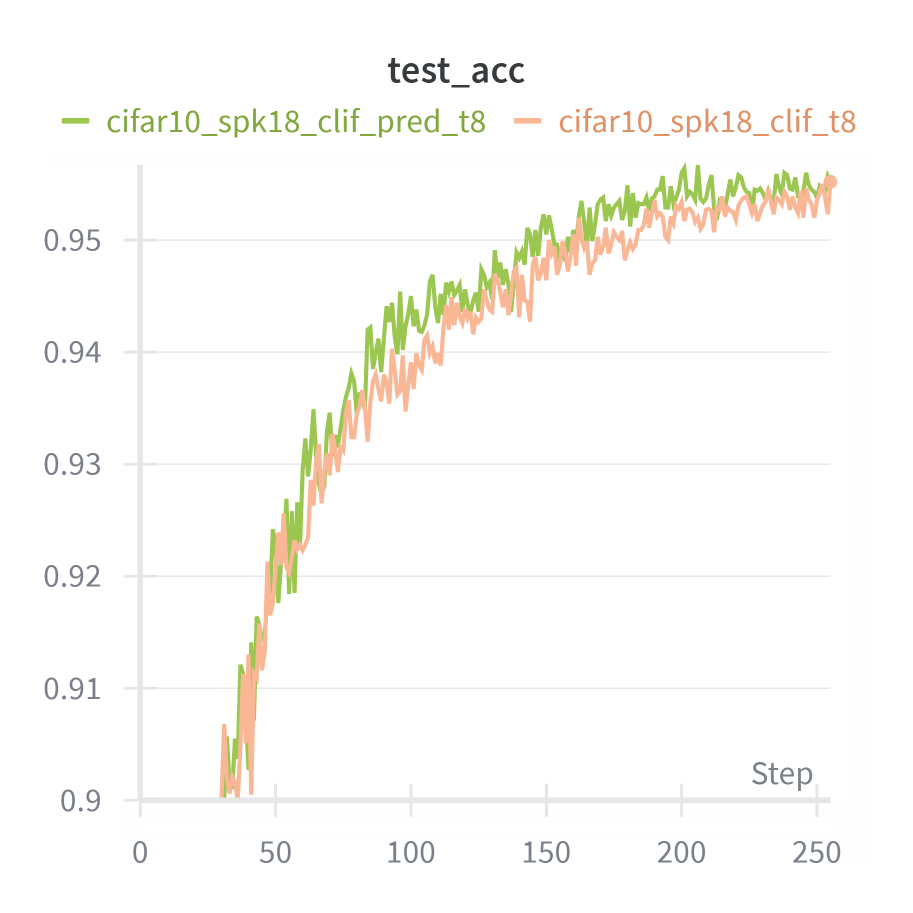}
\includegraphics[width=0.25\columnwidth, trim=0.0cm 0.0cm 0.0cm 0.0cm, clip]{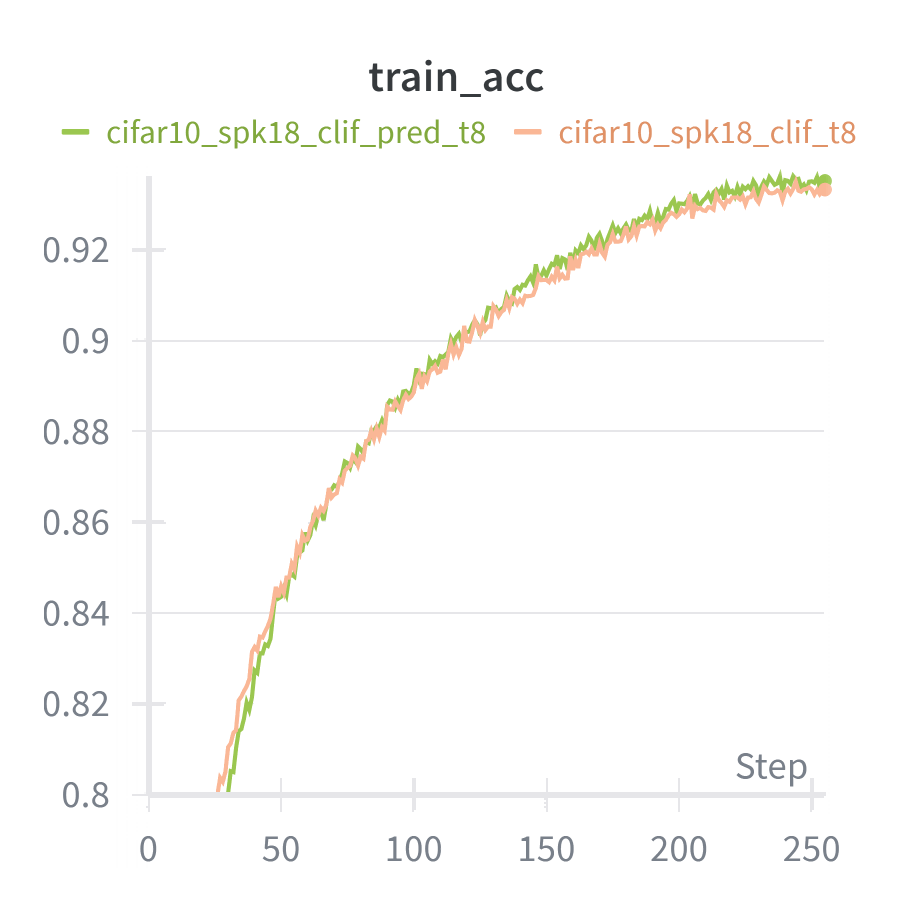}
}
\vskip -0.1in
\caption{The training and testing accuracy curves of the Spiking ResNet18 trained on CIFAR10 dataset with IF,LIF,PLIF and CLIF neurons along with their variants enhanced by self-prediction mechanisms at time-steps T=4 and T=8.}
\label{pic:cifar10_spk18}
\end{center}
\end{figure}

\begin{figure}[h]
\begin{center}
\vskip -0.15in
\centerline{\includegraphics[width=0.25\columnwidth, trim=0.0cm 0.0cm 0.0cm 0.0cm, clip]{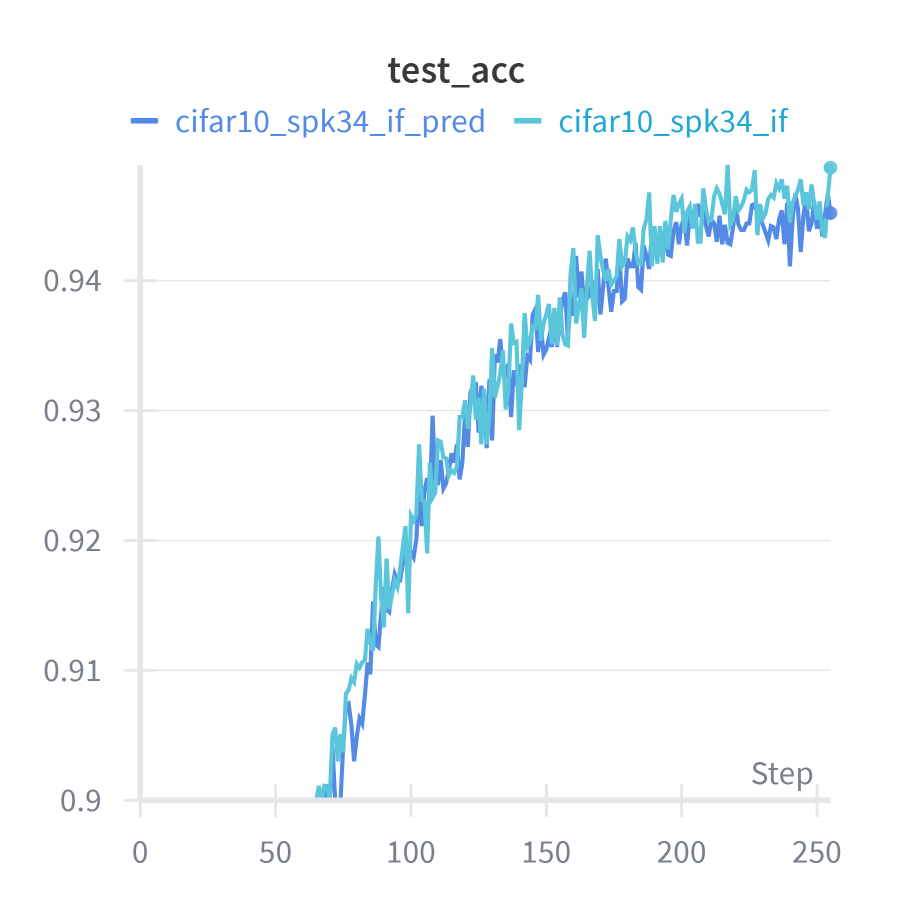}
\includegraphics[width=0.25\columnwidth, trim=0.0cm 0.0cm 0.0cm 0.0cm, clip]{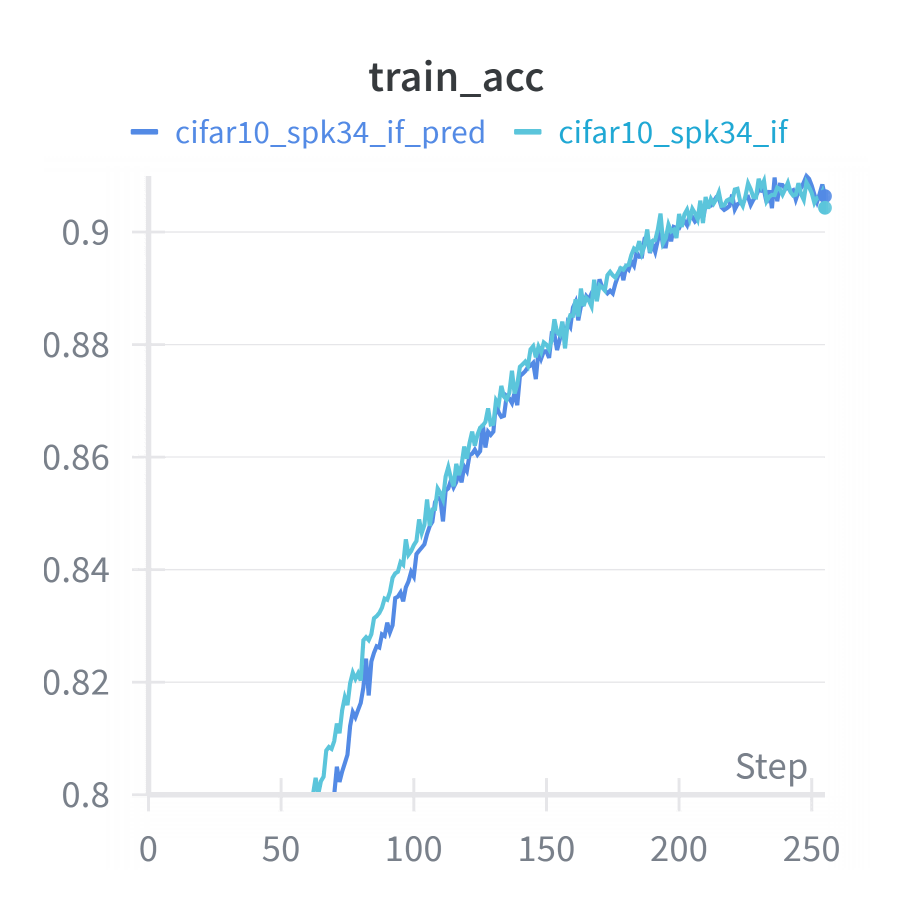}
\includegraphics[width=0.25\columnwidth, trim=0.0cm 0.0cm 0.0cm 0.0cm, clip]{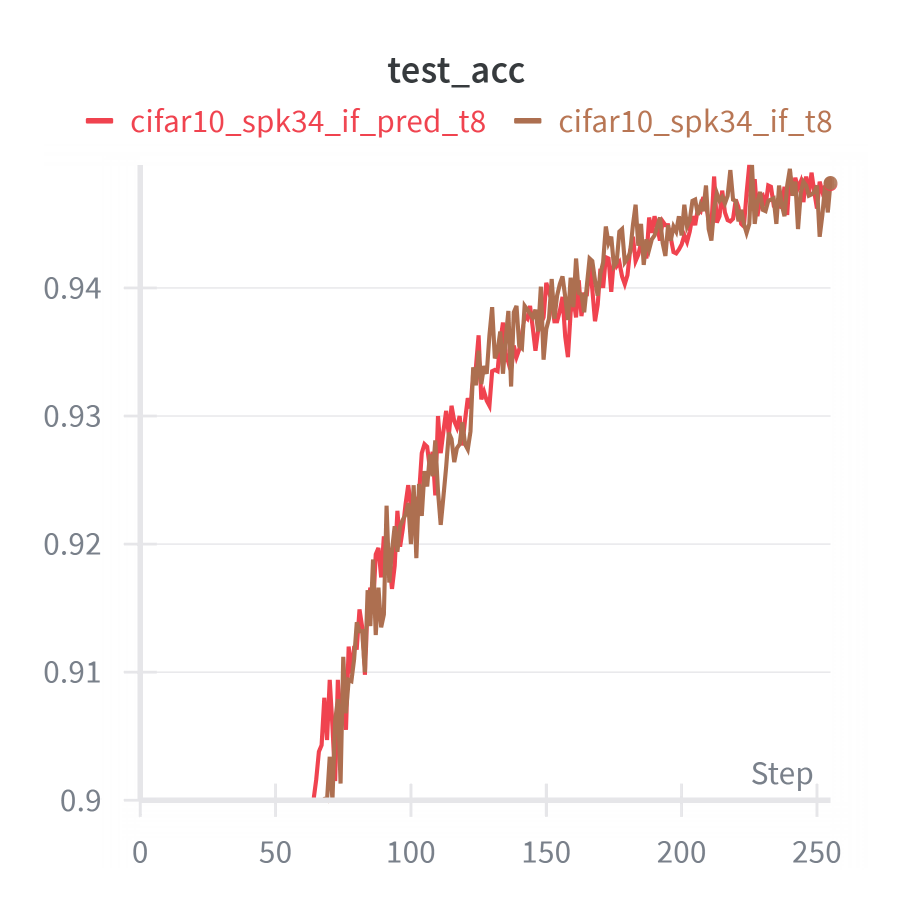}
\includegraphics[width=0.25\columnwidth, trim=0.0cm 0.0cm 0.0cm 0.0cm, clip]{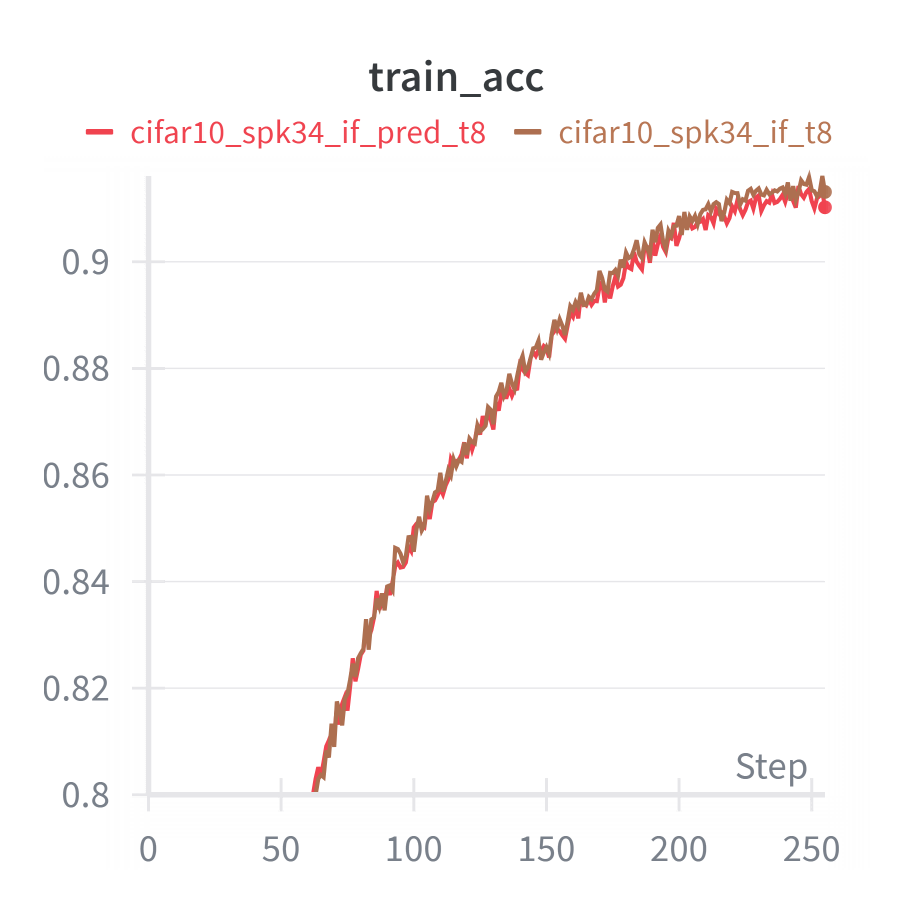}
}
\vskip -0.05in
\centerline{\includegraphics[width=0.25\columnwidth, trim=0.0cm 0.0cm 0.0cm 0.0cm, clip]{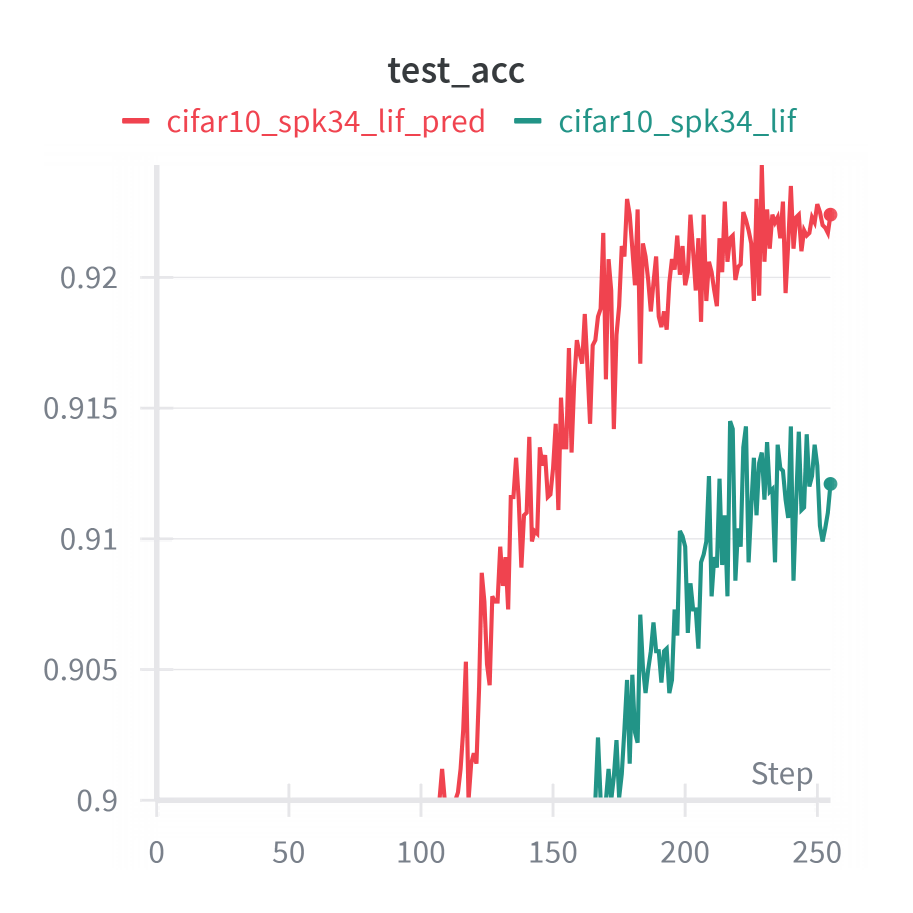}
\includegraphics[width=0.25\columnwidth, trim=0.0cm 0.0cm 0.0cm 0.0cm, clip]{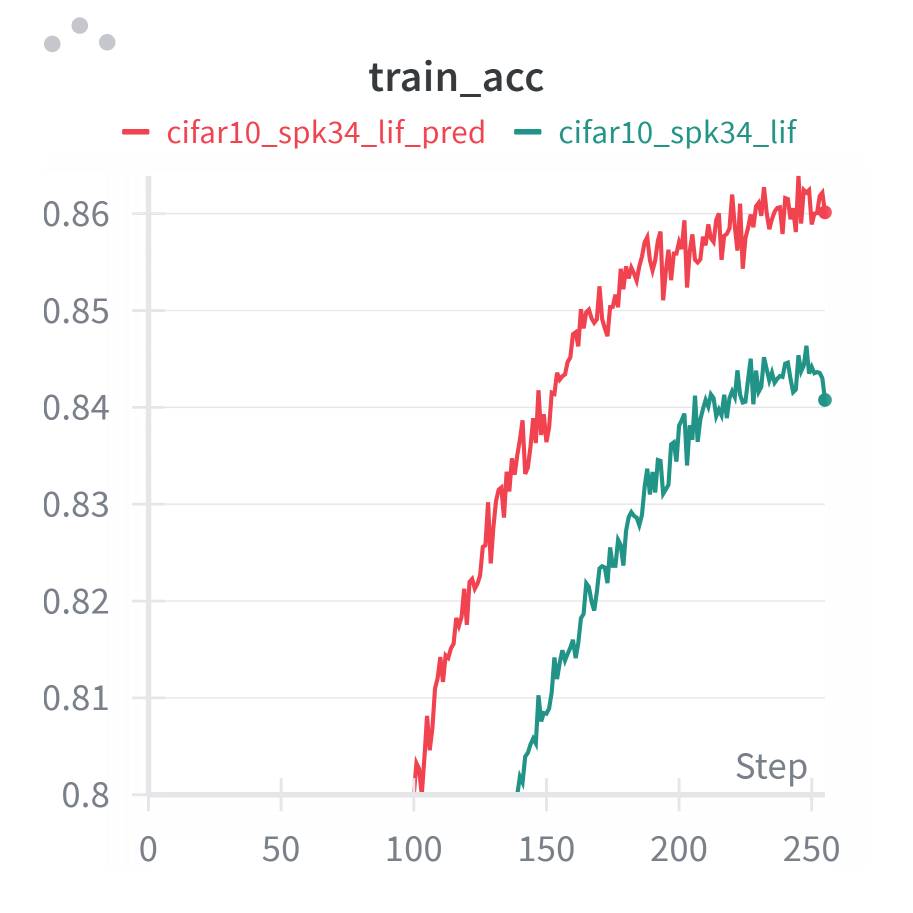}
\includegraphics[width=0.25\columnwidth, trim=0.0cm 0.0cm 0.0cm 0.0cm, clip]{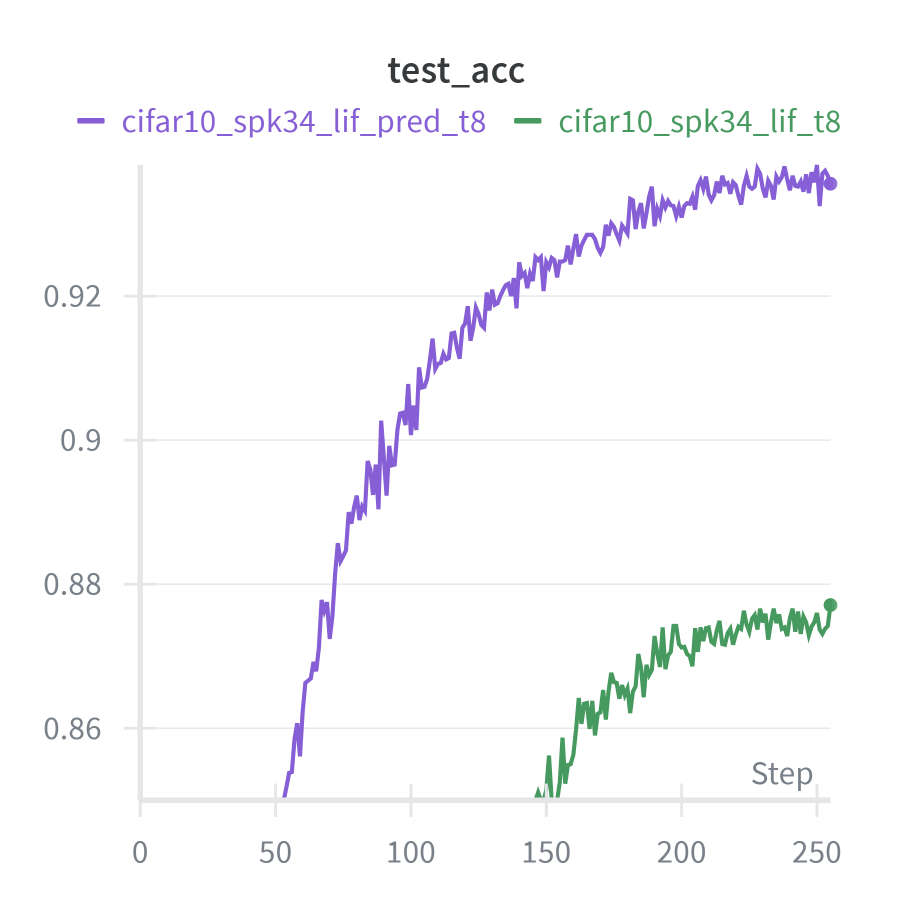}
\includegraphics[width=0.25\columnwidth, trim=0.0cm 0.0cm 0.0cm 0.0cm, clip]{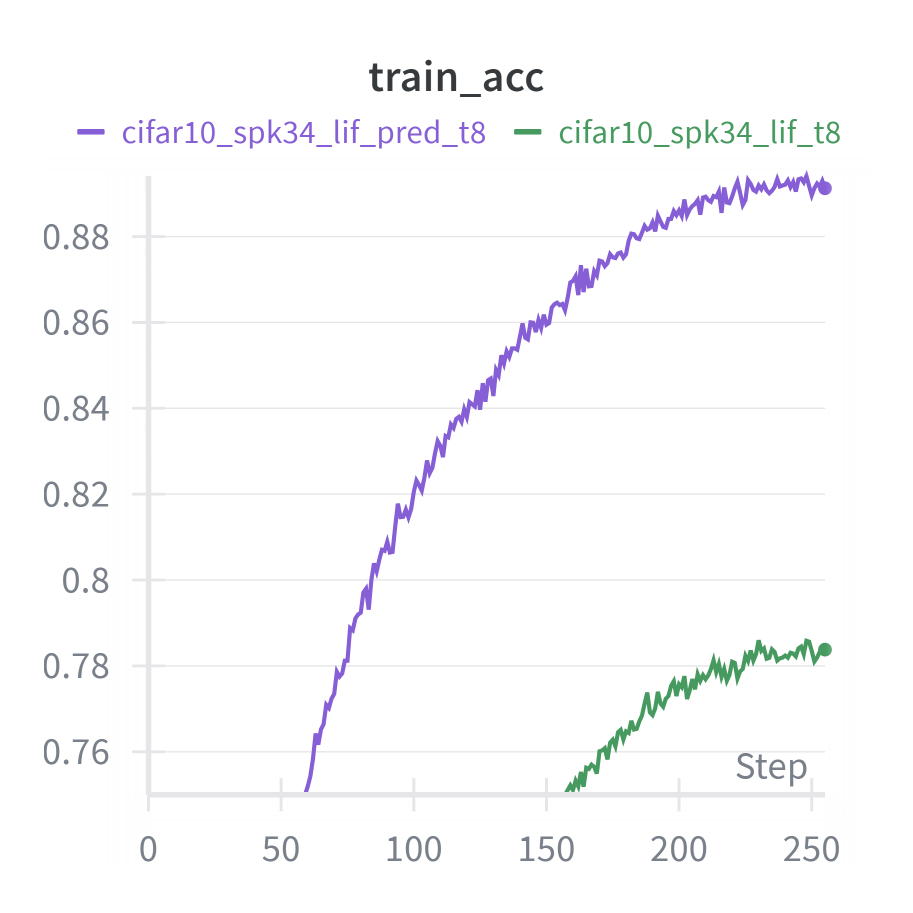}
}
\vskip -0.05in
\centerline{\includegraphics[width=0.25\columnwidth, trim=0.0cm 0.0cm 0.0cm 0.0cm, clip]{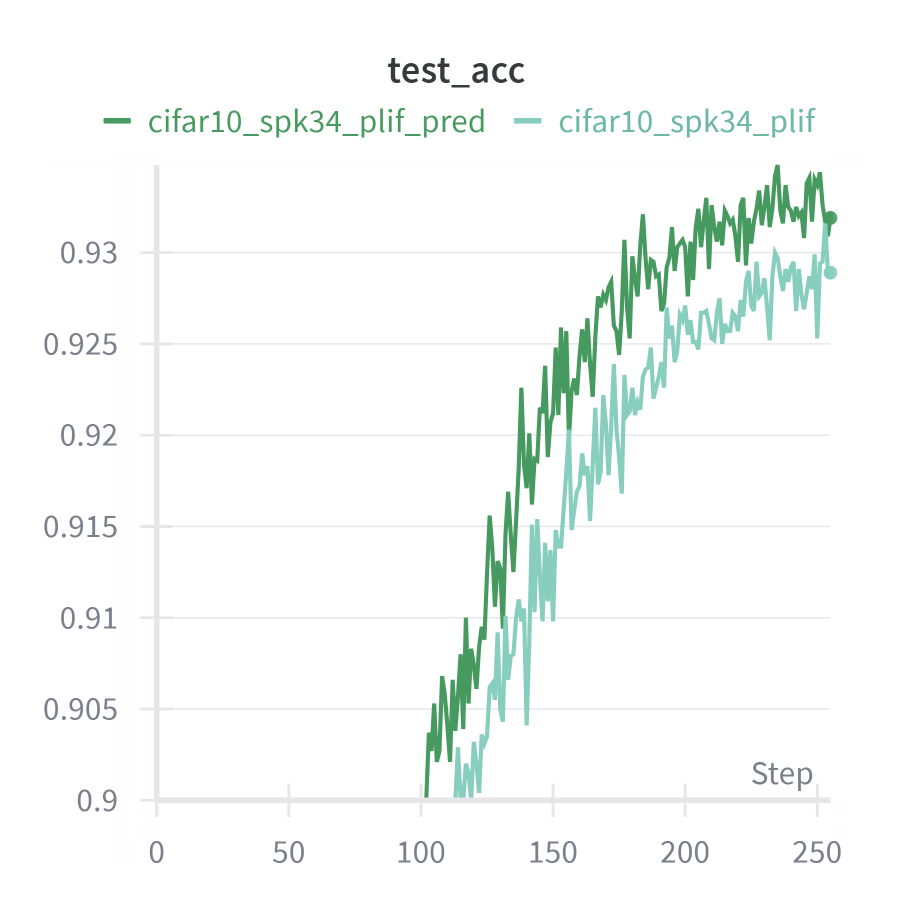}
\includegraphics[width=0.25\columnwidth, trim=0.0cm 0.0cm 0.0cm 0.0cm, clip]{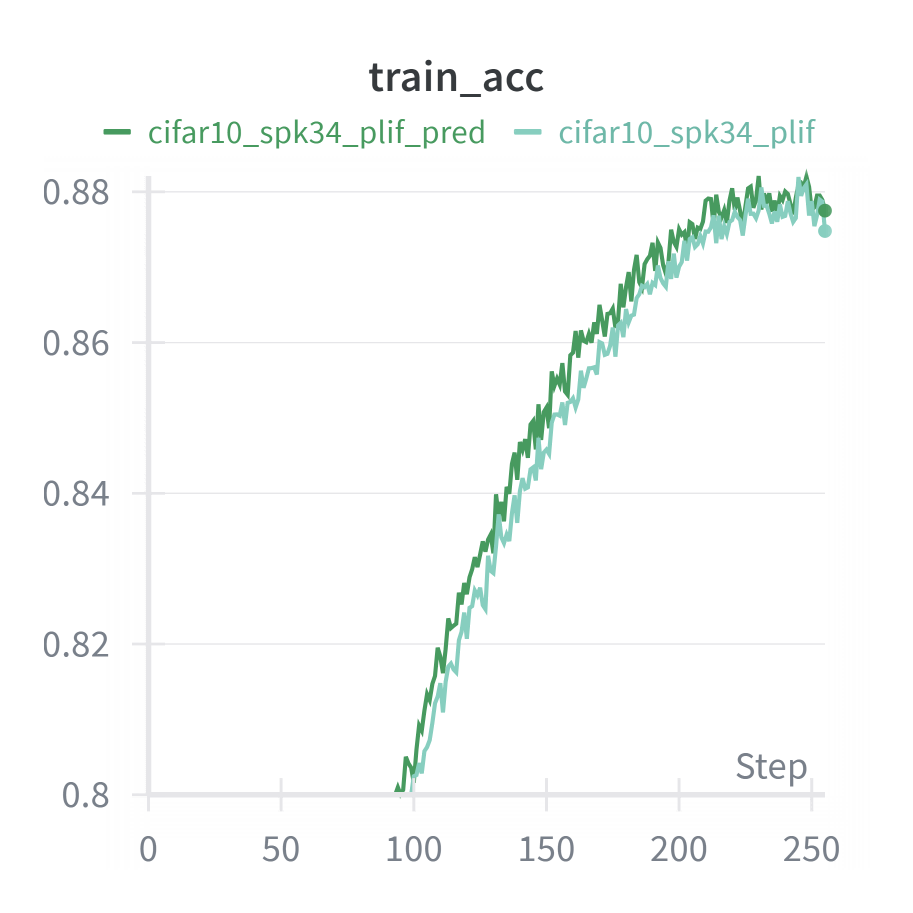}
\includegraphics[width=0.25\columnwidth, trim=0.0cm 0.0cm 0.0cm 0.0cm, clip]{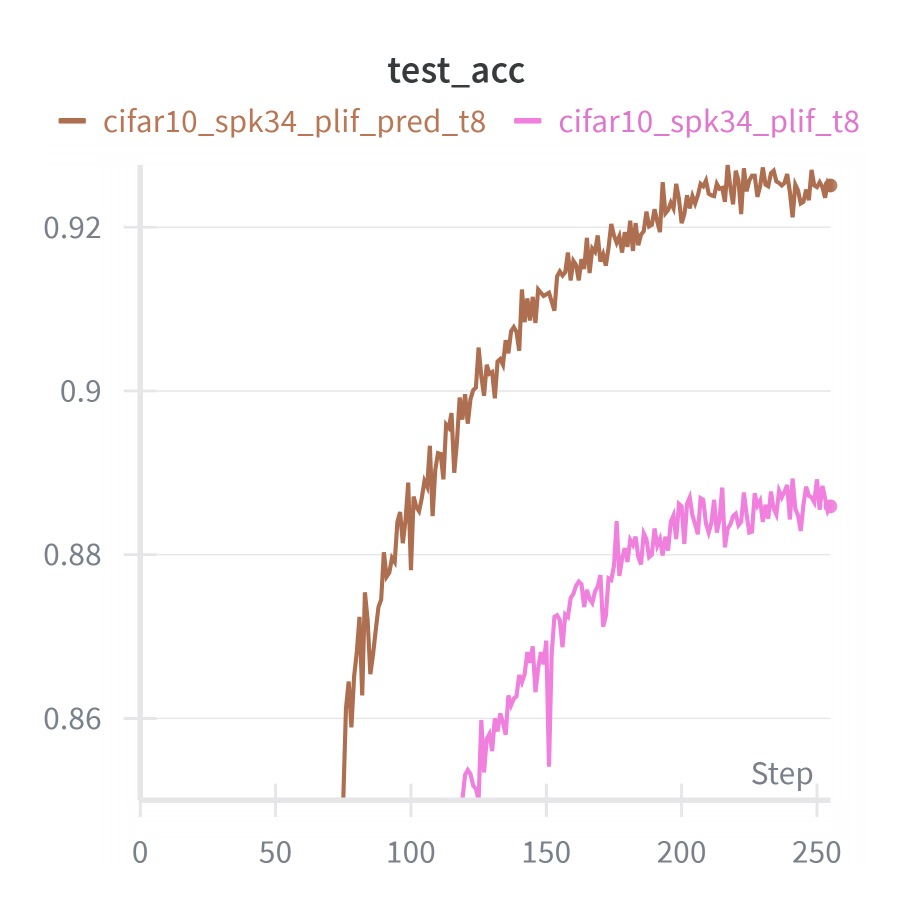}
\includegraphics[width=0.25\columnwidth, trim=0.0cm 0.0cm 0.0cm 0.0cm, clip]{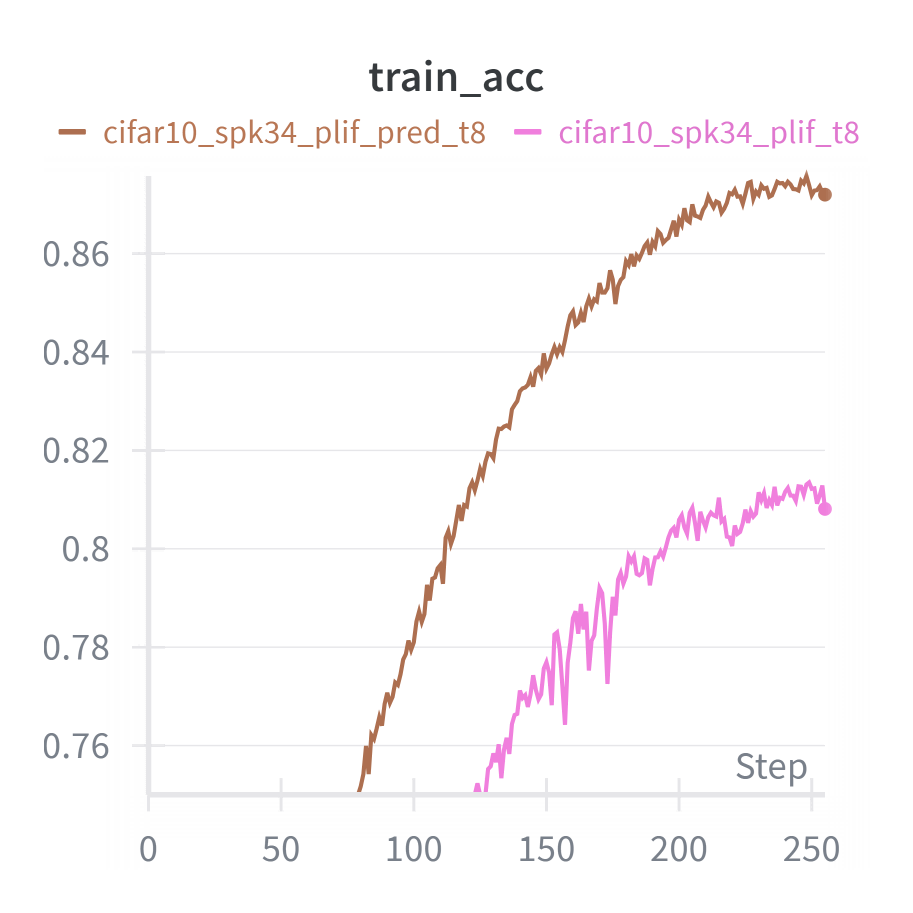}
}
\vskip -0.05in
\centerline{\includegraphics[width=0.25\columnwidth, trim=0.0cm 0.0cm 0.0cm 0.0cm, clip]{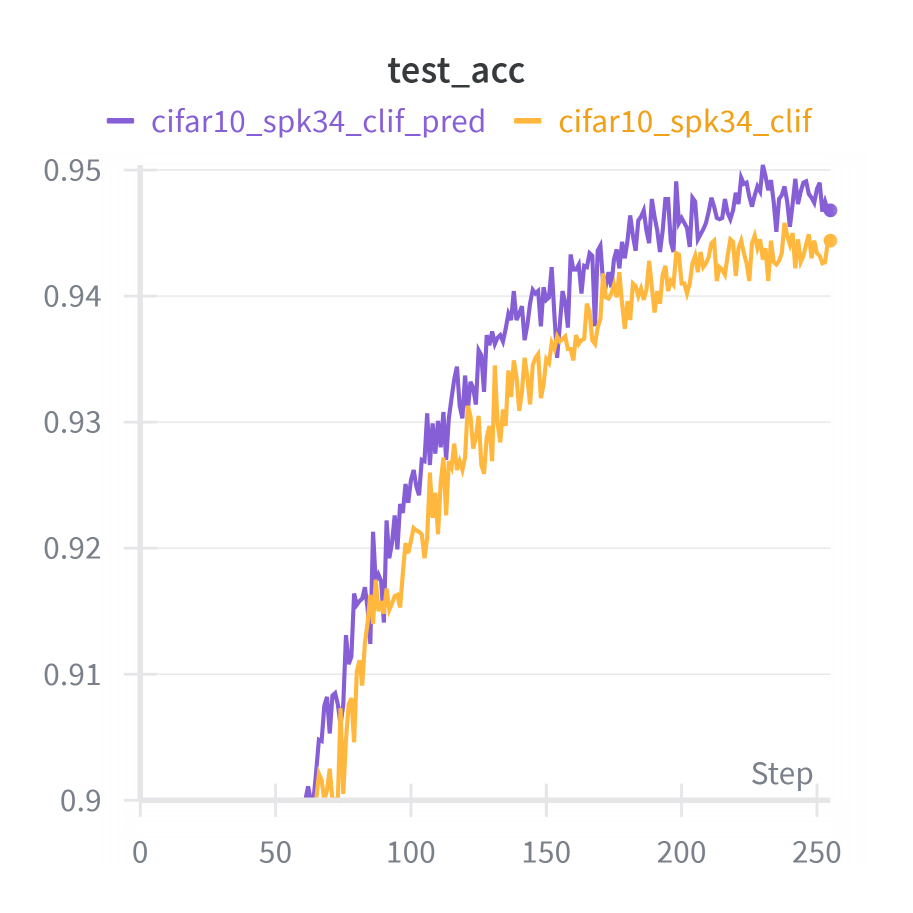}
\includegraphics[width=0.25\columnwidth, trim=0.0cm 0.0cm 0.0cm 0.0cm, clip]{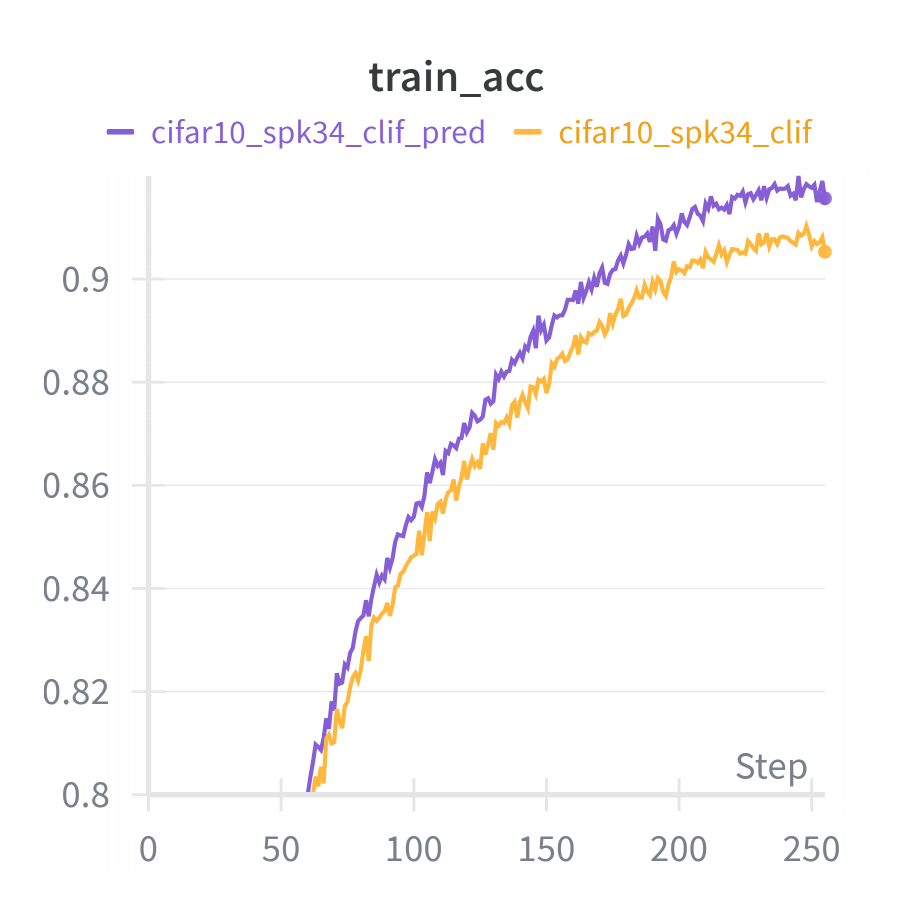}
\includegraphics[width=0.25\columnwidth, trim=0.0cm 0.0cm 0.0cm 0.0cm, clip]{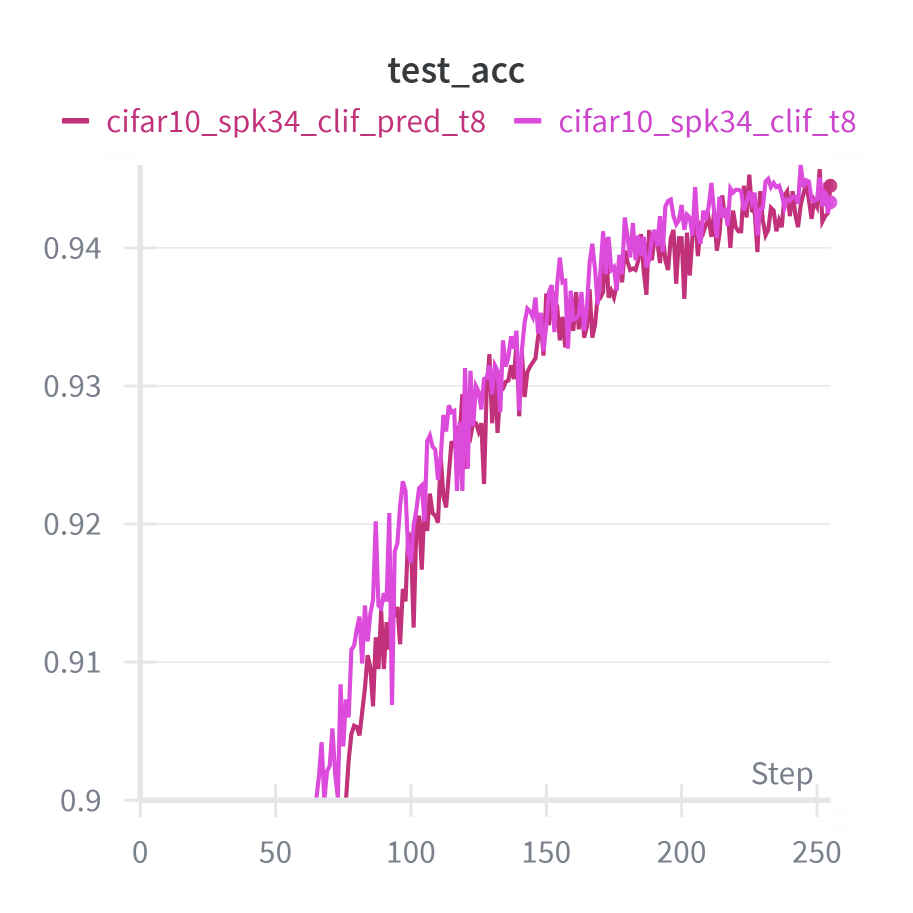}
\includegraphics[width=0.25\columnwidth, trim=0.0cm 0.0cm 0.0cm 0.0cm, clip]{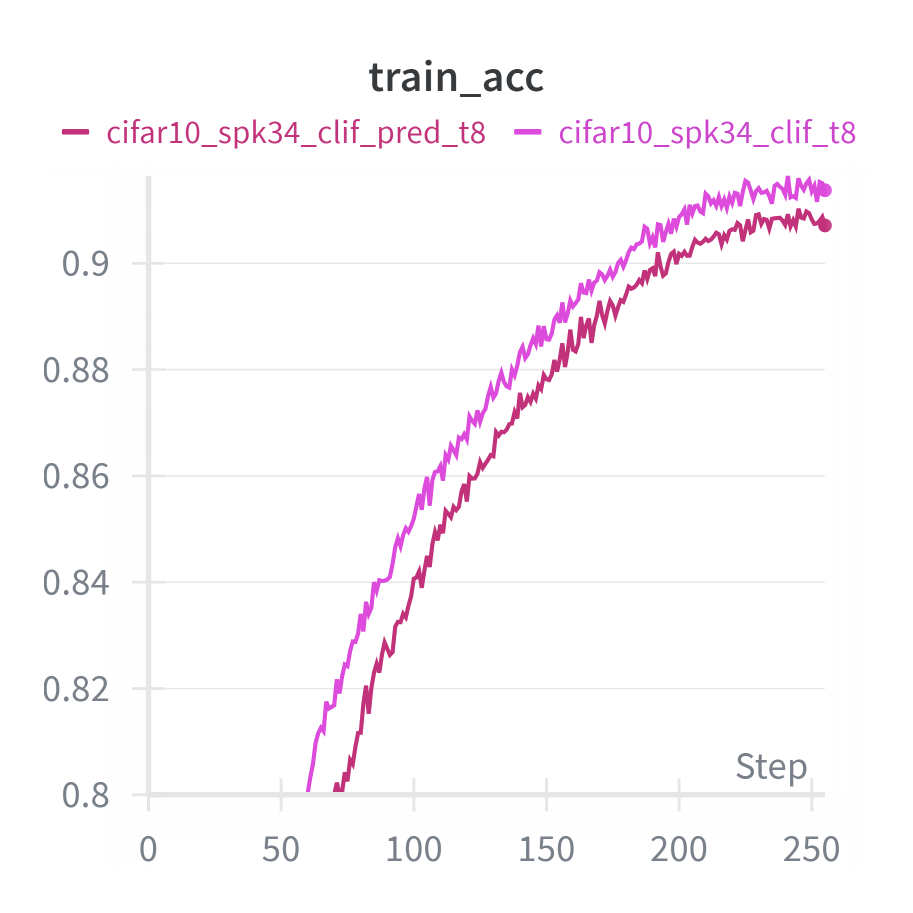}
}
\vskip -0.1in
\caption{The training and testing accuracy curves of the Spiking ResNet34 trained on CIFAR10 dataset with IF,LIF,PLIF and CLIF neurons along with their variants enhanced by self-prediction mechanisms at time-steps T=4 and T=8.}
\label{pic:cifar10_spk34}
\end{center}
\end{figure}

For the Spiking ResNet families which is originally designed for ImageNet, when adapting them to CIFAR-10, we also modify the first convolutional layer parameters, changing kernel size, stride, and padding from 7, 2, 3 to 3, 1, 1, respectively, and replace the initial max pooling layer with an identity mapping. Figure~\ref{pic:cifar10_spk18} shows the training and testing accuracy curve on Spiking ResNet18. Figure~\ref{pic:cifar10_spk34} shows the training and testing accuracy curve on Spiking ResNet34.

\clearpage
\section{Results on ImageNet100 and ImageNet1k Dataset}
\begin{figure}[h]
\begin{center}
\vskip -0.15in
\centerline{\includegraphics[width=0.4\columnwidth, trim=0.0cm 0.0cm 0.0cm 0.0cm, clip]{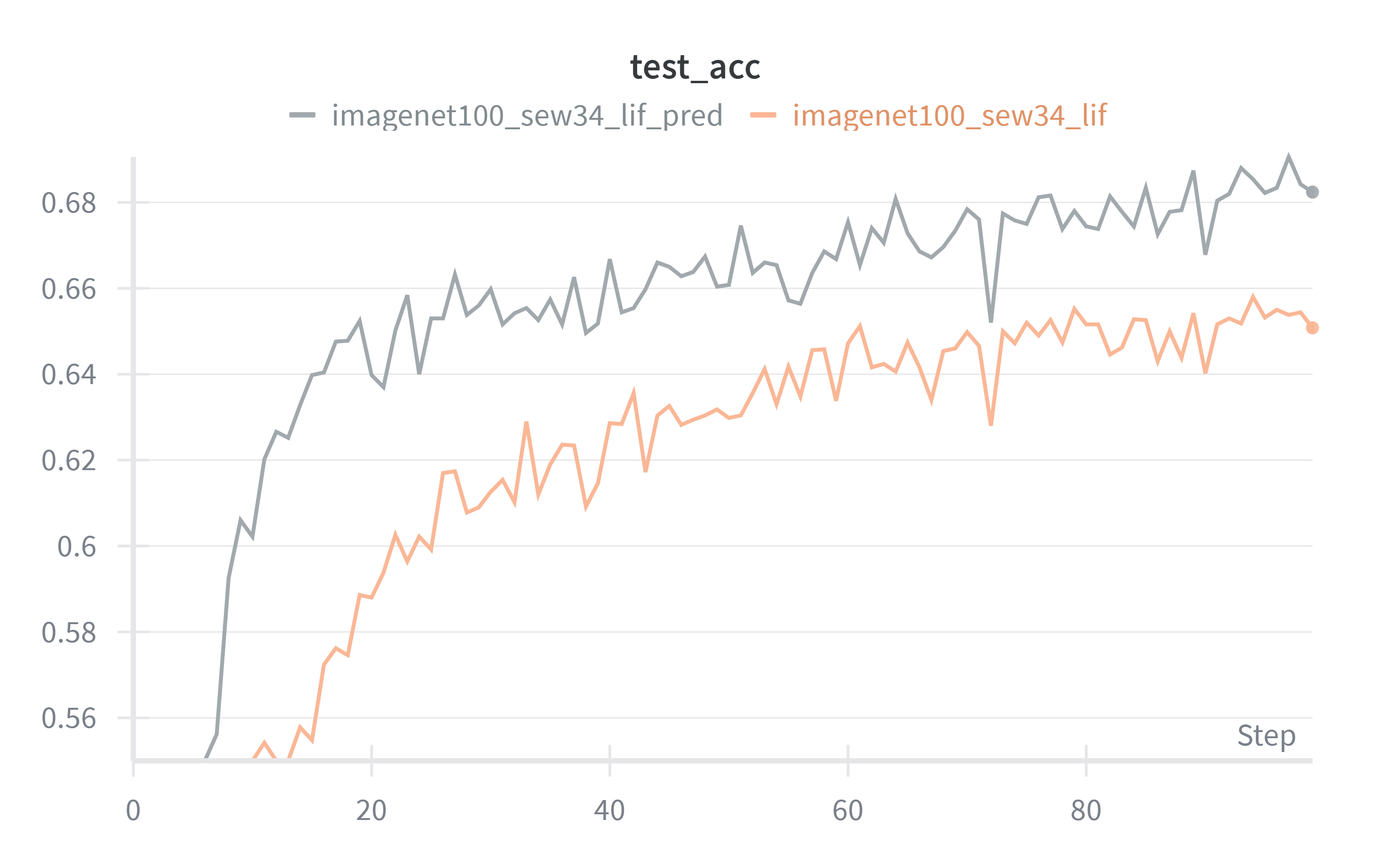}
\includegraphics[width=0.4\columnwidth, trim=0.0cm 0.0cm 0.0cm 0.0cm, clip]{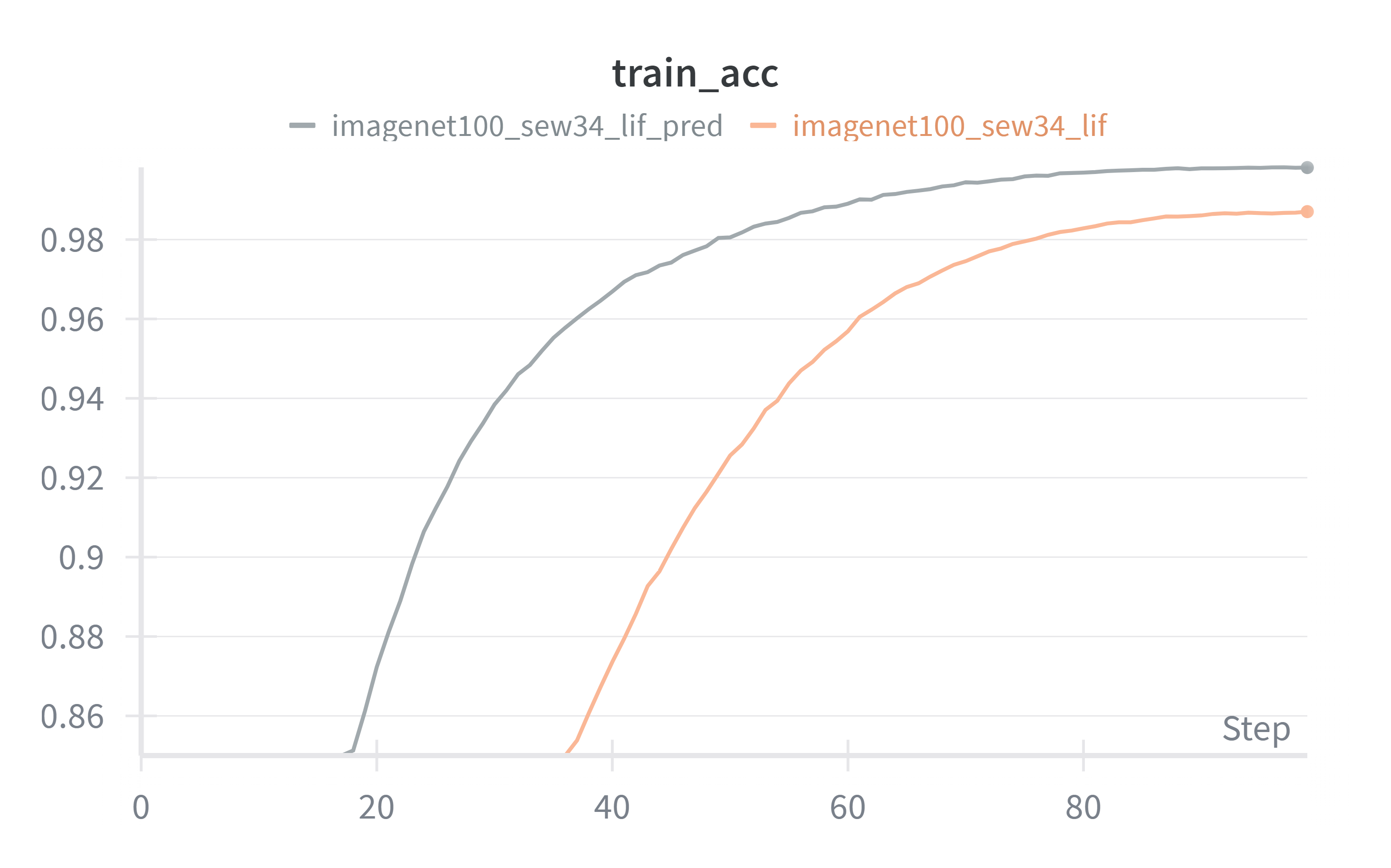}
}
\vskip -0.05in
\centerline{\includegraphics[width=0.4\columnwidth, trim=0.0cm 0.0cm 0.0cm 0.0cm, clip]{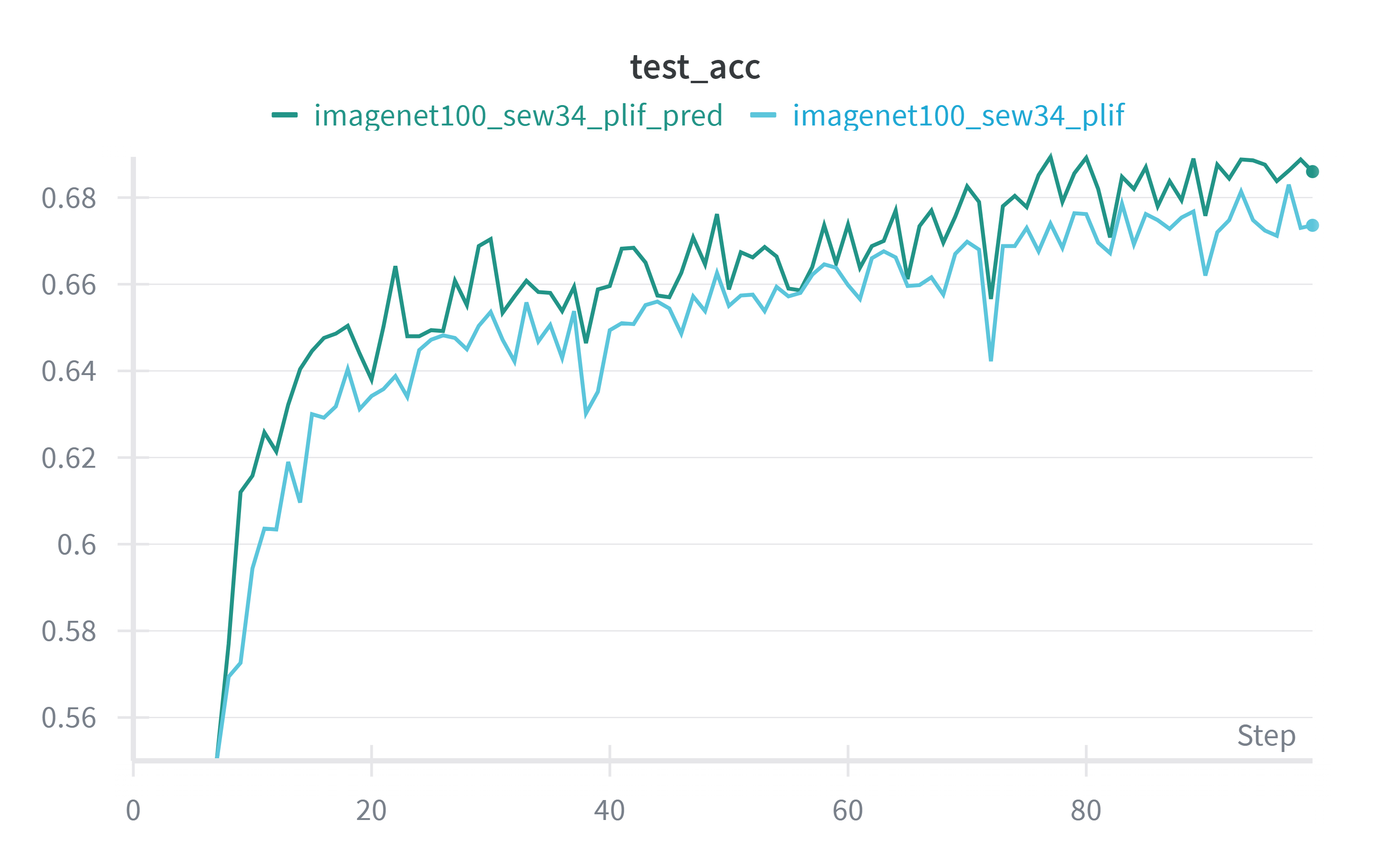}
\includegraphics[width=0.4\columnwidth, trim=0.0cm 0.0cm 0.0cm 0.0cm, clip]{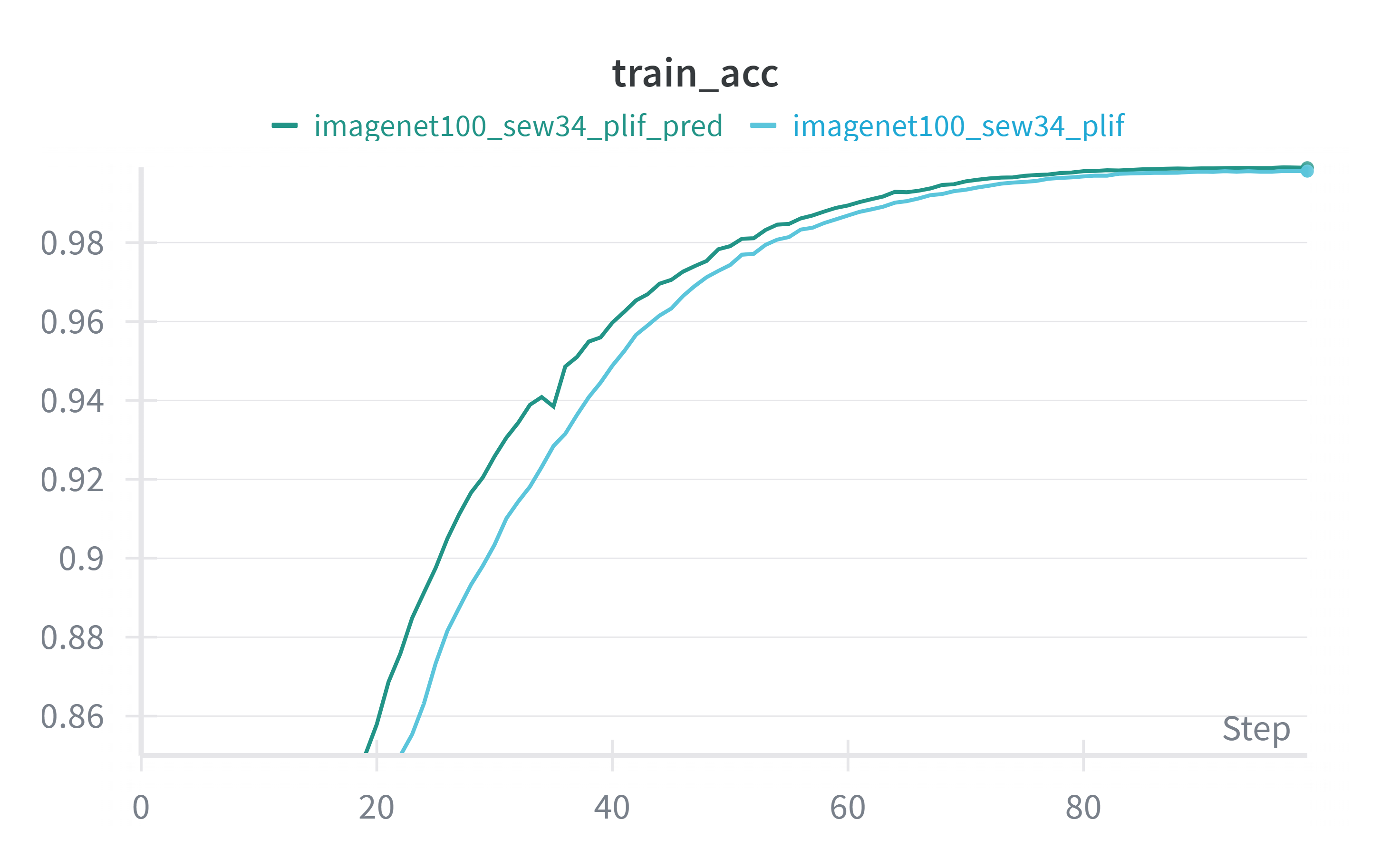}
}
\vskip -0.05in
\centerline{\includegraphics[width=0.4\columnwidth, trim=0.0cm 0.0cm 0.0cm 0.0cm, clip]{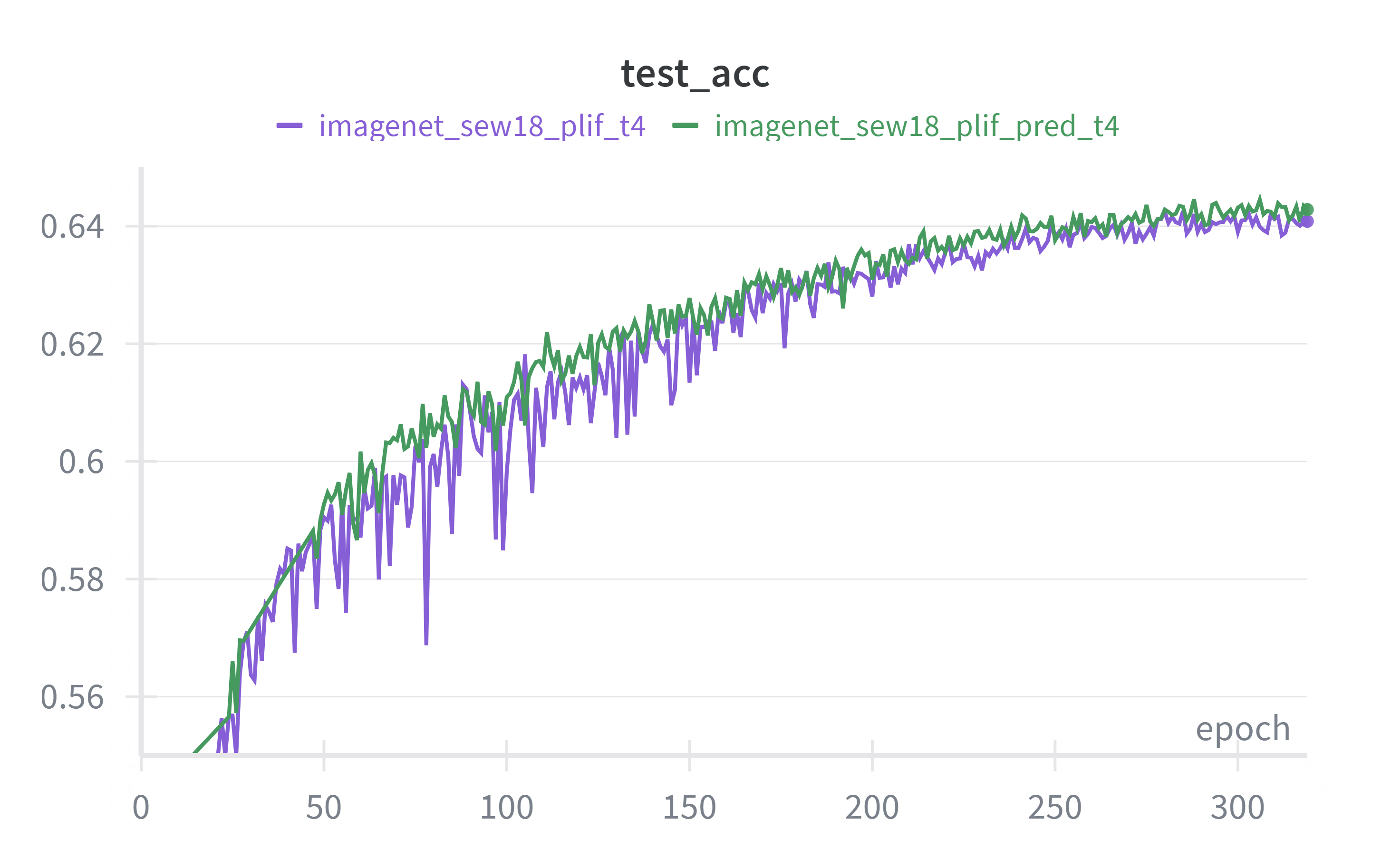}
\includegraphics[width=0.4\columnwidth, trim=0.0cm 0.0cm 0.0cm 0.0cm, clip]{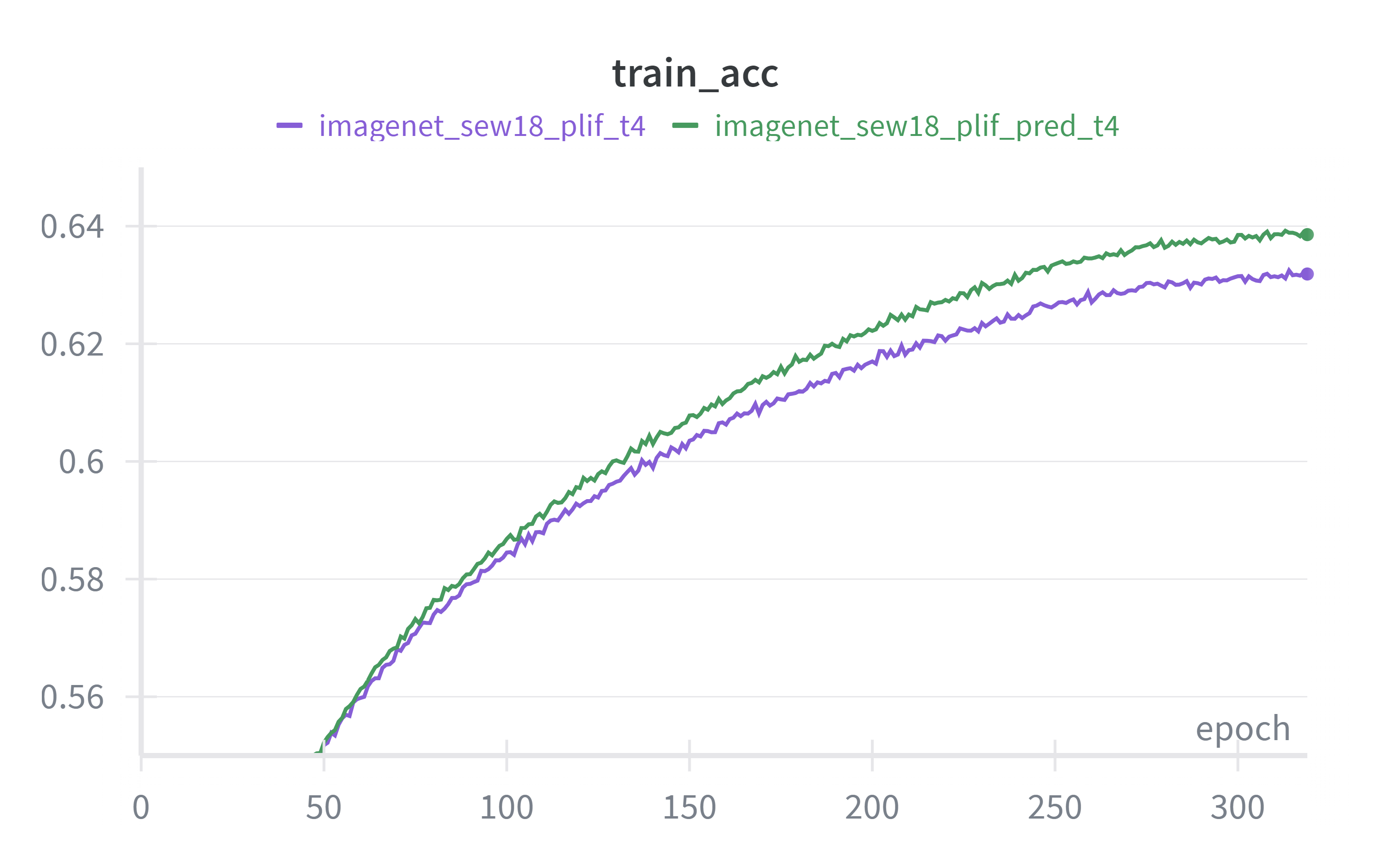}
}
\vskip -0.1in
\caption{Training and testing accuracy curves on ImageNet100 and ImageNet dataset.}
\label{pic:imagenet_curve}
\end{center}
\end{figure}

Due to computational resource constraints, we evaluated the performance gains of the self-prediction enhancement method using the SEW ResNet-34 model on ImageNet-100 with both LIF and PLIF neurons. Additionally, we tested its effectiveness with PLIF neurons on the full ImageNet dataset using the SEW ResNet-18 model. The corresponding training and testing curves are shown in the Figure~\ref{pic:imagenet_curve}.

\clearpage
\section{Results on Sequential CIFAR10 Classification Tasks}
\begin{figure}[h]
\begin{center}
\vskip -0.15in
\centerline{\includegraphics[width=0.4\columnwidth, trim=0.0cm 0.0cm 0.0cm 0.0cm, clip]{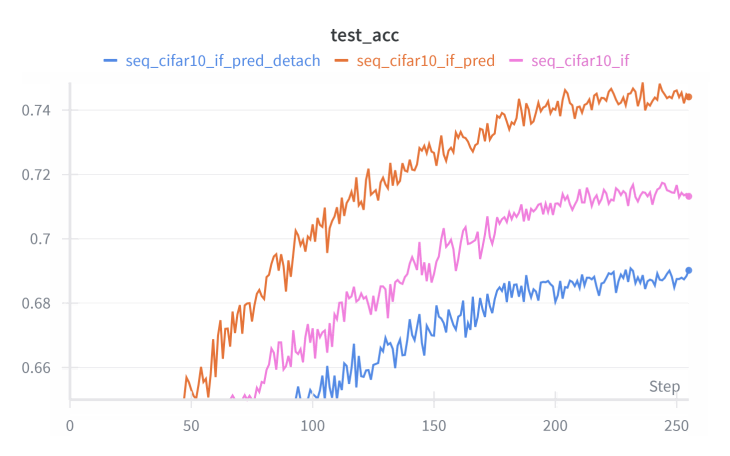}
\includegraphics[width=0.4\columnwidth, trim=0.0cm 0.0cm 0.0cm 0.0cm, clip]{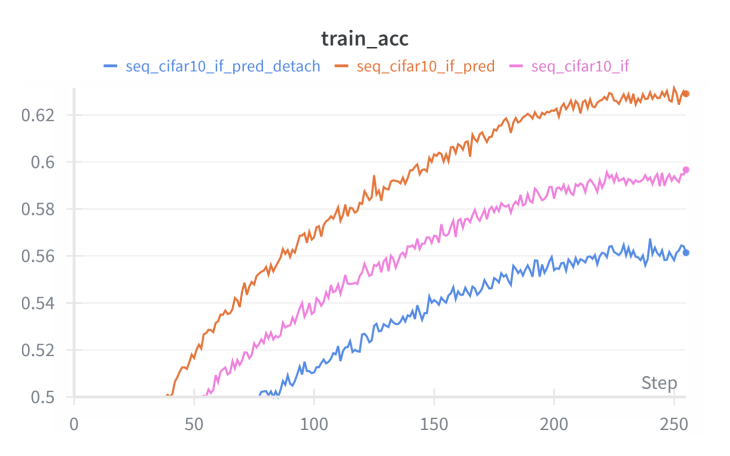}
}
\vskip -0.05in
\centerline{\includegraphics[width=0.4\columnwidth, trim=0.0cm 0.0cm 0.0cm 0.0cm, clip]{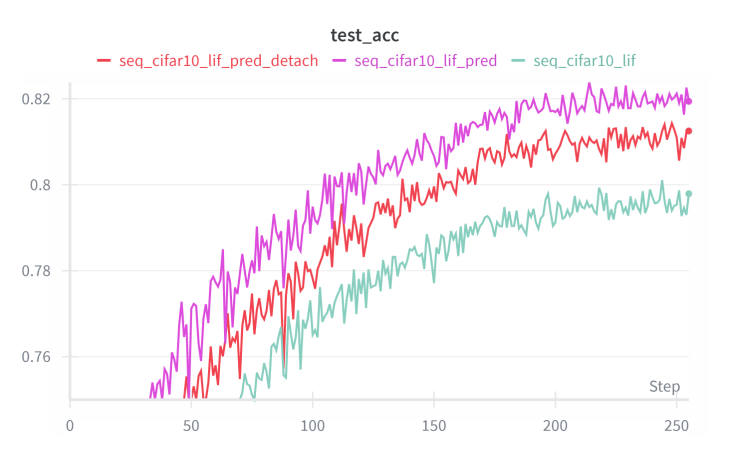}
\includegraphics[width=0.4\columnwidth, trim=0.0cm 0.0cm 0.0cm 0.0cm, clip]{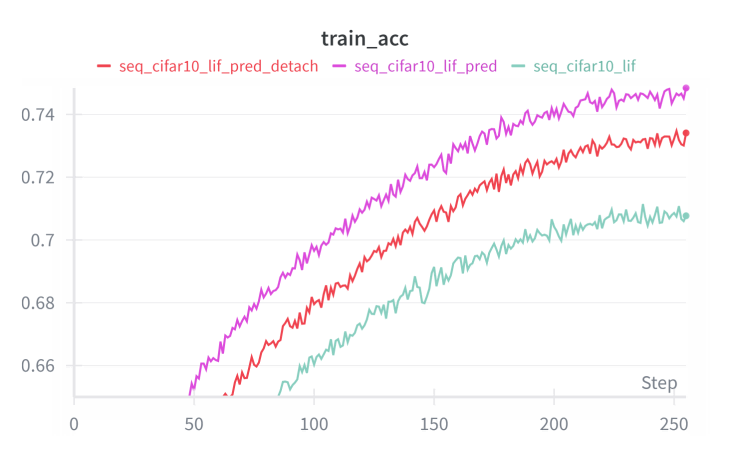}
}
\vskip -0.05in
\centerline{\includegraphics[width=0.4\columnwidth, trim=0.0cm 0.0cm 0.0cm 0.0cm, clip]{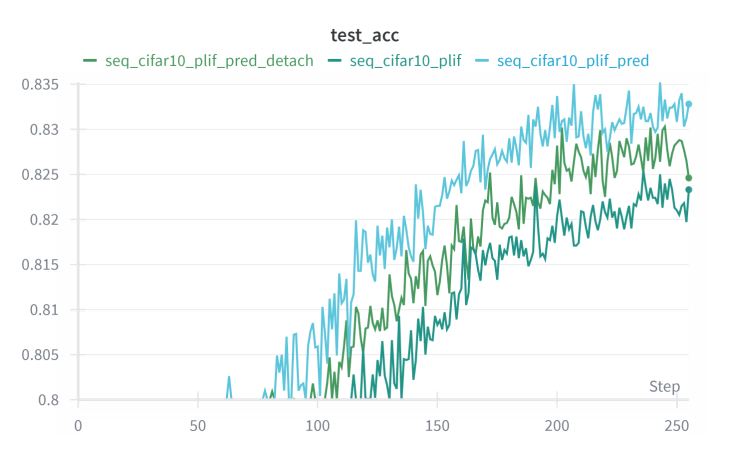}
\includegraphics[width=0.4\columnwidth, trim=0.0cm 0.0cm 0.0cm 0.0cm, clip]{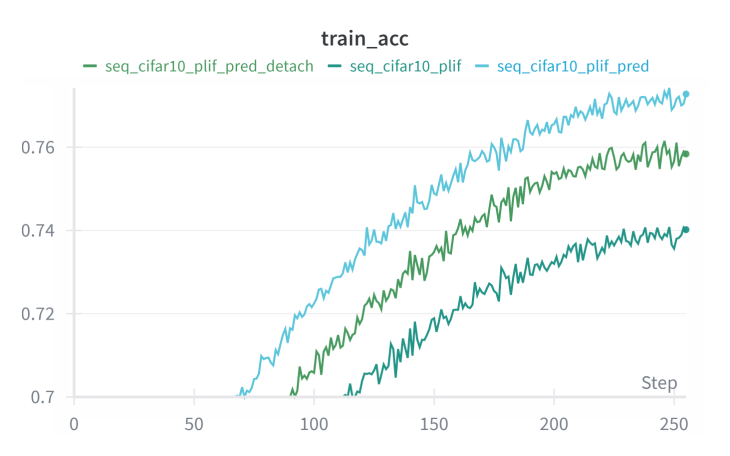}
}
\vskip -0.05in
\centerline{\includegraphics[width=0.4\columnwidth, trim=0.0cm 0.0cm 0.0cm 0.0cm, clip]{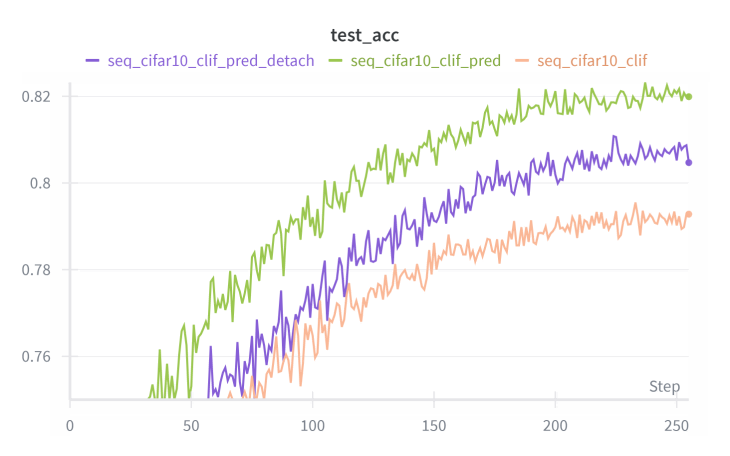}
\includegraphics[width=0.4\columnwidth, trim=0.0cm 0.0cm 0.0cm 0.0cm, clip]{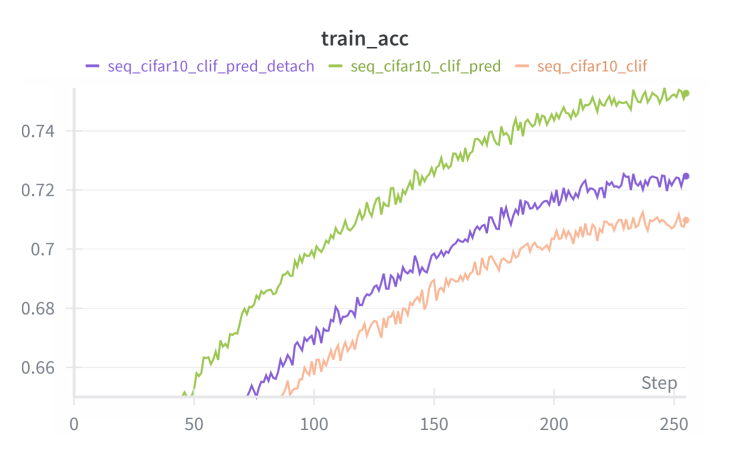}
}
\vskip -0.1in
\caption{Training and testing accuracy curves on sequential CIFAR-10 using IF, LIF, PLIF, and CLIF neurons, including their self-prediction enhanced variants and an ablation on spike detachment in the computation of $\bm{m}^l_p[t]$.}
\label{pic:seq_cifar10_all}
\end{center}
\end{figure}

For Sequential CIFAR10 Classification Tasks, we use our CIFAR10NetSeq architecture , which processes input images in a sequential manner by treating the spatial width dimension as time steps. It consists of the following components in sequence: two macro stages, each containing three convolutional blocks followed by an average pooling layer. Each convolutional block comprises Conv1d(inchannels, channels=128, kernelsize=3, padding=1, bias=False) to BatchNorm1d(128) to spiking neuron. The first block uses 3 input channels (RGB), while all subsequent blocks use 128 channels. After every three convolutional blocks, an AvgPool1d(2) reduces the temporal length by half. Following the convolutional stack, the feature sequence is flattened and passed through two fully connected layers: Linear(128 × 8, 256) with a spiking neuron, then Linear(256, 10) without a neuron. Figure~\ref{pic:seq_cifar10_all} shows the training and testing accuracy curve.



\section{Experiment details of Reinforcement Learning Tasks}
\label{app:exp_para}
\subsection{Specific Parameters for the Proxy Target Framework} 

Table~\ref{tab:proxy} lists the hyperparameters of the proxy target framework for different spiking neuron types. To better capture the dynamics of the SNN, the proxy network uses wider hidden layers than the corresponding online SNN. Because different spiking neurons exhibit distinct temporal dynamics and learning characteristics, both the hidden sizes and learning rates of the proxy network are adjusted accordingly across neuron types. All other hyperparameters remain identical.

\begin{table}[htbp]
  \caption{Hyper-parameters of the proxy network framework with different spiking neurons}
  \label{tab:proxy}
  \centering
  \begin{tabular}{lc}
  \toprule
 Parameter& LIF\\ 
 \midrule
 Proxy network architecture &$(512,512)$\\
 Proxy network activation& ReLU\\
  Proxy network learning rate& $3\cdot 10^{-3}$\\
 Proxy network optimizer & Adam\\
 Proxy update iterations $K$& $1$\\
 Proxy update batch size $N$& $256$\\
 \bottomrule
  \end{tabular}
\end{table}

\subsection{Spiking Actor Network Parameters}
All hyperparameters of the spiking actor network are summarized in Table~\ref{tab:SAN}, following the same configuration as in a wide range of prior studies \citep{popSAN}.

\begin{table}[htbp]
  \caption{Hyper-parameters of the spiking actor network}
  \label{tab:SAN}
  \centering
  \begin{tabular}{lc}
  \toprule
 Parameter& Value\\ 
 \midrule
 Encoder population per dimension $N_{in}$&$10$\\
 Encoder threshold $V_E$& $0.999$\\
  Network hidden units& $(256,256)$\\
 Decoder population per dimension $N_{out}$& $10$\\
 Surrogate gradient window size $\omega$& $0.5$\\
 \bottomrule
  \end{tabular}
\end{table}

\subsection{RL algorithm parameters} 
We conduct our experiments based on the TD3 algorithm \citep{TD3}, using the hyperparameters listed in Table~\ref{tab:TD3}.

\begin{table}[htbp]
  \caption{Hyper-parameters of the implemented TD3 algorithm \citep{TD3}}
  \label{tab:TD3}
  \centering
  \begin{tabular}{lc}
  \toprule
 Parameter& Value \\
 \midrule
 Actor learning rate& $3\cdot10^{-4}$\\
  Actor regularization & None\\
 Critic learning rate& $3\cdot10^{-4}$\\
 Critic regularization & None\\
 Critic architecture &$(256,256)$\\
 Critic activation &ReLU\\
 Optimizer & Adam\\
 Target update rate $\tau$& $5\cdot10^{-3}$\\
 Batch size $N$& $256$\\
 Discount factor $\gamma$& $0.99$\\
 Iterations per time step&$1.0$ \\
 Reward scaling & $1.0$ \\
 Gradient clipping & None \\
 Replay buffer size& $10^{6}$ \\
 Exploration noise $\mathcal{N}(0,\sigma)$& $\mathcal{N}(0,0.1)$\\
 Actor update interval $d$& $2$\\
 Target policy noise $\mathcal{N}(0,\tilde{\sigma})$& $\mathcal{N}(0,0.2)$\\
 Target policy noise clip $c$& $0.5$\\
 \bottomrule
  \end{tabular}
\end{table}

\subsection{Experiment environments}
All environments were used with their default configurations without any modifications.  
Notably, since the state vector can take values in $-\infty$ to $\infty$, it is normalized to the range $(-1,1)$ using a tanh function. Similarly, because actions are bounded by predefined minimum and maximum limits, the actor network’s output is first passed through a tanh function to constrain it to $(-1,1)$, and then linearly scaled to the actual action range $(\text{Min action},\text{Max action})$.

\end{document}